%% file: predictive_representations_of_state_and_knowledge.tex
\pdfoutput=1
\documentclass[notitlepage]{article}
\usepackage{amsfonts}
\usepackage{amsmath}
\usepackage{amssymb}
\usepackage{multirow}
\usepackage{fullpage}

\usepackage[T1]{fontenc}
\usepackage{float}
\usepackage{subfig}
\usepackage{graphicx}
\usepackage{booktabs}
\usepackage{xparse}
\usepackage{etoolbox}
\usepackage{fmtcount}
\usepackage{pgf}
\usepackage[stable]{footmisc}
\newwrite\file
\immediate\openout\file=mydebugging.txt
\immediate\write\file{A line of text to write to the file}
\immediate\write\file{Another line of text to write to the file}

\input{common}

\input{PRSK-latex-commands}
\input{latex-forecast-handling}

\author{Mark Ring}
\date{}
\title{Representing Knowledge as Predictions \\ (and State as Knowledge)\footnote{Other than a small number of corrections and improvements suggested by generous colleagues, this paper has been in its present state since roughly 2013. Thus, some aspects of it now seem dated; for example, GVFs (aka ``forecasts'') and off-policy learning are relatively well known now, which they were not when the paper was written. As a result, many references could be updated to reflect the last decade of progress. Nevertheless, I believe this paper still has many useful insights to offer the community, especially the growing community of enthusiastic researchers in Continual Learning. }}
\begin{document}
\maketitle

\input{forecasts}

\input{options}
\input{aliases}

\input{abstract}
\input{motivation}

\input{framework}
\input{demonstration}

\section{Acknowledgments}
\input{acknowledgements}

\appendix
\section{Appendix}

\input{appendix-forward-view}
\input{appendix-collected-tables}

\section{References}
\bibliography{All}
\bibliographystyle{plain}
\closeout\file
\end{document}

%% file: common.tex
\usepackage[mathletters]{ucs}
\usepackage[utf8x]{inputenc}
\newcommand{\beq}{\begin{equation}}
\newcommand{\eeq}{\end{equation}}
\newcommand{\beqn}{\begin{align}}
\newcommand{\eeqn}{\end{align}}
\newcommand{\be}{\begin{eqnarray*}}
\newcommand{\ee}{\end{eqnarray*}}
\newcommand{\ba}{\[\begin{aligned}}
\newcommand{\ea}{\end{aligned}\]}
\usepackage{color}
\ifx \comment \undefined
  \newcommand{\comment}[1]{\textcolor{blue}{\{{#1}\}}}
\fi

\newcommand{\fncomment}[1]{\footnote{\textcolor{blue}{\{{#1}\}}}}
\newcommand{\commentOut}[1]{}

\newcommand{\CalA}{\mathcal{A}}

\newcommand{\CalO}{\mathcal{O}}
\newcommand{\CalP}{\mathcal{P}}

\newcommand{\CalR}{\mathcal{R}}
\newcommand{\CalS}{\mathcal{S}}

\newcommand{\Reals}{\mathbb{R}}
\newcommand{\ra}{\rightarrow}
\newcommand{\Exp}{\mathbb{E}} 
\ifx \argmax \undefined
  \DeclareMathOperator*{\argmax}{arg\!max}
\fi
\newcommand{\ie}{{\it i.e.,} }
\newcommand{\etc}{{\it etc.}}

\newcommand{\forlater}[1]{\textcolor{purple}{\{{ttd: #1}\}}}

\newcommand{\ttd}[1]{\forlater{#1}}

%% file: PRSK-latex-commands.tex
\newcommand{\Authors}[1]{#1}
\newcommand{\I}{\Authors{I }} 
\newcommand{\my}{\Authors{my }} 
\newcommand{\am}{\Authors{am }} 

\newcommand{\IA}{\Authors{I }} 
\newcommand{\myA}{\Authors{my }} 
\newcommand{\MyA}{\Authors{My }} 

\newcommand{\WeReaders}[1]{#1}
\newcommand{\us}{\WeReaders{us }} 
\newcommand{\our}{\WeReaders{our }} 

\newcommand{\WeH}{\WeReaders{We }} 
\newcommand{\weH}{\WeReaders{we }} 
\newcommand{\usH}{\WeReaders{us}} 
\newcommand{\ourH}{\WeReaders{our }} 

\newcommand{\YouReader}[1]{{#1}}
\newcommand{\you}{\YouReader{you }} 

\newcommand{\commentout}[1]{}

\renewcommand{\vec}[1]{{\bf {{#1}}}}
\newcommand{\concat}{\circ}
\newcommand{\cumu}{cumulating}

\newcommand{\state}{\ensuremath s}
\newcommand{\Estate}{\ensuremath\hat\state}
\newcommand{\allStates}{\ensuremath \CalS}
\newcommand{\absorbingState}{\ensuremath \state^\circ}
\newcommand{\action}{\ensuremath a}

\newcommand{\allActions}{\ensuremath \CalA}
\newcommand{\obs}{\ensuremath o}
\newcommand{\obsvec}{\ensuremath \vec{o}}
\newcommand{\allObs}{\ensuremath \CalO}
\newcommand{\reward}{\ensuremath r}
\newcommand{\Reward}{\ensuremath R}
\newcommand{\allRewards}{\ensuremath \Reals}
\newcommand{\policy}{π}

\newcommand{\accv}{\ensuremath c}
\newcommand{\termv}{\ensuremath z}
\newcommand{\termP}{\ensuremath \beta}
\newcommand{\forecastfun}{\ensuremath f}
\newcommand{\forecast}{\ensuremath \hat\forecastfun}
\newcommand{\allForecastfunctions}{\ensuremath F}
\newcommand{\allForecasts}{\ensuremath \hat\allForecastfunctions}

\newcommand{\transitions}{\ensuremath T}
\newcommand{\prob}{\Pr}
\newcommand{\target}{outcome}

\newcommand{\Target}{Outcome}

\newcommand{\mtarget}{\xi}
\newcommand{\statevec}{\vec{s}}
\newcommand{\Estatevec}{\ensuremath\hat\statevec}

\newcounter{stagecounter}
\newcommand{\subheading}[1]{\vspace{10pt}  \noindent {\it #1.}}

\NewDocumentCommand\case{>{\SplitList{;}}m}
{
  \hfill
    \begin{cases}
      \ProcessList{#1}{ \processcaseitems }
    \end{cases}
  }

\DeclareDocumentCommand \processcaseitems
   { > { \SplitArgument{1} {,} } m}
   {\insertcaseitem #1}

\newcommand\insertcaseitem[2]{#1 & \mbox{#2} \\}

\usepackage{minibox}
\newcommand{\mysplit}[1]{\minibox[c]{#1}}

-lmtk10
\newcommand{\newlayer}{
  \refstepcounter{stagecounter}%
  \subsection*{Layer \thestagecounter}%
}
\newcommand{\nlayers}{12}
\newcommand{\option}{\emph}

\newenvironment{my_list}{
\begin{samepage}
\begin{list}{}{}
  \setlength{\itemsep}{1pt}
  \setlength{\parskip}{0pt}
  \setlength{\parsep}{0pt}}{\end{list}\end{samepage}
}
\newenvironment{my_enumerate}{
\begin{samepage}
\begin{enumerate}
  \setlength{\itemsep}{1pt}
  \setlength{\parskip}{0pt}
  \setlength{\parsep}{0pt}}{\end{enumerate}\end{samepage}
}


%% file: latex-forecast-handling.tex
\newcommand{\compoundname}[1]{\csname #1\endcsname}
\makeatletter
\newcommand{\myassign}[2]{\protected@edef#1{#2}}
\newcommand{\@assign}[2]{\protected@edef#1{#2}}
\newcommand{\assign}[2]{
  \expandafter\@assign\expandafter{\csname #1\endcsname}{#2}
  }
\makeatother

\DeclareExpandableDocumentCommand \first
  {m}
  {\ifnum \mytemp=1 #1
    \mytemp=0
    \fi
  }

\newcounter{forecastcounter}
\newcounter{forecaststart}
\newcounter{forecastlast}

\NewDocumentCommand\defineForecasts{>{\SplitList{\\}}m}
{
   \ProcessList{#1}{ \processForecastItems }
 }

\DeclareDocumentCommand \processForecastItems
  { > { \SplitArgument{5} {&} } m}
  {\defineForecast #1}

\newcommand{\FAbbr}[1]{\textbf{\textsc{\MakeLowercase{#1}}}}
\newcommand{\FName}[1]{{\sc #1}}

\newcommand{\DefFAbbr}[1]{\assign{F#1}{\FAbbr{#1}}}
\newcommand{\DefFName}[2]{\assign{F#1name}{\FName{#2}}}

\newcommand{\DefFLayer}[1]{\assign{Fl#1}{\thestagecounter}}
\newcommand{\DefFTableEntry}[2]{%
  \expandafter\newcommand\expandafter{\csname F#1TE\endcsname}{#2}}

\newcommand\defineForecast[6]{
  \xappto{\allforecastabbreviations}{#3,}
  \DefFAbbr{#3}
  \DefFName{#3}{#2}
  \refstepcounter{forecastcounter}
  \setcounter{forecaststart}{\theforecastlast}
  \refstepcounter{forecaststart}
  \addtocounter{forecastlast}{#1}
  \assign{Fnum\theforecastcounter}{#3}
  \assign{Fni#3}{\theforecaststart}
  \ifnum #1 > 1
     \assign{Fnj#3}{\theforecastlast}
     \assign{Fn#3}{\compoundname{Fni#3}-\compoundname{Fnj#3}}
     \assign{FnA#3}{\FAbbr{#3}$(i)$}
  \else
      \assign{Fn#3}{\compoundname{Fni#3}}
      \assign{FnA#3}{\FAbbr{#3}}
  \fi
  \DefFTableEntry{#3}{\compoundname{Fn#3} & \FName{#2} & \compoundname{FnA#3} & #4 & #5 & #6}
}

\newcommand{\FaName}[1]{\compoundname{F#1name}}
\newcommand{\FaNumber}[1]{\compoundname{Fn#1}}
\newcommand{\FaLayer}[1]{\compoundname{Fl#1}}

\newenvironment{F_table}[2]{
\gdef\tempcaption{#1} 
\gdef\templabel{#2} 
\begin{table}[tbh]
\centering
\small
  \begin{tabular}{ccccccc}
  \toprule
  \mysplit{Forecast \\ Number} & Name & Abbrev & Option & \accv & \termv  \\
  \midrule
}
{
  \bottomrule
  \end{tabular}
  \refstepcounter{table} \caption*{Table~\thetable}
  \label{\templabel}
\end{table}
}
\newcount{\mytemp}

\NewDocumentCommand\TableFA{>{\SplitList{,}}mm}
{
  \begin{F_table}{}{table:F#2}
    \ProcessList{#1}{ \processTableItemsF }
  \end{F_table}
  \ProcessList{#1}{ \makeLayerDefinitionsF }
}

\def\myfirst[#1,#2]{#1}
\newcommand{\TableF}[1]{\TableFA{#1}{\myfirst[#1,dummy]}}

\DeclareDocumentCommand \processTableItemsF
  {m}
  {\compoundname{F#1TE} \\}

\DeclareDocumentCommand \makeLayerDefinitionsF
  {m}
  {\DefFLayer{#1}}

\newcounter{optioncounter}

\NewDocumentCommand\defineOptions{>{\SplitList{\\}}m}
{
   \ProcessList{#1}{ \processOptionItems }
 }

\DeclareDocumentCommand \processOptionItems
  { > { \SplitArgument{6} {&} } m}
  {\defineOption #1}

\newcommand{\OAbbr}[1]{{\it{\MakeLowercase{#1}}}}
\newcommand{\OName}[1]{\emph{#1}}

\newcommand{\DefOAbbr}[1]{\assign{O#1}{\OAbbr{#1}}}
\newcommand{\DefOName}[2]{\assign{O#1name}{\OName{#2}}}
\newcommand{\DefONumber}[1]{\assign{On#1}{\theoptioncounter}}
\newcommand{\DefOLayer}[1]{\assign{Ol#1}{\thestagecounter}}

\newcommand{\DefOTableEntry}[2]{%
  \expandafter\newcommand\expandafter{\csname O#1TE\endcsname}{#2}
}
\newcommand{\DefOTableEntryCZ}[2]{%
  \expandafter\newcommand\expandafter{\csname O#1TECZ\endcsname}{#2}
}

\newcommand\defineOption[7]{
  \xappto{\alloptionabbreviations}{#2,}
  \DefOAbbr{#2}
  \DefOName{#2}{#1}
  \refstepcounter{optioncounter}
  \assign{Onum\theoptioncounter}{#2}
  \DefONumber{#2}
  \DefOTableEntry{#2}{\compoundname{On#2} & \OName{#1} & \OAbbr{#2} & #3 & #4 & #5}
  \DefOTableEntryCZ{#2}{\compoundname{On#2} & \OName{#1} & \OAbbr{#2} & #3 & #4 & #5 & #6 & #7}
}

\newcommand{\OaName}[1]{\compoundname{O#1name}}
\newcommand{\OaNumber}[1]{\compoundname{On#1}}
\newcommand{\OaLayer}[1]{\compoundname{Ol#1}}

\newenvironment{O_table}[2]{
\gdef\tempcaption{#1} 
\gdef\templabel{#2} 
\begin{table}[tbh]
\centering
\small
\begin{tabular}{cccccc}
  \toprule
  \mysplit{Option \\ Number} & Name & Abbrev & I & π & β \\
  \midrule
}
{
  \bottomrule
  \end{tabular}
  \refstepcounter{table} \caption*{Table~\thetable}
  \label{\templabel}
\end{table}
}

\NewDocumentCommand\TableOA{>{\SplitList{,}}mm}
{
  \begin{O_table}{}{table:O#2}
    \ProcessList{#1}{ \processTableItemsO }
  \end{O_table}
  \ProcessList{#1}{ \makeLayerDefinitionsO }
}

\newcommand{\TableO}[1]{\TableOA{#1}{\myfirst[#1,dummy]}}

\DeclareDocumentCommand \processTableItemsO
  {m}
  {\compoundname{O#1TE} \\}

\DeclareDocumentCommand \makeLayerDefinitionsO
  {m}
  {\DefOLayer{#1}}

\newenvironment{OCZ_table}[2]{
\gdef\tempcaption{#1} 
\gdef\templabel{#2} 
\begin{table}[tbh]
\centering
\small
\begin{tabular}{cccccccc}
  \toprule
  \mysplit{Option \\ Number} & Name & Abbrev & I & π & β & c & z \\
  \midrule
}
{
  \bottomrule
  \end{tabular}
  \refstepcounter{table} \caption*{Table~\thetable}
  \label{\templabel}
\end{table}
}

\NewDocumentCommand\TableOCZA{>{\SplitList{,}}mm}
{
  \begin{OCZ_table}{}{table:O#2}
    \ProcessList{#1}{ \processTableItemsOCZ }
  \end{OCZ_table}
  \ProcessList{#1}{ \makeLayerDefinitionsO }
}

\DeclareDocumentCommand \processTableItemsOCZ
  {m}
  {\compoundname{O#1TECZ} \\}

\newcommand{\TableOCZ}[1]{\TableOCZA{#1}{\myfirst[#1,dummy]}}

\newcounter{aliascounter}

\NewDocumentCommand\defineAliases{>{\SplitList{\\}}m}
{
   \ProcessList{#1}{ \processAliasItems }
 }

\DeclareDocumentCommand \processAliasItems
  { > { \SplitArgument{2} {&} } m}
  {\defineAlias #1}

\newcommand{\AAbbr}[1]{\emph{\MakeUppercase{#1}}}
\newcommand{\AName}[1]{\textbf{#1}}
\newcommand{\DefAAbbr}[1]{\assign{A#1}{\AAbbr{#1}}}
\newcommand{\DefAName}[2]{\assign{A#1name}{\AName{#2}}}
\newcommand{\DefANumber}[1]{\assign{An#1}{\thealiascounter}}
\newcommand{\DefALayer}[1]{\assign{Al#1}{\thestagecounter}}
\newcommand{\DefATableEntry}[2]{%
  \expandafter\newcommand\expandafter{\csname A#1TE\endcsname}{#2}
}

\newcommand\defineAlias[3]{
  \xappto{\allaliasabbreviations}{#2,}
  \DefAAbbr{#2}
  \DefAName{#2}{#1}
  \refstepcounter{aliascounter}
  \assign{Anum\thealiascounter}{#2}
  \DefANumber{#2}
  \DefATableEntry{#2}{\compoundname{An#2} & \AName{#1} & \AAbbr{#2} & #3}
}

\newcommand{\AaName}[1]{\compoundname{A#1name}}
\newcommand{\AaNumber}[1]{\compoundname{An#1}}
\newcommand{\AaLayer}[1]{\compoundname{Al#1}}

\newenvironment{A_table}[2]{
  \gdef\tempcaption{#1} 
  \gdef\templabel{#2} 
  \begin{table}[tbh]
    \centering
    \small
    \begin{tabular}{cccc}
      \toprule
      \mysplit{Alias \\ Number} & Name & Abbrev & Definition \\
      \midrule
    }
    {
      \bottomrule
    \end{tabular}
    \refstepcounter{table} \caption*{Table~\thetable}
    \label{\templabel}
  \end{table}
}

\NewDocumentCommand\TableAA{>{\SplitList{,}}mm}
{
  \begin{A_table}{}{table:A#2}
    \ProcessList{#1}{ \processTableItemsA }
  \end{A_table}
  \ProcessList{#1}{ \makeLayerDefinitionsA }
}

\newcommand{\TableA}[1]{\TableAA{#1}{\myfirst[#1,dummy]}
}

\DeclareDocumentCommand \processTableItemsA
  {m}
  {\compoundname{A#1TE} \\}

\DeclareDocumentCommand \makeLayerDefinitionsA
  {m}
  {\DefALayer{#1}}


%% file: forecasts.tex
\defineForecasts{
1            & Touch                      & T      & \Oef                                                                 & 0 & \ST                                                                            \\
1            & TouchLeft                  & TL     & \Orotl                                                               & 0 & \FT                                                                            \\
1            & TouchRight                 & TR     & \Orotr                                                               & 0 & \FT                                                                            \\
11           & Touchmap                   & TM     & $\case{\Orotr, if $i \in \{1...6\}$; \Orotl, if $i \in \{7...11\}$}$ & 0 & $\case{\FTM(i-1), if $i \in \{1...6\}$; \FTM(i+1), if $i \in \{7...11\}$}$     \\
1            & Touch Adjacent             & TA     & \Ortt                                                                & 0 & \FT                                                                            \\
1            & Distance to Touch Adjacent & DTA    & \Orfta\ (or \Orftt)                                                  & 1 & 0                                                                              \\
1            & Nearness to Touch Adjacent & NTA    & \Orfta\ (or \Orftt)                                                  & 0 & $\case{1, if $\FTA>θ$; 0, otherwise}$                                          \\
11           & Distance-to-\FTA\ Map      & DTAM   & $\case{\Orotr, if $i \in \{1...6\}$; \Orotl, if $i \in \{7...11\}$}$ & 0 & $\case{\FDTAM(i-1), if $i \in \{1...6\}$; \FDTAM(i+1), if $i \in \{7...11\}$}$ \\
1            & Wall Right, forward        & WRf    & \OrfWR                                                               & 1 & 0                                                                              \\
1            & Wall Left, forward         & WLf    & \OrfWL                                                               & 1 & 0                                                                              \\
1            & Wall Right, backward       & WRb    & \OrbWR                                                               & 1 & 0                                                                              \\
1            & Wall Left, backward        & WLb    & \OrbWL                                                               & 1 & 0                                                                              \\
1            & Wall Distance A            & WDa    & \OmrW                                                                & 1 & 0                                                                              \\
1            & Wall Distance B            & WDb    & \OmCWp                                                               & 1 & 0                                                                              \\
1            & Distance to Wall           & DW     & \OrfW                                                                & 1 & 0                                                                              \\
11           & Distance-to-Wall Map       & DWM    & $\case{\Orotr, if $i \in \{1...6\}$; \Orotl, if $i \in \{7...11\}$}$ & 0 & $\case{\FDWM(i-1), if $i \in \{1...6\}$; \FDWM(i+1), if $i \in \{7...11\}$}$   \\
1            & Distance to Room           & DR     & \Ogfr                                                                & 1 & 0
}
%
%

%% file: options.tex

\newcommand{\comma}{,}
\defineOptions{
roll forward                                             & rf     & S                                 & -        & 1                                                                                                   &                                                                          &       \\
roll backward                                            & rb     & S                                 & -        & 1                                                                                                   &                                                                          &       \\
rotate left                                              & rotl   & S                                 & -        & 1                                                                                                   &                                                                          &       \\
rotate right                                             & rotr   & S                                 & -        & 1                                                                                                   &                                                                          &       \\
extend finger                                            & ef     & S                                 & -        & 1                                                                                                   &                                                                          &       \\
rotate to touch                                          & rtt    & S                                 & maximize & $\case{1, if \FT; 0.1, otherwise}$                                                                  & $\case{0, if $\action_t \in \{ \Orotl \comma \Orotr \comma \Oef\}$; -\infty, otherwise}$ & \FT   \\
roll forward toward touch adjacent                       & rfta   & S                                 & \Orf     & \FTA                                                                                                &                                                                          &       \\
roll forward until touch threshold                        & rftt   & S                                 & \Orf     & $\case{1, if $\FTA > \theta$; 0, otherwise}$                                                        &                                                                          &       \\
\mysplit{roll forward along \\ wall on right}            & rfWR   & $\theta_1 < \FDTAM(3) < \theta_2$ & \Orf     & $\case{1, if $(\FTA > \theta_3)$ or \emph{Not} $(\theta_1 < \FDTAM(3) <\theta_2)$; 0.1, otherwise}$ &                                                                          &       \\
\mysplit{roll forward along \\ wall on left}             & rfWL   & $\theta_1 < \FDTAM(9) < \theta_2$ & \Orf     & $\case{1, if $(\FTA > \theta_3)$ or \emph{Not} $(\theta_1 < \FDTAM(9) <\theta_2)$; 0.1, otherwise}$ &                                                                          &       \\
\mysplit{roll backward along \\ wall on right}           & rbWR   & $\theta_1 < \FDTAM(3) < \theta_2$ & \Orb     & $\case{1, if $(\FTA > \theta_3)$ or \emph{Not} $(\theta_1 < \FDTAM(3) <\theta_2)$; 0.1, otherwise}$ &                                                                          &       \\
\mysplit{roll backward along \\ wall on left}            & rbWL   & $\theta_1 < \FDTAM(9) < \theta_2$ & \Orb     & $\case{1, if $(\FTA > \theta_3)$ or \emph{Not} $(\theta_1 < \FDTAM(9) <\theta_2)$; 0.1, otherwise}$ &                                                                          &       \\
\mysplit{move randomly until \\ wall left or right}  & mrW    & S                                 & random   & $\case{1, if \AWLR; 0.1, otherwise}$                                                                &                                                                          &       \\
move to canonical wall position                          & mCWp   & S                                 & maximize & $\case{1, if \AWLR; 0.1, otherwise}$                                                                & 0                                                                        & \AWLR \\
roll forward to wall                                     & rfW    & S                                 & \Orf     & $\case{1, if \AWA; 0.1, otherwise}$                                                                 &                                                                          &       \\
go to middle of hallway                                  & gmh    & S                                 & maximize & $\case{1, if \AMH; 0.1, otherwise}$                                                                 & 0                                                                        & \AMH  \\
go to center of room                                     & gcr    & S                                 & maximize & $\case{1, if \ACR; 0.1, otherwise}$                                                                 & 0                                                                        & \ACR  \\
go forward into room                                     & gfr    & S                                 & \Orf     & \AR & & 
}
%


%% file: aliases.tex
\defineAliases{
Wall Left or Right          & WLR          & (\FWRf\ > θ) or (\FWLf\ > θ) or (\FWRb\ > θ) or (\FWLb > θ) \\
Wall Adjacent               & WA           & $(\FTA\ > \theta_1)$ and $(\FWDa < \theta_2)$               \\
Left-Right Free Space       & LRFS         & $\FDWM(3) + \FDWM(9)$                                       \\
Front-back Free Space       & FBFS         & $\FDWM(0) + \FDWM(6)$                                       \\
Left-right Centered         & LRC          & $\FDWM(3) ≈ \FDWM(9)$                                       \\
Front-Back Centered         & FBC          & $\FDWM(0) ≈ \FDWM(6)$                                       \\
Room                        & R            & $\FDWM(i) < θ_1, i ∈ \{0,3,6,9\}$                           \\
Centered in a Room          & CR           & \AR\ and \ALRC\ and \AFBC                                   \\
Middle of Hall              & MH           & \ACR\ and $\ALRFS < θ_3$ and $\AFBFS > θ_4$                 \\
Room Area                   & RA           & $\ALRFS × \AFBFS$                                           \\
Small Room                  & SR           & \ACR\ and $(\ARA < θ_5)$                                    \\
Large Room                  & LR           & \ACR\ and $(θ_6 < \ARA < θ_7)$                              \\
Doorway                     & D            & $\FDTA(3) < θ_1$ and $\FDTA(9) < θ_1$ and $DR < θ_2$ 
}                


%% file: abstract.tex
\begin{abstract}
%

This paper shows how a single mechanism allows knowledge to be constructed layer by layer directly from an agent's raw sensorimotor stream.
This mechanism, the General Value Function (GVF) or ``forecast,'' captures high-level, abstract knowledge as a set of predictions about existing features and knowledge, based exclusively on the agent's low-level senses and actions.

Thus, forecasts provide a representation for organizing raw sensorimotor data into useful abstractions over an unlimited number of layers—a long-sought goal of AI and cognitive science.

The heart of this paper is a detailed thought experiment providing a concrete, step-by-step formal illustration of how an artificial agent can build true, useful, abstract knowledge from its raw sensorimotor experience alone.
The knowledge is represented as a set of layered predictions (forecasts) about the agent's observed consequences of its actions.
This illustration shows twelve separate layers: the lowest consisting of raw pixels, touch and force sensors, and a small number of actions; the higher layers increasing in abstraction, eventually resulting in rich knowledge about the agent's world, corresponding roughly to doorways, walls, rooms, and floor plans.
\IA then argue that this general mechanism may allow the representation of a broad spectrum of everyday human knowledge.
\end{abstract}


%% file: motivation.tex
\section{Motivation}
\label{sec:overview}

One of the great problems in AI and Cognitive Science is that of connecting a representation to what it represents.
Despite decades of research developing ontologies and datasets for encoding information that most children take for granted, no representation has yet emerged capable of capturing the richness of a child's understanding.
As humans, \weH draw on vast resources of intuition to justify \ourH judgments.
\WeH do not merely know that pillows are soft and that basketballs bounce, we know what they look and feel like; we know what to expect when we take hold of a balloon or a drinking glass, when we sit down on a tricycle or when we jump into a pile of leaves.
We know their sizes, shapes and textures; we can guess how hard or heavy they are or how likely they might be to blow away in a strong wind.
And \weH know these things without having been told them explicitly; \weH have learned them through \ourH everyday experiences.
Yet this knowledge is represented in \usH\ in an accessible way: \weH can assess its validity quickly, and \weH can use it for deeper reasoning.
Such a powerful representation has eluded the best efforts of AI researchers, but \IA believe that recent research has made substantial progress towards reproducing this power.

\subsection{Why a New Knowledge Representation?}
\label{sec:why-new-repr}

Reasoning systems in AI rely on information encoded by humans and represented by humans in forms that humans think are useful for reasoning.
Adult humans, especially adult human AI researchers, tend to consider \emph{knowledge} to consist of symbols bound together by a network of relationships, while \emph{reasoning} is the application of rules of inference to these symbols and relationships.
Yet the symbols themselves tend not to be related to the universal basis of all human knowledge: everyday sensorimotor interaction with the world.

The CYC project---the world's largest relational knowledge base---recently solicited help online to answer queries it could not resolve within the knowledge base itself.
The website visitor was asked to confirm, if possible, whether a given statement is true.
In some cases, CYC requested confirmation of objective facts, such as:
\begin{my_list}
\item Bauxite ore is a natural resource of Greece.
\item Aramaic is spoken in Syria.
\item Frank Nelson appeared on ``I Love Lucy''.
\end{my_list}
In these cases, CYC lacked some specific bits of world knowledge that the visitor was asked to supply.
But in many other cases, a different kind of confirmation was requested:
\begin{my_list}
\item Most balloons are taller than most cages.
\item Most canes are heavier than most bed pillows.
\item Most hominids are heavier than most lions.%
\end{my_list}
So what was CYC lacking in these cases?
Did it simply require more knowledge about pillows, balloons, and lions, or a better definition of heaviness or height?
What does the human have that the knowledge-based reasoning system did not?
Though perhaps hard to define exactly, there is an obvious intuitive answer here: what CYC lacks is experience with the real world.\footnote{By \emph{experience} I am referring here and throughout this paper to sensorimotor interaction with the world, and specifically to the low-level interactive data stream: raw sensory input and motor output.}

Humans are indispensable to CYC's interaction with the world.
Humans are not just necessary for encoding CYC's knowledge, but also for interpreting its symbols.
This human ability that knowledge bases lack is, perhaps, in some sense, the real story of intelligence.
\WeH know approximately how heavy canes are, as well as hominids, lions, and bed pillows---despite not having placed each of them on a scale or having read anywhere what they should weigh.
\WeH can imagine interacting with these objects, and this allows \usH\ to answer an unlimited number of questions about their properties and relationships.
\WeH can consider how heavy, or how soft, or how comfortable, a pillow is by imagining what happens when \weH manipulate it.
\WeH can guess how heavy a lion might be just by imagining the act of trying to lift one up.
And these acts of imagination ultimately require a representation that allows us to predict the possible perceptual outcomes of the interactions.

Researchers studying general intelligence and developmental learning largely appreciate that high-level knowledge must ultimately be connected to and formed from low-level sensorimotor data, yet the chasm between raw data and real knowledge has seemed too immense to bridge.
No toolset, language, or framework has yet proven sufficient for constructing this connection, and the search is still on for a robust mechanism that allows useful abstractions to be built up from the sensorimotor stream in an unlimited number of layers, though many powerful candidates have been proposed reaching back many decades.

Jean Piaget proposed that children \emph{construct} their understanding of reality as they grow~\cite{Piaget:constructivism}.
In support of this \emph{constructivist} theory of knowledge, Piaget enlisted the \emph{schema} framework of Kant as a method of representation that could account for the development of intelligence across multiple stages, from low-level sensorimotor interactions to abstract knowledge.
Piaget's schemas later inspired computational models, such as the simulated agent of Drescher~\cite{Drescher}, which started from primitive senses and actions and constructed a rudimentary understanding of object permanence.
Earlier, in the 1970's, Cunningham~\cite{Cunningham} had attempted to combine Piaget's relatively high-level schemas with Hebb's low-level ``cell assemblies''~\cite{Hebb:1949} (a computation-based constructivist proposal more precisely defined than schemas) to produce a hybrid model of intelligence.
Cunningham was notably deliberate in his search for ``a small basic set of data structures'' from which ``all the complex structures and operations eventually recognized as being intelligent'' could develop.
The JCM system of Becker~\cite{Becker} and (though less explicitly) the classifier system of Holland~\cite{Holland} were also originally conceived with similar constructivist goals in mind.

In the 1980's the neural-network resurgence returned the sub-symbolic representational perspective to the forefront of AI, and at roughly the same time, reinforcement learning also experienced renewed interest, as researchers strove to increase the autonomy of artificial agents by allowing them to learn about uninterpreted sensorimotor signals from positive and negative rewards.
In addition, handcrafted agents that learned about the world in explicit stages of development~\cite{Kuipers:critter,KuipersByun:AAAI88,PierceKuipers:AIJ} helped to sway the community toward an agent-centric viewpoint.
Each of these in its way was attacking some part of the constructivist problem, seeking agents that autonomously build up an understanding of the world.

%
``Continual learning'' and ``continual development'' were the terms first applied to open-ended, hierarchical, constructivist learning set within a modern connectionist and reinforcement-learning framework~\cite{Ring:ML91,Ring:SAB92,Ring:Thesis,Ring:ML}.\footnote{The term ``continual learning'' will be used throughout this paper to refer to the full learning process of the constructivist agent, as it concretely and succinctly captures the following critical attributes: continual and unlimited growth through interactive, autonomous, online, and incremental learning from positive and negative rewards using a single learning mechanism at all layers.
See Ring (1994) or Ring (1997) for a precise description and detailed discussion of this term.}
The resulting agent, CHILD~\cite{Ring:Thesis}, was designed explicitly for continual, hierarchical, incremental learning and development; it started from raw sensorimotor interaction and autonomously built up context-sensitive skills.
This and other work from the 1990's on learning to learn~\cite{Schmidhuber:L2LLS} led eventually toward theories of optimality in reinforcement-learning agents that begin from sensorimotor interaction alone~\cite{Hutter:book}.
Current trends in AI also include methods for agents to learn probabilistic models of their environments through interaction~\cite{Ghahramani:BayesianReview}, which is often combined with reinforcement learning methods~\cite{Ghavamzadeh:BayesianRL}.
In addition, at least one annual conference (the International Conference on Developmental Learning) is devoted specifically to the problem of modeling cognitive development, and at least one large EU-funded project was recently mandated to contribute toward constructivist goals~\cite{IM-Clever}.
Other cousins of the current paper include Predictive State Representations (PSRs)~\cite{LittmanSuttonSingh:PSRs}, TD Networks~\cite{SuttonTanner:TDNets,Rafols,SuttonRafolsKoop,TannerKoopRafolsBulitko} and, most directly, Horde~\cite{Sutton:Horde}.
(More will be said about these later).
Thus, the constructivist movement has a long history in AI and Cognitive Science, and though its goal is tremendously ambitious, much progress has already been made.

Yet despite these advances, one piece of the puzzle has remained elusive: finding a single toolset with which new skills and knowledge can be built from existing skills and knowledge through unlimited levels of meaningful abstractions.
While Drescher's schema mechanism succeeded at creating a few clear levels of abstraction, it was less clear that the mechanism was general enough to support many such levels.
In contrast, CHILD was able to span an arbitrary number of levels, but it was not clear that these levels were general enough to represent a broad class of abstract knowledge.

In this article, \IA argue that General Value Functions (GVFs)---which \I nickname, ``forecasts''~\cite{Sutton:Horde,GQ,SuttonRafolsKoop,ModayilWhiteSutton:2012,SchaulRing:IJCAI2013}---may be the right representation to allow the constructivist program to proceed.
Forecasts provide a framework for building abstract knowledge in layers,\footnote{By ``layers'' \IA mean only that more abstract, higher-level features and knowledge may depend on less abstract, lower-level features and knowledge.
For example, if \I know that pillows are soft, then \I must also have some knowledge of pillows and of softness.
Thus, it is the knowledge that is layered, and the forecasts are able to capture and reflect these dependencies.
  } organizing the world the way a baby (or any beginning creature) might: by making predictions about the outcomes of behavior---posing and answering questions of the form, ``if \I behave \emph{this} way, will \I perceive \emph{that}?''
For example, ``if \I pick up this pillow, will it feel soft?''

This paper attempts to explain how forecasts work and how they can be used to encode knowledge predictively.
%
%
I believe that before machines can reason about the world in the way that humans can, they must first be able to represent knowledge in terms of their sensorimotor stream, as humans do.
Thus, the paper's primary contribution is a demonstration of how an agent using forecasts can build up increasing layers of abstract knowledge directly from the sensorimotor stream: from extremely limited sensors and actions, the agent constructs a complex awareness of its surroundings and an understanding of how it can interact with its world.

Forecasts provide an important piece of the constructivist puzzle, but many pieces remain to be found.
\MyA overall goal here is to show how knowledge can be built from predictions—--in sufficient detail that it motivates \our community to start filling in the remaining gaps.

\subsection{Building from the Bottom Up}
\label{sec:bloomingbaby}

A baby comes into the world knowing very little but having the latent ability to learn all the facts, skills, places and things it will come to know over the course of its life.
William James famously wrote that ``the baby, assailed by eyes, ears, nose, skin, and entrails at once, feels it all as one great blooming, buzzing confusion''~\cite{James:bloomingbuzzing}.
Yet starting from within this confusion, the baby begins to build up knowledge and skills, piece by piece, eventually developing a rich understanding of itself and its environment.
But how?

Though perhaps lost in its initial confusion, the baby soon begins to notice regularities, certain relationships between its actions and its perceptions.
It cries and it gets picked up; it drinks and its stomach feels full; it moves its eyes to the left, and what was on the right side of its retina moves to the center; it puts its hands together and perceives a constellation of somatosensory signals.
As these experiences are repeated, they become more predictable, and the baby comes to expect the perceptions that result from its actions.
By organizing and refining these expectations, the baby can begin to understand its world.

\IA believe that this predictive way of representing the world can potentially bridge all stages of growth, remaining as valuable in the adult's world as in the baby's.
Furthermore, with the right representational mechanisms, an artificial, continual-learning agent might build up an understanding of its world in the same way: simply by continually extending the predictions it makes about the consequences of its actions—at first learning to predict frequent and coarse regularities,
then gradually expanding its understanding of its world, step by step, as its predictions become more refined, complex, and larger in scope.

\subsection{Isolaminar, Continual Learning}
%
As \weH humans develop, learn, and build up layers of abstractions, \ourH experience is \ourH only link to the world.
Everything \weH understand, \weH have learned through interaction with the environment.
All \ourH knowledge derives from this small stream of sensorimotor data.

If an artificial agent is to gain human-like knowledge, it  will need to build up and refine that knowledge through its experience.
As it interacts with the world, it will learn continually---constantly developing, constantly revising what it knows, constantly building on top of what it has already learned, using what it knows now as the basis for what it learns next.
This continual-learning process is critical to the development of knowledge from sensorimotor activity.

\commentout{need in here something about the constancy of the learning/tuning/verification process ? This becomes important in section~\ref{sec:Framework}.}

Yet continual learning is not the standard approach pursued in AI (though it is increasingly common in AI for an agent's learned models to be \emph{refined} through experience).
An important branch of current research, for example, focuses on the use of probabilistic methods to refine abstract models of the world using sensorimotor data~\cite{Dayan:Switches,Ghavamzadeh:BayesianRL,KordingWolpert:BayesianSMLearning,Kording:Review,Wolpert:MotorLearning}.
This is essential work and addresses a particular part of the learning process.

The subject of this article, however, is different.
Its focus is on knowledge \emph{construction}, and in particular on \emph{isolaminar} knowledge construction.
By ``isolaminar,'' \IA refer to any constructive method that uses the same mechanism for all layers.
Thus, a brick wall is isolaminar; a suspension bridge is not.
A deep neural network is also isolaminar: it uses the same mechanism, repeated in any number of layers.
Most current methods for learning in complex environments are not isolaminar.
Forecasts are isolaminar.



To build knowledge from the bottom up means building at multiple levels (and probably building at many such levels simultaneously).
To build at multiple levels requires either inventing new methods at each level to construct that level from the previous one, or using an isolaminar method that will work at every level equally well.
Finding such a single method is not easy, because it must be just as suitable for building knowledge from low-level sensorimotor data as from high-level abstractions; it must be equally suited to all kinds of experience and all kinds of knowledge, from gustatory and kinesthetic knowledge to musical and mathematical knowledge.

Forecasts are a representation that allows sensorimotor data to be knitted together into ever more abstract sensorimotor constructs, every layer having the same form and behaving according to the same principles as every other. 



\subsection{Knowledge and generalization}
\label{sec:generalization}


Before describing forecasts in detail, it is worth considering their desiderata.
A good representation for continual learning would:

\renewcommand{\labelenumi}{\alph{enumi})}
\begin{my_enumerate}
\label{sec:good-reps}
\item describe knowledge in terms of senses and actions;
\item allow ongoing, incremental learning from experience;
\item support the isolaminar development of knowledge in ever-increasing layers of abstraction;
\item afford access to the knowledge for planning and reasoning.%
\footnote{%
Noticeably absent from this list are qualities such as amenability to symbolic interpretation, suitability for logical manipulation, readiness for integration into existing software or databases, etc.
While these qualities are desirable in a software knowledge base designed principally for human access, they are not immediate goals of the constructivist paradigm as they are not in themselves essential to the agent's continually increasing ability to understand and navigate the complexities of its world.
(And bearing in mind that we humans must amass knowledge and experience for years before we are capable of explicit logical manipulation of symbols, it is also not an unreasonable long-term goal for continual-learning agents to eventually develop an ability to use their knowledge for symbolic reasoning.
In fact, the knowledge constructed in the demonstration of Section~\ref{sec:demo} allows human symbolic interpretation, though it is not the focus of the current article.)
However, the agent's own immediate interests do demand a system of knowledge that supports the planning of actions and reasoning about their possible outcomes, and such concerns will be addressed in future work.}

\end{my_enumerate}
In turn, the represented knowledge should:
\begin{my_enumerate}
  \item be constructed through interaction with the world;
  \item be modifiable and extensible with new experience;
  \item be verifiable through interaction with the world;
  \item be useful to the intelligent agent.
\end{my_enumerate}

\commentout{\fncomment{i might also modify the focus here and describe to a greater extent why real Intelligence is about self interest, and that self interest reflects value.
Introducing it here might allow a deeper discussion about why certain representations emerge.
Currently, I plan on returning to this later, but maybe it should go here instead.}}

The last of these is perhaps the quality most critical for an agent pursuing its own desires and goals.
\WeH humans are agents in a universe that is unfathomably complex and unfolds according to a vast collection of predictable regularities, of which no human can understand and represent more than the tiniest fraction.
To make choices based on what is likely best for \usH, \weH are forced to evaluate the possible futures these choices entail, and \weH must do so using a considerably incomplete representation of \our world.
Yet the approximation \weH have of \ourH environment is quite useful to \usH, even if it is inaccurate and vastly incomplete.
The regularities it captures are sufficient for \usH\ to make successful and useful predictions.
In other words, \emph{\ourH representation generalizes well}.

A representation that generalizes well captures useful regularities of the environment in a way that allows accurate prediction of the consequences of actions, despite incomplete information. It is \myA contention that this is also a good definition of knowledge.

\subsection{Predictions as knowledge}

As will be described in detail below, forecasts are predictions.
How can predictions capture knowledge?
Though it may seem a strange assertion at first, perhaps knowledge is nothing \emph{other} than prediction.
Consider the piece of knowledge, ``my keys are in my pocket.''
This statement specifies a fact about the world that can be verified: \my keys are in \my pocket if \I can put \my hand in \my pocket and find them there.
The procedure for verifying the statement can be phrased as a prediction: ``if I put my hand in my pocket, I will detect my keys.''
\I can also estimate the probability that this prediction will come true—the probability of detecting \my keys if \I put \my hand in \my pocket.
This probability estimate will be different at different times in different situations.
If \I estimate the probability to be high, \I can convey that by saying, ``my keys are in my pocket.''
From a predictive viewpoint, this everyday piece of knowledge \emph{is the same as} \my prediction that \I will feel the keys in my pocket if \I search for them there.%
%
%
%

It is the claim of this paper that the above example is neither unique nor rare but in fact exemplifies the way that \emph{all} knowledge works.
This is a very strong statement and might best be expressed as a hypothesis; let us call it the \emph{predictive knowledge hypothesis}: All knowledge is private and can be reduced to a set of predictions about one's sensorimotor stream.
According to this view, a statement such as, ``the cat is on the mat,'' fundamentally means something subjective and predictive: ``I estimate that if I execute a certain set of actions, then I will make a certain set of observations.''
Thus, statements a person makes about a conveniently hypothesized objective world have meaning to the extent that they reduce to subjective predictions about the speaker's (and listeners') sensorimotor streams.
These are not outright predictions about \emph{the} future---what \emph{will} happen; they are instead predictions about \emph{possible} futures---what \emph{could} happen.
What distinguishes the possible futures from the actual future is one's own implied involvement in the prediction, one's own interaction with the environment, one's own sensorimotor stream.\footnote{The concept of action-conditional observation as the basis for knowledge has roots at least back to the philosophy of Pragmatism.
William James, for example, asserted that all useful distinctions must have practical consequences: thus, to distinguish two things, one must be able to interact with them through the same series of actions and arrive at two different observations~\cite{James:squirrel}.}

The predictive knowledge hypothesis stands in contrast to the physical symbol systems hypothesis~\cite{NewellSimon:PSSH}, which postulates the necessity and sufficiency of symbol manipulation for producing intelligence.
Whereas the latter asserts that intelligence requires symbol manipulation, the former asserts that the interpretation of symbols and their referents is always subjective---constructions of predictions about one's own sensorimotor stream.
\footnote{More specifically, the ability to interpret symbols—to understand their references, and to relate them to the world—requires an enormous network of connected knowledge, and thus the ability to construct abstractions of great sophistication.
Only in a tautological sense is even the ``necessary'' claim of the physical symbol system hypothesis compatible with the predictive knowledge hypothesis; i.e., only if we were to define the essential characteristic of intelligence to be the ability to manipulate symbols.
But such a definition would be arbitrary and would ignore the vast foundation of (predictive) knowledge required for symbols to be understood.
} When \weH say, for example, that a particular wall is red, or that pillows are not heavy, or that most people like the taste of ice cream, what are \weH saying?
In each case, the sentence expresses a prediction about a possible future; each sentence puts into words what \weH predict \weH would probably \emph{perceive} in a certain future scenario.
If \weH look at the wall, \weH will perceive red.
If \weH try to lift something \weH believe to be a pillow, \weH will almost certainly succeed and perceive very little resistance.
If \weH give a person ice cream, there is a good chance \weH will be able to perceive an improvement in mood in the recipient (a very complex set of predictions).
Of course, all of these examples entail some rather sophisticated sub-components of their own: how do \weH determine something to be a wall, a pillow, or a person?
Yet if the hypothesis is correct, then these sub-components must also refer to pieces of predictive knowledge, and, furthermore, there will be no circularity; instead, all statements of knowledge are eventually reducible to predictions about raw experience, i.e., to predictions about interactions between senses and actions.


\subsection{Forecasts as Predictive Knowledge}
%
\label{sec:predictingwithforecasts} 

Predictive models are common in AI, but these are typically too precise to be generally useful.
Predictive models in robotics and reinforcement learning nearly universally make single-step predictions; i.e., they predict what the agent's next observation will be if the agent takes a specific action in a specific state.
Similarly predictive state representations (PSRs) have been in the literature now for over a decade, but (nearly) all previous work has dealt exclusively with single-step predictions. %
In principle, single-step predictions can be quite useful and can provide the foundation for making more complex predictions.
But they are not the kinds of predictions needed for representing abstract, human-level knowledge.

In normal life, \weH humans need to predict many different kinds of events and quantities within a rather non-specific time frame.
\IA may need to know, for example, whether the door will open if \I turn the handle, or what the chance of crashing into another car is if \I pull out onto the street right now, or how likely the ball is to go into the hoop if \I throw it from over here, or how full \my stomach will be if \I eat the entire piece of cake.
In none of these cases, and in very few cases throughout life, is it critical to know the precise time step at which anything will occur, or to know anything about what will happen exactly one time step from now.
In general, prediction at fine temporal granularity is something that is preferred in AI for its technical convenience rather than for its usefulness to the agent.

In common parlance, the word ``prediction,'' is very broad and can mean many things.
A good thesaurus lists a dozen or more synonyms, each slightly different.
In AI, ``prediction'' generally refers specifically to one-step predictions.
The word ``forecast'' has fewer established associations in AI and has a more nuanced traditional meaning outside of AI indicating a specific kind of prediction---generally a probability estimate about a specific future of interest---and for this reason it seems an appropriate description for the kind of predictions that I claim underly all knowledge.
I therefore use ``forecast'' as a convenient nickname for the long-term, action-conditional predictions embodied by GVFs (general value functions), which I will describe formally in Section~\ref{sec:framework}.

A \emph{forecast} is not a one-step prediction and does not (generally) estimate the value of a state variable or feature at the next time step.
Instead, it is an estimate of a measurable quantity over the course of a possible future. 
But unlike the traditional use of the word, which refers to statements such as ``the chance of sunshine on Thursday'' or ``how much snow will fall in January,'' the forecasts in this article are explicitly conditioned on the agent's activity, and thus  resemble statements such as, ``how cold it will it be in the city I am visiting next week,'' or, ``how wet my clothes will become if I run to the market in this rain storm.''
Thus, forecasts are temporally extended predictions that depend on the agent's actions and can involve probabilities and cumulative quantities, but they always represent estimates of scalar values.

Just as one can make any number of predictions about the weather, an agent can make an unlimited number of forecasts (though each forecast must adhere to strict formal guidelines, as will be presented in Section~\ref{sec:forecast-details}).
Because of this variety, and because forecasts can make predictions about other forecasts, their expressiveness is profound and can cover a broad array of knowledge, much more easily, cleanly, and transparently than previous isolaminar mechanisms such as those of Drescher, CHILD, or, PSRs.
Thus forecasts represent a particularly powerful kind of predictive representation, and the demonstration of Section~\ref{sec:demo} tests and explores the predictive representation hypothesis by attempting to build world knowledge using forecasts.
The knowledge created there is abstract enough to lend some credibility to the hypothesis, and shows that forecasts are capable of representing at least some of the most important kinds of knowledge we want our robots to have.
And it is not inconceivable that the same principles could eventually allow an agent to represent all such knowledge, including such abstract forms as mathematical and historical facts and such abstract entities as ancient Rome, Bauxite Ore and \emph{I Love Lucy}.
But it is only a first step---an initial thought experiment---and we must keep two things in mind: first, that even humans require many years of continual learning to construct the foundation necessary for the comprehension of such abstractions; and second, that forecasts may not be the ultimate predictive representation, and that refinements and improvements may reveal themselves as real artificial learning agents begin to build up to such vast collections of actual, predictive knowledge.

\subsection{Overview of Paper}
The next section (Section~\ref{sec:framework}) provides a precise specification of forecasts, describing in formal terms how they can represent general predictive knowledge as described above.
%
The following section  (Section~\ref{sec:demo}) constitutes the central contribution of the paper: a thought experiment in which forecasts are combined in an isolaminar fashion to construct everyday abstract knowledge directly from the sensorimotor stream.
The final section briefly discusses the high-level ramifications of the thought experiment.

%


%% file: framework.tex
\section{An Isolaminar Framework for Predictive Knowledge}
\label{sec:framework}

How can an agent represent knowledge as a set of predictions?
This paper is an attempt to answer that question, and the technical details are shown below, but the answer is---conceptually, at least---fairly simple; the agent:  (1) makes predictions that are contingent on currently known behaviors; (2) learns new behaviors that maximize or minimize currently known predictions; and (3) repeats.
The initial predictions and behaviors are very simple and seem to have little to do with knowledge, but as layering increases, so does the degree of abstraction, and the predictions start to resemble---and then become---true high-level knowledge.
(This story, from simple forecasts to high-level knowledge, will be played out in great detail in Section~\ref{sec:demo}.)
The next three paragraphs provide a slightly expanded version of the above description.

First, the agent limits all its predictions (and thus all its knowledge) to the one thing it can measure and verify: its own sensorimotor stream.
It does this by creating and maintaining a set of forecasts.
%
Each forecast estimates a specific scalar quantity computable directly from the agent's future sensorimotor stream, and the value of this quantity depends on the agent's way of behaving.
To the extent that the agent behaves in that specified way, the forecast can be verified: the predicted value can be compared against the actual value.
The agent may have a repertoire of many ways of behaving, and it may be interested in predicting many different quantities from its future sensorimotor stream.

Second, the agent can create new behaviors that learn to maximize (or minimize) any of the agent's existing forecast estimates, and these new behaviors can become part of the agent's repertoire.

Third, the agent can create new forecasts to predict the future value of existing forecasts, contingent upon any of the agent's known behaviors, and in this way the forecasts can be layered.
A newly initialized agent's first forecasts will predict only the immediate value of specific sensors and will be contingent upon specific primitive actions.
Then, new forecasts can be built that make predictions about the future value of existing forecasts.
The agent thereby builds up knowledge in layers, continually creating new forecasts that are contingent on existing behaviors, and continually learning new behaviors that maximize or minimize any of its existing forecast estimates.
Thus, new behaviors and new forecasts are built up in an isolaminar fashion and through this process eventually result in long-term predictions about very complex relationships.
These predictions, all encoded as forecasts, can seem remarkably similar to what is typically called ``knowledge.''

%

Thus, forecasts encode behavior-contingent predictions in a form that allows isolaminar knowledge to be learned, tuned, and verified continually.%
\footnote{This paper does \emph{not} answer the very important question implied by this process: how can the agent decide which quantities and behaviors are of interest and which quantities should be optimized?
This is still an open question.
The purpose here is only to examine whether forecasts can capture abstract, isolaminar knowledge, fully connected to the sensorimotor data.
In the demonstration section of this article, \IA will therefore choose all forecasts by hand.}
Section~\ref{sec:demo} will demonstrate this process in great detail.

\subsection{Forecasts}
\label{sec:forecast-details}
\commentout{
Forecasts can be succinctly and formally described within the framework of Markov decision processes (MDPs).
Forecasts are general value functions (GVFs)~\cite{Sutton:Horde,GQ,SuttonRafolsKoop}.
Like standard value functions from reinforcement learning~\cite{SuttonBarto:book}, they can be succinctly and formally described within the framework of Markov decision processes (MDPs).
An MDP consists of a set of states ($\state\in \allStates$), actions ($\action \in \allActions$), observations ($\obs\in\allObs$) and rewards ($\reward\in\allRewards$).
At every time step, the agent receives an observation $\obs_t$ and reward $\reward_t$ in its current state $\state_t$ and takes action $\action_t$ which leads the agent to the next state $\state_{t+1}$ and reward $\reward_{t+1}$ according to state transition probabilities $\transitions(\state,\action,\state') = \prob(\state_{t+1} = \state' \mid \state_t = \state, \action_t = \action)$ and reward probabilities $\Reward(\reward_{t+1} = \reward)$ \ttd{fix this!}
\footnote{This paper does not deal extensively with the reward, but it is included for completeness.
Also, it is not critical that the agent has access to the full state information (even though the existence of such underlying states is presumed by the theoretical framework), because useful partitions of the state space can be drawn without it.}
} 
%
Forecasts can be succinctly formalized using the standard reinforcement-learning (RL) framework (see Sutton and Barto, 1998)\nocite{SuttonBarto:Book}.
In this framework a learning agent's interaction with its environment is modeled as a \emph{Markov decision process} (MDP), which unfolds over a series of discrete time steps $t\in\{0,1,2,...\}$.
At each time step $t$ the agent takes an action $\action_t \in \allActions$ from its current state $\state_t\in\allStates$.
As a result of the action the agent transitions to a state $\state_{t+1}\in\allStates$, and receives a reward $r_{t+1}\in\Reals$.
The dynamics underlying the environment are fully described by the MDP's state-to-state transition probabilities $\CalP^\action_{\state\state'} = Pr\{\state_{t+1}\!=\!\state' \mid \state_t\!=\!\state,\action_t\!=\!\action\}$ and rewards $\CalR^\action_{\state\state'}\in\Reals$, defined for all actions $\action$ and states $\state$ and $\state'\!$, where $r_{t+1} = \CalR^\action_{\state\state'}$ when $\state_t\!=\!\state,\action_t\!=\!\action$, and $\state_{t+1}\!=\!\state'$.
The agent's preference for each action in each state is described by a \emph{policy}, $\policy(s,\action)=Pr\{\action_t\!=\!\action \mid s_t\!=\!s\}$.
%
The reinforcement-learning agent's task is to find an optimal policy $\policy^*$\!\!\!, which specifies an action in every state that will maximize the agent's expected future receipt of reward.
The amount of reward the agent can expect to receive from any state $\state$ is called $V^\policy(\state)$:
\commentout{if it follows In an MDP, maximizing reward is generally the same as maximizing $V(s)$ the expected sum of all future rewards the agent from every starting state $\state$.
This value,}
\begin{align} 
  V^\policy(\state) = \Exp\left[\sum_{t = 0}^\infty \gamma^t r_{t+1} \mid \state=\state_0, \policy\right],
  \label{eq:efdr}
\end{align}
which is the expected sum of all future rewards the agent will receive if it starts in state $\state$ and follows policy $\policy$ forever, where future rewards may be exponentially discounted by $\gamma\in [0,1]$.
$V(s)$ is called the \emph{value} of state $\state$, and $V$ is known as the \emph{state-value function}.
The reinforcement learner generally refines its policy through repeated interaction with the environment so as to maximize $V(\state)$ for all states $\state \in \allStates$.
In standard reinforcement learning, $V(\state)$ is often estimated using an approximation technique, and the agent learns to improve these estimates through repeated interaction with the environment.
The policy is regularly modified to take into account the latest state values, and the state values are regularly updated to reflect the latest policy modifications.
This process of updating the policy and the values generally continues until the agent's performance is satisfactory or until no further improvements are possible.

There are many learning algorithms in common practice that will guarantee convergence to an optimal policy, given some fairly reasonable constraints.
Most algorithms make estimates of either the state-value function $V(s)$ or the \emph{action-value function} $Q: (\allStates, \allActions) \ra \Reals$, which is different from $V(s)$ only in that it defines the expected future reward for each state-action pair $(\state, \action)$ rather than for each state.
Thus, $\hat{V}^π(\state)$ and $\hat{Q}^π(\state,\action)$ are the agent's estimates of its expected future reward if it starts with state $\state$—or state-action pair $(\state, \action)$—and takes actions according to policy $π$ forever after that.
\commentout{
Most algorithms either estimate the state values $V(s)$ or the \emph{action-value function} $Q: (\allStates, \allActions) \ra \Reals$, which define the expected future discounted reward (Equation~\ref{eq:efdr}) for each state (or each state-action pair).
These values are generally estimated online, each usually contingent on the agent's current policy $π$.
Most algorithms either build an estimate of the \emph{state-value function} $V(s)$ or an \emph{action-value function} $Q: (\allStates, \allActions) \ra \Reals$, which define the expected future discounted reward (Equation~\ref{eq:efdr}) for each state (or each state-action pair).
Most algorithms either  make use of a \emph{state-value function} $V: \allStates \ra \Reals$ or an \emph{action-value function} $Q: (\allStates, \allActions) \ra \Reals$, which define the expected future discounted reward (Equation~\ref{eq:efdr}) for each state (or each state-action pair).
These values are generally estimated online, each usually contingent on the agent's current policy $π$.
Thus, $\hat{V}^π(\state)$ and $\hat{Q}^π(\state,\action)$ are estimates of the agent's expected future discounted reward from the current state $\state$—or state-action pair $(\state, \action)$—when the agent takes actions according to policy $π$.
}

General value functions (GVFs)~\cite{Sutton:Horde,GQ,SuttonRafolsKoop}—which \IA refer to by the nickname, ``forecasts,'' for explanatory convenience—are extended versions of value functions.
Like value functions, forecasts are behavior-dependent functions of state, 
but they are general-purpose predictors that can be used for estimating a broad class of values computable from the agent's future sensorimotor stream.
There are two kinds of values the forecast predicts: \cumu\ values, which are summed up during the course of the specified behavior, and termination values, which are added to the sum when the behavior terminates.
Therefore, unlike value functions, forecasts are contingent upon behaviors with specific initiation and termination conditions.
%
%
%
%
Accordingly, each forecast function $\forecastfun^i$ consists of two parts, an \emph{option}~\cite{SuttonPrecupSingh:Options}, which specifies the behavior (including its initiation and termination conditions), and an \emph{\target}, which specifies the \cumu\ and termination values.
Both (options and outcomes) are described next.

The \emph{option} is a 3-tuple $(\pi, I, \termP)$, where $\pi$ is a policy; $I:\allStates\ra\{0,1\}$ is an \emph{initiation set}---the states in which the policy can be started; and $\termP:\allStates\ra [0,1]$ is the \emph{termination probability}---the probability for each state that the option will terminate should the agent visit that state.
The initiation set and termination probability allow specification of the conditions by which the policy can start and stop---something a policy alone cannot do.
Each option describes one possible way for the agent to behave, including how that way of behaving can begin and end.

The \emph{\target} is a tuple $(\accv, \termv)$, where $\accv:(\allStates\times \allActions)\ra\Reals$ is a \emph{\cumu} value defined for every state-action pair reachable while the option is being followed and summed (accumulated) while the agent is following the option; and $\termv:\allStates\ra\Reals$ is a \emph{termination} value, defined wherever the option might terminate.
These values, $\accv$ and $\termv$, are similar to the reward value from traditional reinforcement learning, but they can in principle be any values available from the agent's sensorimotor stream.
(Many examples of $\accv$ and $\termv$ appear later in the article.)

Thus, every forecast $\forecastfun^i$ is a function of the state as specified by the five components of the forecast definition:
\begin{equation}
  \label{eq:definition}
  \forecastfun^i(\state) \equiv \forecastfun^{\pi^i\!,\ I^i\!,\ \termP^i\!,\  \accv^i\!,\  \termv^i}(\state),
\end{equation}
but because the agent will have many forecasts of this form, \IA will drop the superscript $i$ for clarity. 

Each forecast is a scalar function of state, whose output in each state is the expected sum of the \target s the agent will encounter if it follows the option (\ie if the agent behaves as described by the option).
That is, the output of each forecast function $\forecastfun$ is the  expected sum of all the \cumu\ values $\accv$ the agent encounters while following the option, plus the termination value $\termv$ encountered when the option terminates at some future time step $k$; \ie
\begin{equation}
\label{eq:ideal}
\forecastfun(\state) = \Exp\left[\accv_1 + \accv_2 + \ldots + \accv_{k-1} + \termv_k \mid \pi, β, \state_0 = s \right].
%
%
%
\end{equation}
Thus, the output of the forecast is a scalar \emph{prediction} about the sum of the $\accv$ and $\termv$ values the agent will encounter if it follows the option.
A good way of thinking about the forecast function is that it corresponds to a question the agent might pose to itself, ``if I act in this way, what will the outcome be?''
(In fact, forecast definitions are referred to as ``questions'' in the Horde architecture~\cite{Sutton:Horde}.)

Just as with the value function, the agent constantly refines its internal estimate of each forecast's output through interaction with the world in a process of continual improvement.
Like the ideal value, each forecast estimate is also a function of the agent's state, but it is a learned, \emph{approximated} function of the agent's \emph{approximation} of its state:
\begin{equation}
\label{eq:estimate}
\forecast(\Estate) \approx \forecastfun(\state)
\end{equation}
More will be said about this approximation below in Section~\ref{sec:forecast-estimation}.

Thus, each ``forecast'' is  really  three different things: a specific, unique function of state as defined in Equation~\ref{eq:definition}; the true output or ``ideal value'' of that function as shown in Equation~\ref{eq:ideal}; and the agent's continually improving estimate of that ideal value, as represented in Equation~\ref{eq:estimate}.
Because the risk of confusion is high, I will choose the lesser risk of redundancy and refer to these explicitly as the forecast ``function'' or ``definition''; the forecast ``output,'' ``value,'' or ``ideal value''; and the agent's forecast ``estimate,'' respectively.
\begin{table}[h]
  \center
\begin{tabular}{ll}
\hline
forecast \emph{function} or \emph{definition} & Equation~\ref{eq:definition} \\
forecast \emph{output}, \emph{value}, or \emph{ideal value} & Equation~\ref{eq:ideal}\\
forecast \emph{estimate} & Equation~\ref{eq:estimate}\\
\hline
\end{tabular}
\end{table}

\subsection{Forecast Examples} For many predictions, it is useful for $\termv$ to be zero everywhere.
For example, the following prediction is a description an agent might use to estimate its number of steps to the nearest wall.

\vspace{12 pt}
Example forecast {\bf A}: Steps to wall from here.\nopagebreak%
\footnote{Notice that this prediction, expressed in the first person from the agent's subjective view, might tempt one to re-interpret the meaning of the forecast as something objective, such as the agent's ``distance to the wall,'' but in general this is a risky interpretation, as there is no guarantee that much or even \emph{any} of a continual-learning agent's knowledge will correspond to any objective entity.
In fact, one suggestion of this paper is that it is perhaps the confusion between subjective descriptions (which are personal yet verifiable from one's sensorimotor stream) and objective descriptions (which promote consensual and social agreement, but are not necessarily verifiable from one's sensorimotor stream) that has lead to some seemingly insoluble problems in AI, particularly symbolic AI, when objective definitions are sought for ultimately subjective phenomena.}
\begin{my_list}
\vspace{-5 pt}
\item Option: ``Walk to nearest wall''
\begin{my_list}
\item    $I = 1$ if there is a wall nearby; $0$ otherwise.
\item    π = turn to face nearest wall, or, if I am already facing the nearest wall, walk forward one step.
\item    $β = 1$ when at wall; $0$ otherwise.
\end{my_list}
\item \Target:
\begin{my_list}
\item $\accv = 1$ if $\action = $ forward step; 0 otherwise
\item $\termv = 0$
\end{my_list}
\end{my_list}
%
This forecast function estimates the sum of \accv\ values that the agent will see before option termination.
Since \accv\ is always 1, this estimate corresponds to the total number of time steps that will elapse as the agent walks toward the nearest wall and then stops.

For other predictions, it is useful for $\accv$ to be zero everywhere, such as for the following forecast, which predicts whether the agent can shoot a basket from its current position.
In this instance it is useful to use an option that (hypothetically) already exists, namely, the option for shooting baskets.%
\footnote{Note that these examples are quite abstract and rely on complex perceptions (such as holding a basketball) and options (such as throwing a basketball towards a hoop), while the forecasts in the demonstration will build up all abstractions entirely from the sensorimotor stream rather than relying on those that you and I already know.}

\vspace{12 pt}
Example forecast {\bf B}: Probability of shooting a basket from here. \nopagebreak%
\begin{my_list} 
\vspace{-5pt}
\item Option: ``Shoot basket''
\begin{my_list}
\item $I = 1$ if I am on a basketball court holding a basketball; $0$ otherwise
\item π = raise basketball, throw towards hoop, watch ball's trajectory
\item $β = 1$ when ball passes through or bounces away from net; $0$ otherwise
\end{my_list}
\item \Target:
\begin{my_list}
\item $\accv = 0$
\item $\termv = 1$ if ball passes through net; $0$ otherwise
\end{my_list}
\end{my_list}
One might describe this forecast function as formally specifying the question: ``if I try to shoot a basket from here, with what probability will I succeed?''
The forecast's output or \emph{ideal value} is the specific probability that the ball will go in.
In general, the ideal value is never known exactly; but it can be estimated from experience.


\subsection{Forecast Estimation}
\label{sec:forecast-estimation}
In any actual agent, the forecast ideal values are not known but must be estimated from the agent's experience with its environment.
The forecast estimate $\forecast(\Estate)$ is an approximation of the forecast's ideal value $\forecastfun(\state)$ and is a function of the agent's estimate of its current state.
The predictive agent represents its state as a vector, $\statevec_t$, a set of values encapsulating all the information the agent has about its environment at the current time step, $t$.
This information consists of its lowest-level sensory information (possibly including proprioceptive information) as well as current estimates for all its forecasts.
\begin{align} 
  \label{eq:states}
  \statevec_{t} \equiv \obsvec_{t} \concat\allForecasts(\Estatevec_{t-1}), 
\end{align}
where $\obsvec$ is the vector of lowest-level sensory information, $\allForecasts$ is the full set of individual forecast estimates, and the symbol $\concat$ concatenates two vectors.
Each forecast estimate is computed as a parameterized function of this state vector.

Learning can be done online, modifying the parameters at every time step to reduce the prediction error.
But to compute an error, of course, requires having a target value.
It is therefore critical that the predictions correspond to measurable quantities to supply the targets needed to calculate the errors, update the parameters, and improve future estimates.
But a continual-learning agent has no access to the ideal forecast target values.
How can it learn without knowing the correct values of what it is estimating?
This requires a trick, and that trick is temporal-difference learning~\cite{Sutton:TD}.

The definition in Equation~\ref{eq:ideal} leads to the following relationship (see  Appendix~\ref{sec:forwardview} for derivation),  between a forecast estimate at time step $t$ and its estimate at time step $t+1$:
\begin{align} 
\label{eq:forecastTD}
\forecastfun(\state) &=
    \termP(\state) \termv(\state) 
          + (1-\termP(\state)) 
          \left(
             \accv(\state) 
             + 
             \Exp[\forecastfun(\state' \mid π,β, \state=\state_t,\state'=\state_{t+1})]
             \right).
\end{align}
This equation shows that when an agent follows a forecast's option, the forecast's value for any state is a function of its expected value at the following state, combined with the \accv\ and \termv\ values the agent encounters on that time step (which depends on whether the option terminates on that time step).
Thus, at every time step, the forecast option either terminates or not, and the target, $\mtarget_t$ is formed accordingly from Equation~\ref{eq:forecastTD}:
\begin{align*}
  \mtarget_t &= \begin{cases}
                           \termv_{t} & \text{if } π \text{ terminates at } t, \\
                           \accv_{t+1} + \forecast(\statevec_{t+1}) & \text{otherwise};
                         \end{cases} \\
\end{align*}
Comparing the current estimate to the target produces the TD-error, $\delta$:
\begin{align*}
  \delta_t & = \mtarget_t - \forecast(\Estatevec_t).
\end{align*}
The parameters of the function approximator $\forecast$ can then be modified to reduce this error.
Thus each forecast is a prediction that is both verifiable and learnable through the agent's own sensorimotor experience, but crucially the agent never needs to rely on labelled data or human validation of its knowledge.
Like the rest of \usH\ the agent has only its experiences to help it make sense of its data stream, and ultimately, verification through interaction is the only recourse the agent has to ascertain the extent to which its predictions are accurate.

It should be noted that Equation~\ref{eq:forecastTD} assumes the agent takes all actions according to the forecast's policy, but different forecasts will generally have different policies, and the agent cannot follow all policies at all times.
Yet the agent would learn more \emph{efficiently} if it could apply its experiences from following one policy to all forecasts for which those experiences might be informative.
Training a predictor based on one policy while following the policy of a different predictor is called ``off-policy'' learning.
As a real-world example, consider forecast A in the previous section, which predicts ``the number of time steps to the nearest wall.''
Using off-policy learning, the agent can improve its estimate for this forecast whenever it takes a step toward the wall, even if it is currently following a different policy, say one that takes it on a path toward the kitchen.
But if its current experience is informative for both forecasts, then both estimates can be improved using the same data.
Since continual-learning agents will likely acquire a very large number of forecasts encoding a great deal of knowledge and skills, it is essential for each to be updated using all applicable experience.
Fortunately, a great deal of recent research on off-policy learning has made forecasts finally feasible:
the Horde Architecture~\cite{Sutton:Horde}, for example, has demonstrated that an agent can train forecasts off policy---thousands of them---simultaneously.

But continual learning requires more than just training many forecasts at once. It requires the isolaminar construction of knowledge and skills.
While many learning algorithms allow the composition of intermediate outputs (for example, the hidden units of neural networks), much less work has been done to allow the composition of temporal predictions, where new predictions are built to estimate the future value of existing predictions.
A particularly powerful property of forecast functions is that they can be composed or \emph{layered}, and it is this compositional power that introduces abstraction into the representation, allowing specification of a very wide range of predictions.

\subsection{Composing Forecast Functions}
Once a forecast's estimate becomes reliable, it can be used as a building block for layering with other forecasts.
There are three primary ways that a forecast can be used for layering: (1) as an input to other forecast estimates, (2) as the \accv\ or \termv\ value of a new forecast, and (3) as the value that a new option learns to optimize.

For example, the agent may maintain a set of forecast estimates such as the example forecasts A and B above.
The agent continually maintains these estimates as part of the agent's state vector, each estimate thereby serving as an \emph{input} to calculate and update all other forecast estimates (Equation~\ref{eq:states}).
This ``composition of estimates'' is the first method of layering.

The second method of layering is to use an existing forecast value as the \accv\ or \termv\ component in a new forecast definition.
For example, the agent may find it useful to have a new forecast (call it forecast C) that predicts what the value of Forecast A will be after the agent has executed a particular policy.

\vspace{12 pt}
Example forecast {\bf C}: Steps to wall on my left\nopagebreak%
\begin{my_list} 
\vspace{-5pt}
\item Option: Turn 90° left
\item \Target:
\begin{my_list}
\item $\accv = 0 $ 
\item $\termv = $ Forecast A
\end{my_list}
\end{my_list}
This forecast predicts the value that Forecast A will have after the agent turns 90° to its left. (Assuming that the agent has an option that reliably turns the agent 90° left.)

The third method of layering is to use an existing forecast estimate for training an option policy.
In general, this means using reinforcement-learning methods to modify the option's policy to navigate to a state where the forecast's value is maximized (or minimized).
For example, an option could be created to maximize the value of Forecast B.
The policy, which improves with experience, would lead the agent to a place where its chance of shooting a basket is greatest.
\commentout{
 \vspace{12 pt}
\begin{my_list}
\item Option: ``Maximize Forecast B''
\begin{my_list}
\item $I = 1$ if I am on a basketball court holding a basketball; $0$ otherwise
\item π = Maximize Forecast B
\item $β = 0.1$
\end{my_list}
\end{my_list}
}
%
Once the option is trained, a new forecast D can be created that uses this new option. Forecast D might, for example, estimate how many time steps would be required to reach a state with a high chance of shooting a basket.
%
%

\vspace{12 pt}
Example forecast {\bf D}: Steps to sweet spot\nopagebreak%
\begin{my_list} 
\vspace{-5pt}
\item Option: ``Maximize Forecast B''
\begin{my_list}
\item $I = 1$ if I am on a basketball court holding a basketball; $0$ otherwise
\item π = Maximize Forecast B
\item $β = 1$ if Forecast B > 0.9; $0$ otherwise
\end{my_list}
\item \Target:
\begin{my_list}
\item $\accv = 1$
\item $\termv = 0$ 
\end{my_list}
\end{my_list}

\noindent
Similar to Forecast A above, this forecast predicts the agent's subjective sense of distance (number of time steps away) from a set of states that are defined by a subjective property (high likelihood of making a basket). 
That subjective property is a set of states where Forecast B has a high value, defining a sweet spot on the court.
Forecast D uses Forecast B as a kind of landmark or navigational aid, estimating how quickly the agent can reach the sweet spot from wherever it is on the court. 

In a similar way, a home robot might have something like a ``recharge'' forecast indicating the probability of success if the robot tries to recharge itself in its current position.
Though that forecast seems to have only one purpose—to estimate whether the robot can successfully recharge—it can be used and built upon.
For example, the robot could make predictions about what actions will increase that probability estimate.
It can also create an option that is trained to maximize that probability. 
Since only a small fraction of its state space has a high value for that forecast, the robot can use it as a sort of navigational landmark to orient itself when in the nearby vicinity. 

Thus, as these examples begin to show, composition is the glue that binds forecasts together, allowing this single, isolaminar mechanism---forecasts---to be built into layers of increasing abstraction to form high-level knowledge.
The next section demonstrates many different examples of forecast composition, starting from the sensorimotor stream and building up complex knowledge of the agent's surroundings.


\commentout{
\subsection{Forecasts}
\label{sec:forecasts}

For example, a specification might formally describe something like: ``The probability that I will detect my keys\emph{if} I put my hand in my pocket,'' or ``The amount of time that will probably pass before I ski into a tree if I now lean a bit to the right.''
This specification is the ``forecast function'' also sometimes called the ``question.''

An example of an event forecast is, ``I will see the patio light come on if I flip one of those switches,'' or ``I will find food if I go to the party tonight.''
An example of a cumulative forecast is ``How long it will take me to find the patio switch,'' or ``How much food I will see at the party tonight.''
And an example of a mixed forecast is, ``How many lights I will have turned on when I finally stop flipping the switches,'' or ``How much I will eat at the party tonight'' (knowing that \I may stop eating either because the party ends or because \I \am too stuffed).
\comment{It's very hard to find examples that also makes sense in the third case.}

It is often useful to predict how much of some quantity will accumulate during a certain activity; for example, ``how much food I will eat at dinner tonight.''
These cumulative forecasts are particularly useful for measuring time or distance; for example, ``How much time will elapse before I see the ball hit ground if I throw it into the air,'' or
``How many steps it will take me to get to the door from here,'' or
``How much time I will have after the light turns yellow to make it through the intersection.''

 
Predictions require a time frame.
If \I predict an earthquake in California, the prediction is meaningless without an associated time frame.
Different predictions have different time frames.
When \I predict that a person's mood will improve if \I give the person ice cream, this improvement may fall within a few minutes' time, while if \I predict that \I will see red if \I look at a particular wall, \I expect to see red immediately upon looking at the wall, not minutes or hours later.
Thus, \my representation should somehow include the likely time span of the prediction.

}


%% file: demonstration.tex

\section{Demonstration}
\label{sec:demo}

This section demonstrates the power of forecasts by way of a detailed thought experiment.
The goal of the thought experiment is to investigate the ability of forecasts to form high-level knowledge from an agent's raw sensorimotor stream alone.
Therefore, this section will describe an agent, its sensorimotor apparatus, and a microworld environment the agent inhabits.
It then builds up—in an isolaminar fashion—a series of layered forecasts in an attempt to capture high-level knowledge the agent could have about its environment.
\MyA hope is to describe the agent, its environment, and its forecasts in sufficient formal detail that the careful reader can verify, layer by layer, the reasonableness and plausibility of the thought experiment's conclusions.

Before continuing, however, there are two questions that call out for an immediate answer; the first: ``what is high-level knowledge?''; the second: ``why a thought experiment?''
\IA address each of these directly below before proceeding with the demonstration. 

\subsection{What is high-level knowledge?}
The goal of the demonstration is to determine whether forecasts can form high-level knowledge, but this goal would be vacuous without a clear understanding of the term ``high-level knowledge.''
Ideally, it should have a formal definition that more or less matches the intuitive notion.
But finding a complete, formal and uncontroversial definition of high-level knowledge may not be significantly less involved than building a complete, formal, and uncontroversial method for representing it.
\IA therefore offer an informal description.
By ``knowledge'' \IA am referring to an agent's ability to detect and exploit statistical regularities in its sensorimotor stream.
(An agent might, for example, use its knowledge for making plans and choosing intelligent actions).
``Low-level knowledge'' refers to statements about the raw sensorimotor signals: for example, predictions about the observations that immediately result from actions.
By ``high-level knowledge'' \IA am referring to statements about \emph{abstractions}—regularities based on but removed from the raw sensory or motor signals.
For example, an agent that has knowledge about the arrangement of walls or rooms in its vicinity could be said to have abstract, high-level knowledge, despite the fact that this knowledge may ultimately correspond to patterns the agent has found in its sensorimotor stream.%
\footnote{In traditional knowledge-based systems such as CYC, there is \emph{only} high-level knowledge—abstractions having no connection to the sensorimotor stream.
In traditional sensorimotor systems (agents that compute actions in response to observations) such as reinforcement-learning systems, there is relatively little abstraction: generally there is no other knowledge than predictions about the agent's raw observations immediately following its actions.}
The crunch of an apple, 
the smile on someone's face, 
the skills involved in playing the piano, 
the smoothness of glass, 
the softness of a pillow, 
the layout of rooms in a house—one can discuss these all in terms far removed from the raw sensorimotor signals, though (\IA maintain) they all correspond to statistical regularities found in the sensorimotor stream.



Somewhat more formally, an abstraction is a feature or function of the sensorimotor stream, whereby greater abstraction generally results from a deeper layering of features.
The high-level outputs of a deep neural network, removed by many isolaminar layers from the inputs, are nevertheless ultimately a function of the inputs, where each intervening level provides the working vocabulary for the next higher level.
Analogously, with an effective isolaminar method for discovering regularities in the sensorimotor stream, each layer should allow construction of a richer, more descriptive, easier-to-use vocabulary for the next-higher layer.
That is, each layer provides the opportunity for more meaningful, more useful abstraction.
\IA propose that forecasts do exactly that: they provide a framework for the representation of meaningful, useful abstractions from the sensorimotor stream.

\subsection{Why a Thought Experiment?}

There are multiple ways one could investigate the ability of forecasts to represent abstract, high-level knowledge.
One would be to come up with a method for creating forecasts automatically, implement it in a robot, then see if the robot exhibits behavior that somehow demonstrates high-level knowledge.
While that endeavor may someday be feasible, it seems at the moment premature---a distant prospect for a still nascent technology.
A more feasible demonstration might be an agent that learns a deeply layered representation of a \emph{simulated} environment.
While simpler and faster than an actual robot, the simulation would still require invention of an efficient and as-yet-unknown method for finding and creating new forecasts, which could likely be a major undertaking.

\IA have instead opted for a more immediate route that investigates the \emph{potential} ability of forecasts to capture high-level knowledge through an extended thought experiment in which forecasts are hand built for an imagined agent in an imagined environment.
The agent and environment are chosen to be sophisticated enough to allow an interesting set of forecasts, yet simple enough that \you the reader can validate \myA claims without a computer.
\IA then formally describe a set of forecasts and ask \you to assess whether (1) they are learnable and verifiable by the agent, and (2) they succeed at capturing high-level knowledge.

The thought experiment is focused exclusively on the representational capacity of forecasts, and \IA therefore assume that all other parts of the system function perfectly.
In particular, forecasts require an (unspecified) feedforward function approximator, so the thought experiment assumes that this function approximator is perfect and is always capable of finding a concise mapping from inputs to outputs \emph{if such a mapping exists}.
This assumption is for explanatory convenience, keeping the focus on the forecasts; and as this is a thought experiment in which any function approximator is possible, \IA prefer to use the perfect one, since the price is the same, and there is less maintenance involved.
In any case, \IA will not ask it to find regularities that might seem unreasonable, and in all cases, it should be clear that the necessary information is present in the data stream.

After describing the agent and environment, \IA will start building up layers of forecasts.
Each layer relies on some degree of learning from experience, and in line with the nature of thought experiments, this one will assume the agent's experience is sufficient to allow the function approximator to achieve the required degree of accuracy.
Every layer is built by hand because the goal is to decide whether forecasts will ultimately be capable of representing high-level knowledge.
Each layer will use only forecasts and options built on top of previously constructed forecasts and options, and after it is built \IA will discuss the degree to which that layer develops the agent's abstract understanding of its environment, and at the end of the thought experiment \IA will discuss to what extent the forecasts, combined together, have in fact provided the agent with high-level knowledge.
Thus, the two overall goals of this thought experiment are: (1) to teach, in a quasi tutorial style, how forecasts can be layered, and (2) to find out whether forecasts can be layered robustly such that sophisticated abstractions can be built and then used as the building blocks for even more sophisticated abstractions, resulting eventually in high-level knowledge.


\subsection{The Agent and its Microworld}
\label{sec:a&m}

\begin{figure}
  \centering%
  \includegraphics[height=0.25\textheight]{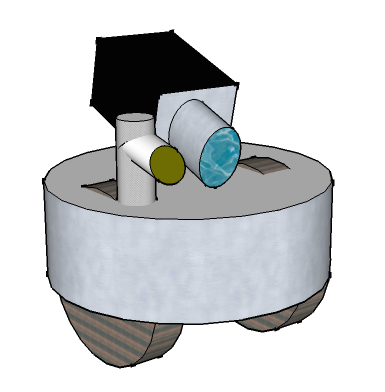}
  \caption{The robot is the agent's interface to its world.
It has a camera and a finger mounted atop a round base.
The robot can extend the finger, which has a contact sensor at its tip.
The robot can also roll forward and backward and rotate in place.}
\label{fig:robot}
\end{figure}

The agent is embodied within the robot shown in Figure~\ref{fig:robot}.
The robot is essentially cylindrical, having a round base supporting a camera and finger, and it has two wheels allowing it to roll backward and forward (in the direction the finger and camera are pointing) and to rotate in place.

\begin{figure}
  \begin{minipage}[c]{0.48\textwidth}
    \centering%
    \includegraphics[width=0.99\textwidth]{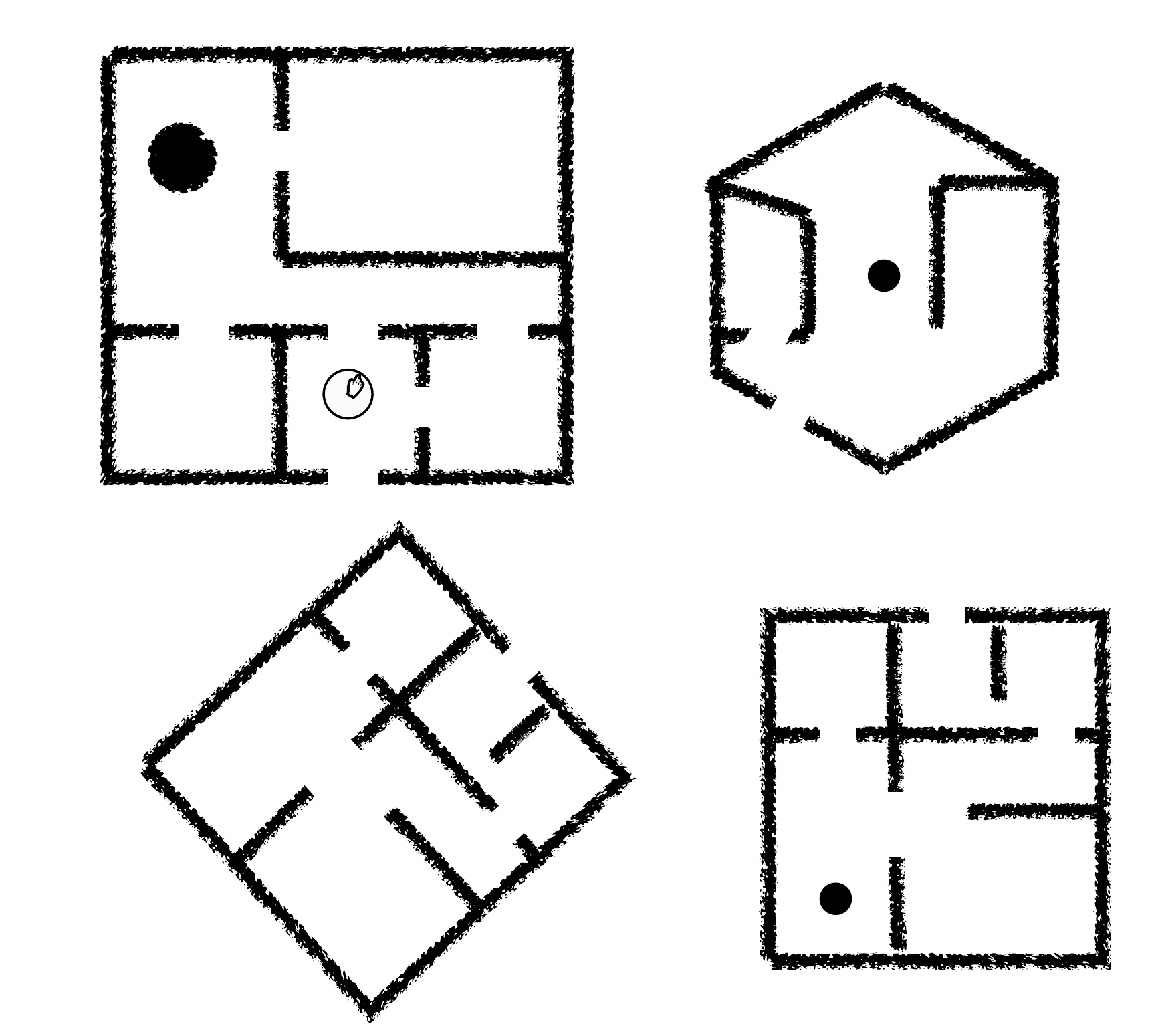}
  \end{minipage}
  \begin{minipage}[c]{0.48\textwidth}
    \centering%
    \includegraphics[width=0.8\textwidth]{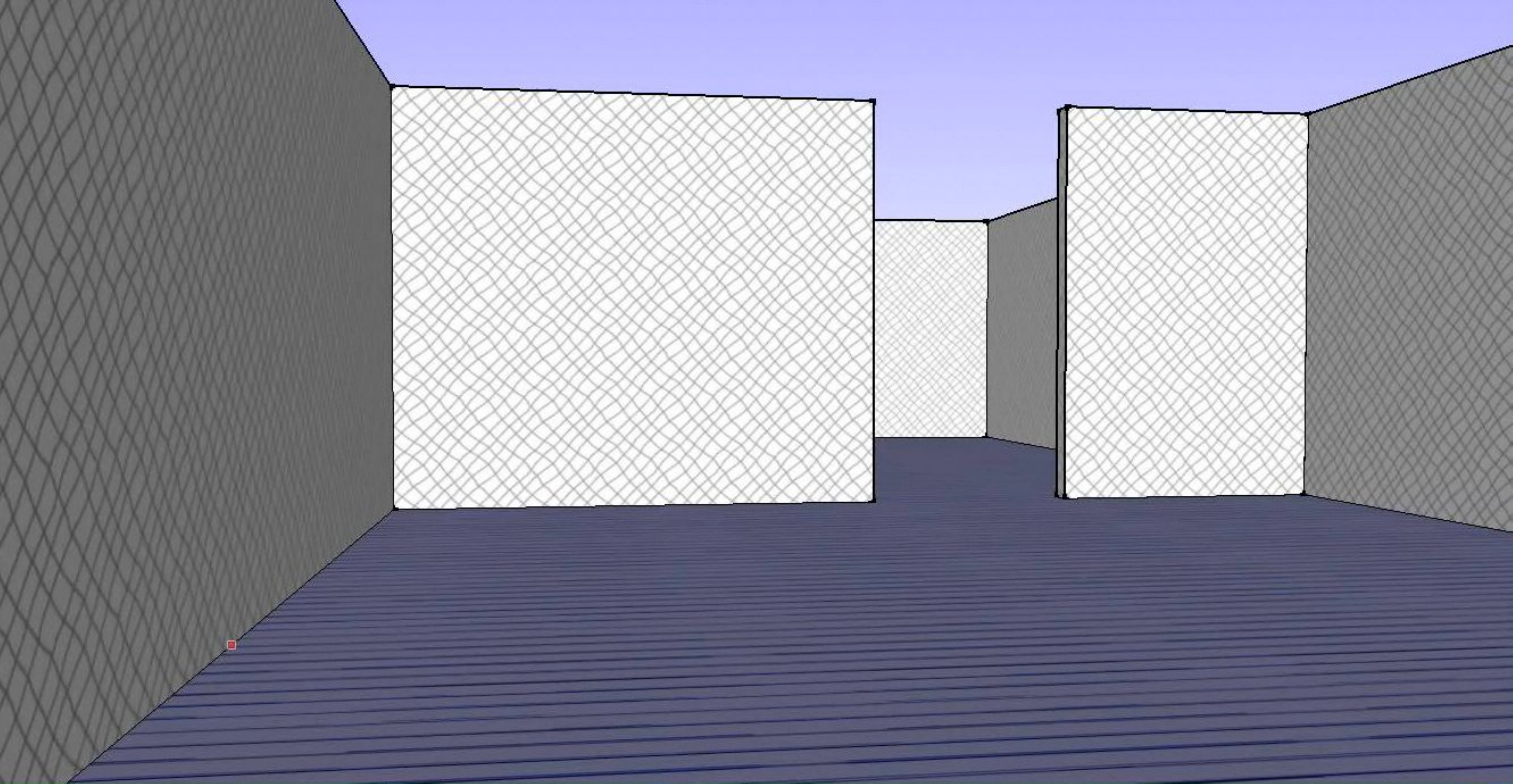}
    \includegraphics[width=0.8\textwidth]{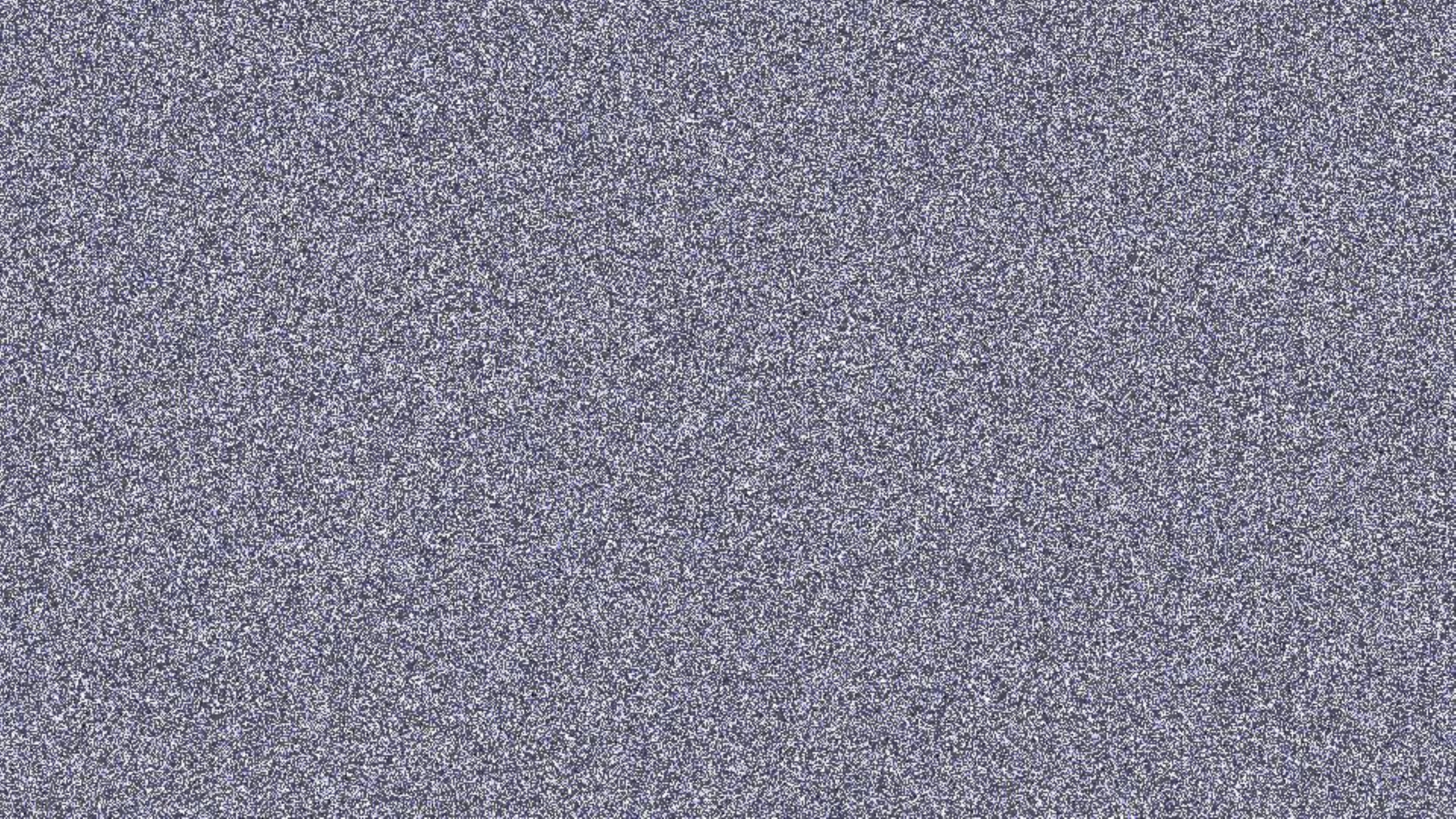}
  \end{minipage}
  \caption{The agent inhabits a microworld.
{\bf Left (looking from the top):} The microworld is made up of a number of \emph{houses}, each consisting of its own \emph{floor plan}.
Each floor plan consists of a number of obstacles and barriers that partition the microworld into spaces or rooms of various sizes and shapes, with doorways that connect these rooms.
The agent is shown as a circle with a hand pointing forward.
{\bf Top~Right (looking through the agent's camera lens):} The walls extend vertically and are decorated with a distinct pattern.
The floors have a different, contrasting pattern.
{\bf Bottom~Right (seeing what the agent sees):} The agent's camera pixeles are randomized, and before it has experience with the world, it has no way to interpret the images it sees.
}
\label{fig:world}
\end{figure}

The robot resides in a $2½$ dimensional world as shown in Figure~\ref{fig:world}.
Viewed from above, the world looks like a 2-dimensional surface divided by barriers,
some of which are round obstacles, while others are straight, dividing the surface into open spaces laid out as houses, rooms, and doorways.
Viewed from the position of the agent, however, the barriers rise up from the floor, impede forward motion, and are painted with a visual pattern that distinguishes them from the floor.

The agent interacts with the world through the robot's sensors and motor commands, summarized in Table~\ref{table:s&a}.
The robot provides two sensory modalities: touch and vision.
Vision is implemented by a rudimentary camera with a 30° field of view whose exact specifications are not important to the thought experiment.
\IA assume again, however, that the hardware is sufficient to allow the robot to make the visual distinctions required by the demands of the thought experiment as laid out below; for example, whenever the robot is facing a wall and one or more time steps away (rolling forward), the camera should be able to see both the wall and the floor with sufficient resolution that it can distinguish the pattern on each.
Touch is simpler: there is a single binary touch signal generated by a sensor at the end of the robot's finger.
It returns a value of 1 only when the fingertip has made contact with something in the environment during the previous time step.

\newcommand{\ST}{\textcircled{\tiny{T}}}
\newcommand{\SC}{\textcircled{\tiny \sc{C}}}

\begin{table}
  \begin{minipage}[t]{0.35\textwidth}
    \vspace{0pt}
    \begin{tabular}{clc}
      \multicolumn{3}{c}{Senses} \\
      \toprule
      Number & Description              & Abbrev \\
      \midrule
      1 & finger contact & \ST \\
      2 & camera & \SC \\
      \bottomrule
    \end{tabular}
  \end{minipage}
  \hfill
  \begin{minipage}[t]{0.58\textwidth}
    \vspace{0pt}
    \begin{tabular}{clc}
      \multicolumn{3}{c}{Actions (Primitive Options)} \\
      \toprule
      \mysplit{Option \\ Number} & Description              & Abbrev \\
      \midrule
      \Onrf   & roll forward             & \Orf   \\
      \Onrb   & roll backward            & \Orb   \\
      \Onrotl & rotate left (counterclockwise) 30°          & \Orotl \\
      \Onrotr & rotate right (clockwise) 30°         & \Orotr \\
      \Onef   & extend finger (retracts automatically) & \Oef   \\
      \bottomrule
    \end{tabular}
  \end{minipage}
\caption{The robot's sensorimotor apparatus consists of two sensors and five possible actions.
\textbf{Left:} The touch sensor \ST\ is binary whereas the camera sensor \SC\ provides an array of pixel values in [0,1].
\textbf{Right:}~Exactly one action must be chosen at every time step.
The robot executes the action to completion by the end of the time step.
The resulting sensor values are available as inputs at the following time step.
Each action can be considered a primitive option whose termination probability is always 1.
}
\label{table:s&a}
\end{table}

\forcsvlist\DefOLayer{rf, rb, rotl, rotr, ef}

At every (discrete) time step, the robot may execute one of the five actions.
In the case of action \Onef~(\Oef~\!), the robot extends its finger beyond the edge of its base and then immediately retracts it, all within the course of a single time step.
The fingertip can come into contact with a barrier only when extended, and then only if the robot is directly facing a wall (within 10°) and already contacting the wall at its base.
Rotation commands (actions \Onrotl\ and \Onrotr) cause the robot to rotate in place by 30° clockwise or counterclockwise.
These actions can be carried out everywhere in the microworld, with no exceptions, and will always result in the desired rotation.%
\footnote{Thus, for clarity of exposition only, the actions in the demonstration are deterministic. However, there is nothing in the theory or in the learning algorithms that necessitates this constraint. Because forecasts are defined as expectations, stochastic actions will simply result in different forecast values, but will not fundamentally alter the continual-learning process.}
In contrast, the robot will not move past a barrier if a \Orf~(roll forward) or \Orb~(roll backward) action is selected when a barrier impedes motion (\ie\ when the robot is in contact with a wall and any component of its motion vector is directed toward the wall).
Thus, if the robot is in contact with a wall, it can only roll away from the wall or parallel to it.
The exact amount that the robot moves forward or backward is not critical to the thought experiment, but it should be sufficiently small that the robot requires many such actions to cross a normal-size room, but large enough that the visual input generally changes as a consequence of taking the action.

\subsection{Layered Forecasts}

It is now time to begin constructing abstractions.
The following demonstration builds up abstractions in a series of \nlayers\ numbered \emph{layers}, where layer zero is the raw sensorimotor interface, given in Table~\ref{table:s&a}.
Thereafter, each successively more abstract layer is built on top of the previous layer and is required by the next: if forecast A is required by forecast B, then A is described before B.
Furthermore, the layers are functionally rather than structurally dependent; i.e., it is the functionality of A that is required to produce the functionality of B, rather than the internal structure of B that contains A or calls A as a software component.
In other words, it is the abstractions that depend on each other, and 
the forecasts are defined so as to capture these dependencies.
Layer by layer, the function approximator learns to estimate forecast values based on the relationships specified in the forecast definitions; these estimates can then be incorporated into the estimates of the forecasts defined in subsequent layers.

\begin{figure}
  \centering%
  \includegraphics[height=0.25\textheight]{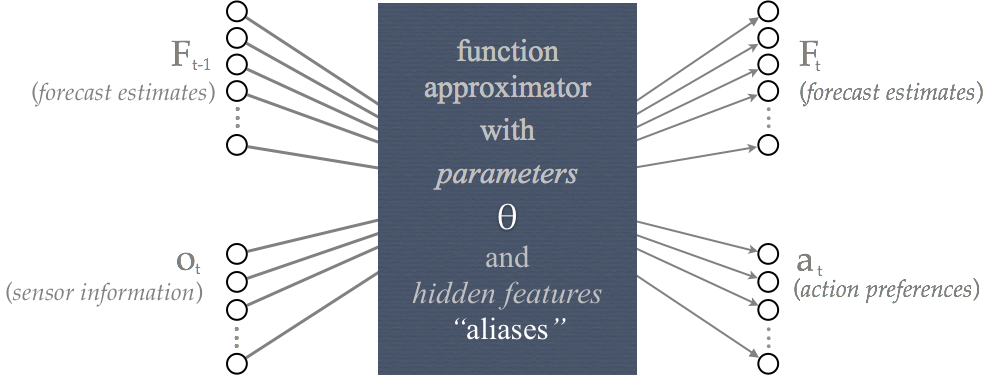}
  \caption{The thought experiment imagines a perfect feedforward function approximator: one that discovers any (non-temporal) statistical regularities in its input-target mapping.
In doing so, it may discover any number of hidden features (or ``aliases'') that can be used to assist estimation of any forecast values.}
\label{fig:FA}
\end{figure}

The fictional, perfect, feedforward function approximator mentioned above is shown in Figure~\ref{fig:FA}.
It takes as input the current sensor observations and forecast estimates from the previous time step.
Using these and adjustable parameters $\Theta$, it generates values for all forecasts and action preferences at the next time step.%
\footnote{\label{fn:policy-actions} If the agent in this demonstration were fully autonomous and \IA were not designing the forecast definitions by hand, it would also require a mechanism for choosing actions, and, while this would be a fairly minor extension, it is not required for the thought experiment, and so \IA leave out its description.}
Note that any information required for one forecast estimate can in principle be used for any other.
Thus, one may postulate internal hidden variables or ``aliases''—--simple functions of the inputs, that might serve as building blocks, component features useful for the estimation of multiple different forecasts. (Some examples are given later in the text.) Parameters are adjusted using a temporal-difference method such as described in Section~\ref{sec:framework}, with the forecast values at the following time step as targets.

The challenge of this demonstration is to show how high-level knowledge can be built from the agent's raw sensorimotor apparatus.
Thus, the agent has no initial knowledge and no built-in interpretation of its sensors.
The order of the camera's inputs (pixels) is randomized so that the physical structure of the sensor is not conveyed by the arrangement of the inputs.
The agent does not know that the these values represent visual information.
It does not even know that its two sensors are distinct.
It has no knowledge regarding the purpose or effects of its motor signals, or any knowledge of the relationships between its senses and actions.
In other words, before learning, it experiences the ``blooming, buzzing confusion'' of its interface to its world, just as William James described the baby's initial experiences (Section~\ref{sec:bloomingbaby}).

But more specifically, the challenge of the demonstration is to show how the robot can build up an understanding of its world from these raw components using only forecasts, options and a feedforward function approximator, building layer by layer, in increasing levels of abstraction.
And the purpose of this effort is to investigate the hypothesis that forecasts provide an isolaminar mechanism for building layered knowledge from the sensorimotor stream, able to capture ever greater complexity, detail, and abstraction.

Thus, the focus of the thought experiment is on \emph{depth} of abstraction, on the plausibility of higher-level knowledge and abstractions using forecasts.
To convey this depth, all forecasts described here are predictions (generally nested predictions about predictions) regarding the activity of a single sensor: the touch sensor.
The agent will come to understand its environment in great depth based on predictions it makes about how it can behave so as to achieve a signal of 1 from this sensor.
It should be understood that additional \emph{breadth} can be added to the forecasts by including predictions about other sensory modalities, whether visual or through other added sensors.

Beginning at layer one, the thought experiment builds up layers of abstraction that recognize, predict and utilize prominent regularities of the sensorimotor stream, eventually arriving at knowledge of the world consisting of walls, rooms of different sizes, doorways and houses, all of these regularities ultimately captured exclusively through the raw sensorimotor stream.
\IA believe this is a sufficiently challenging endeavor that it is nearly inconceivable using any other currently existing mechanism.

\newlayer

The first forecast simply predicts the value of the touch sensor if the finger is extended.
Table~\ref{table:FT} summarizes this forecast, abbreviated \FT.\footnote{This section introduces a considerable number of tables describing a layered set of forecasts, options, and aliases, which are generally referred to by their abbreviations.
For ease of reference, all abbreviations presented in this section are summarized and indexed in Appendix~\ref{sec:collected-tables}.}
All forecasts can be thought of in two different ways: formally and informally.
Formally, there is the exact definition, which is technically precise and correct: this forecast estimates the sum of \accv\ and \termv\ values from the current time step until termination of the \option{extend finger} option.
That option terminates immediately, so the effect is simply to estimate what the value of the touch sensor would be at the next time step if the agent were to choose option~\Onef.
But one can also think of the forecast in somewhat less precise yet more convenient informal terms: \FT\ estimates the probability that the robot will touch something if it extends its finger right now.
All forecast descriptions below provide both formal and informal perspectives; the informal perspective will become increasingly convenient as the layers increase in abstraction.

\TableF{T}

Is the forecast learnable?
The camera, as described above, must be good enough that it can distinguish the image of the wall when the robot is touching it at its base from when the robot is not touching it or when the angle to the wall is too great.
Therefore, there is sufficient information in the camera image alone to allow the agent to distinguish these two cases and learn a good estimate of the forecast value.

Can the agent verify that the forecast is correct?
Yes, because the forecast is described exclusively in terms of the sensorimotor stream, the agent can extend its finger at any time step it chooses and compare the predicted value of the touch sensor with its forecasted value.

Is forecast~\FnT\ high-level and abstract? Like the first rung of a ladder, it does not seem to be very high, but this first forecast is essential for later stages.

Is it cheating to choose in advance to predict the touch sensor given that the agent cannot tell one sensor signal from another?
No, the task at the moment is to build knowledge by hand, and in \myA role as designer and demonstrator, it is fair to bring any of \myA own knowledge to bear to show what is possible within the constraints of the system.
The function approximator is not informed that the touch sensor signal is different from any others.
It has no special, prior knowledge as to the source of its input signals but must decipher them and figure out how to use them to improve its predictions.

\newlayer

The next forecast \FTL\ is built using the last.
In fact, it is a prediction of the value of \FT\ after the agent chooses action number \Onrotl, \Orotl.
Table~\ref{table:FTL} gives a full formal description of the forecast.
As with \FT, forecast \FTL\ is based on a one-step option (\ie\ termination occurs immediately), so the forecast estimates the value of the \FT\ forecast at the following time step if the agent were to choose option \Onrotl~(\Orotl~\!) at the current time step.
\emph{Informally} one might think of this forecast as encoding knowledge about whether there is something the agent can touch 30° to its left.

\TableF{TL, TR}

Is the forecast learnable?
If forecast \FnT~(\FT) has already been learned well, then the agent will have a learning target every time it selects option \Orotl (rotate left). 
If the camera can distinguish between the image of a wall 30° to its left and the wall 60° to its left (or more)—a reasonable precondition of the thought experiment—then the agent can learn to identify the situations in which it can rotate to the left and from there correctly predict the value of forecast \FT.
This result makes no great demands on the camera or the function approximator.

Can the agent verify that \FTL\ is correct?
Yes, the meaning of the forecast is: what is the probability that the sensor \ST\ will have a value of 1 if the agent chooses option \Orotl\ followed by option \Oef$ \!$  ?
At any moment the agent can rotate left and compare the actual value of \FT\ with the predicted value.
Since \FT\ is described exclusively in terms of the sensorimotor stream, so is \FTL.

Is forecast~\FnTL~(\FTL) high-level and abstract?
The knowledge of whether something can be touched to the agent's left is something distinctly more informative than the raw sensory information.
It is the agent's first bit of knowledge about its world that is not immediately observed through its senses nor verifiable in a single time step.
Yet it is only the second rung of the ladder.

Table~\ref{table:FTL} also summarizes forecast \FnTR, \FTR, which is the mirror image of \FTL\ $\!\!$—the same as \FTR\ except for a rotation to the right instead of the left.
Informally, it predicts whether there is something the agent can touch 30° to its right.

\newlayer

\begin{figure}
  \begin{minipage}[c]{0.18\textwidth}
    \centering%
    \includegraphics[width=0.9\textwidth]{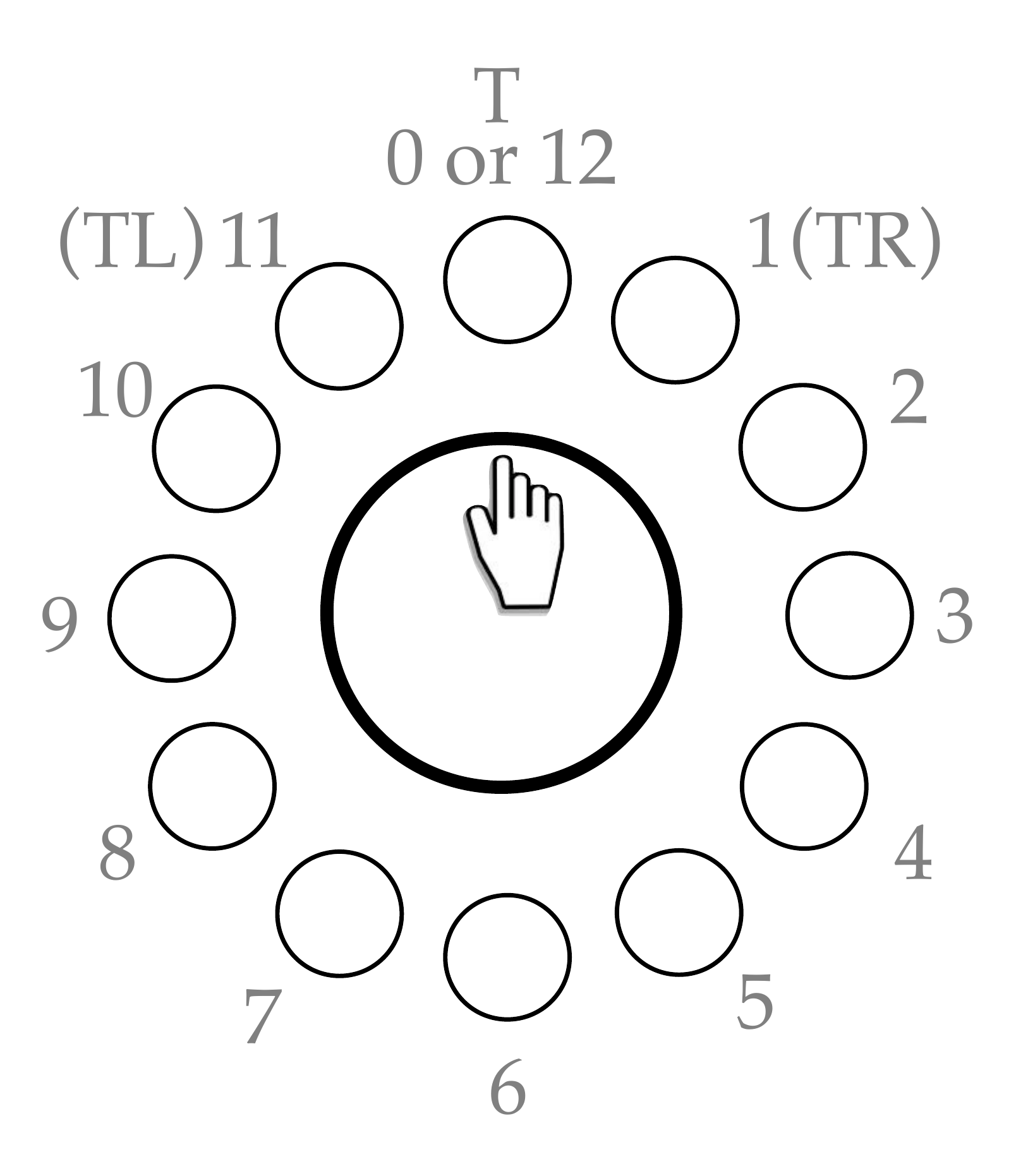} \\
    A
  \end{minipage}
  \begin{minipage}[c]{0.2\textwidth}
    \centering \vspace{20pt}%
    \includegraphics[width=0.99\textwidth]{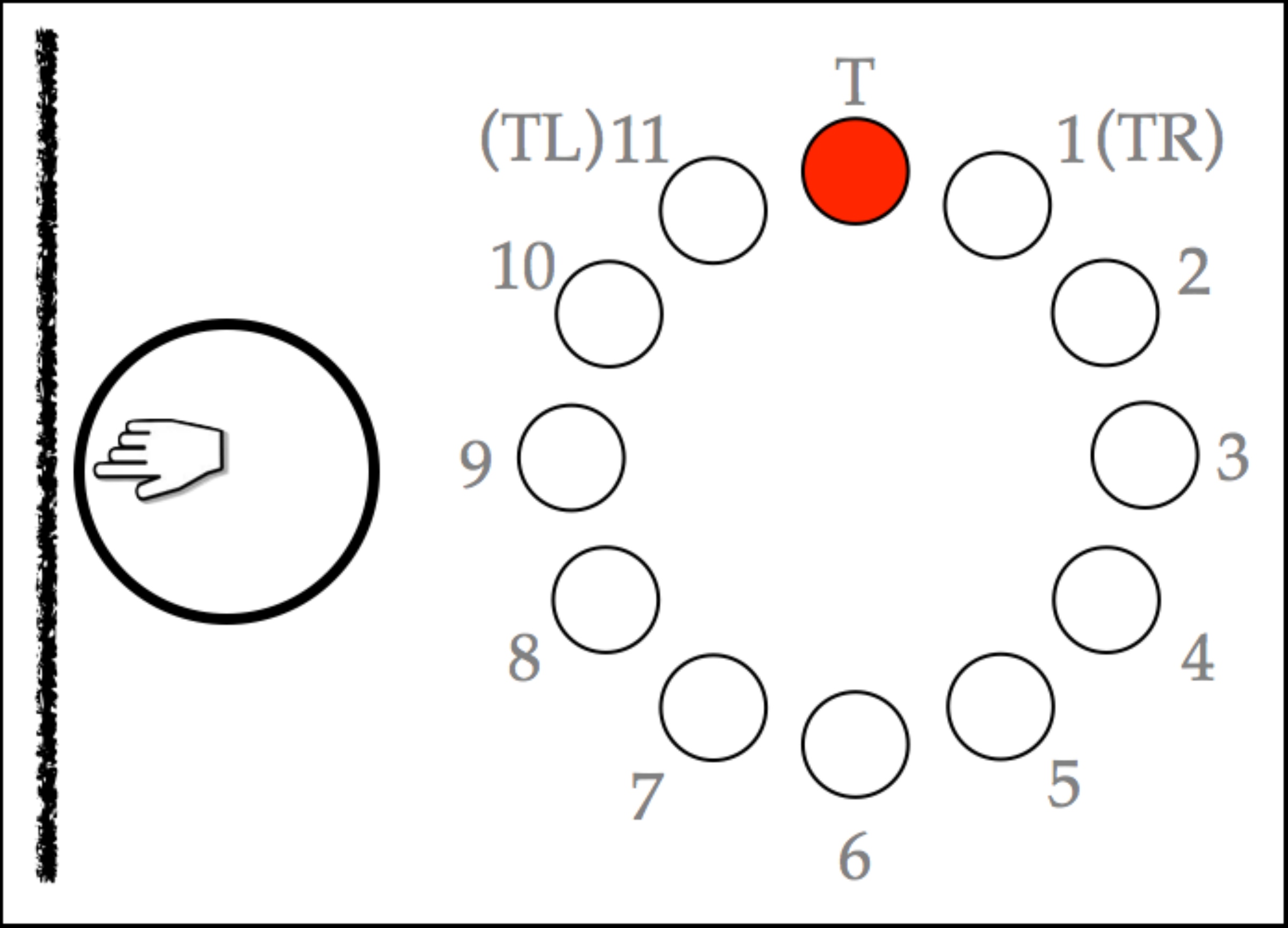} \\
    B
  \end{minipage}
  \begin{minipage}[c]{0.2\textwidth}
    \centering \vspace{20pt}%
    \includegraphics[width=0.99\textwidth]{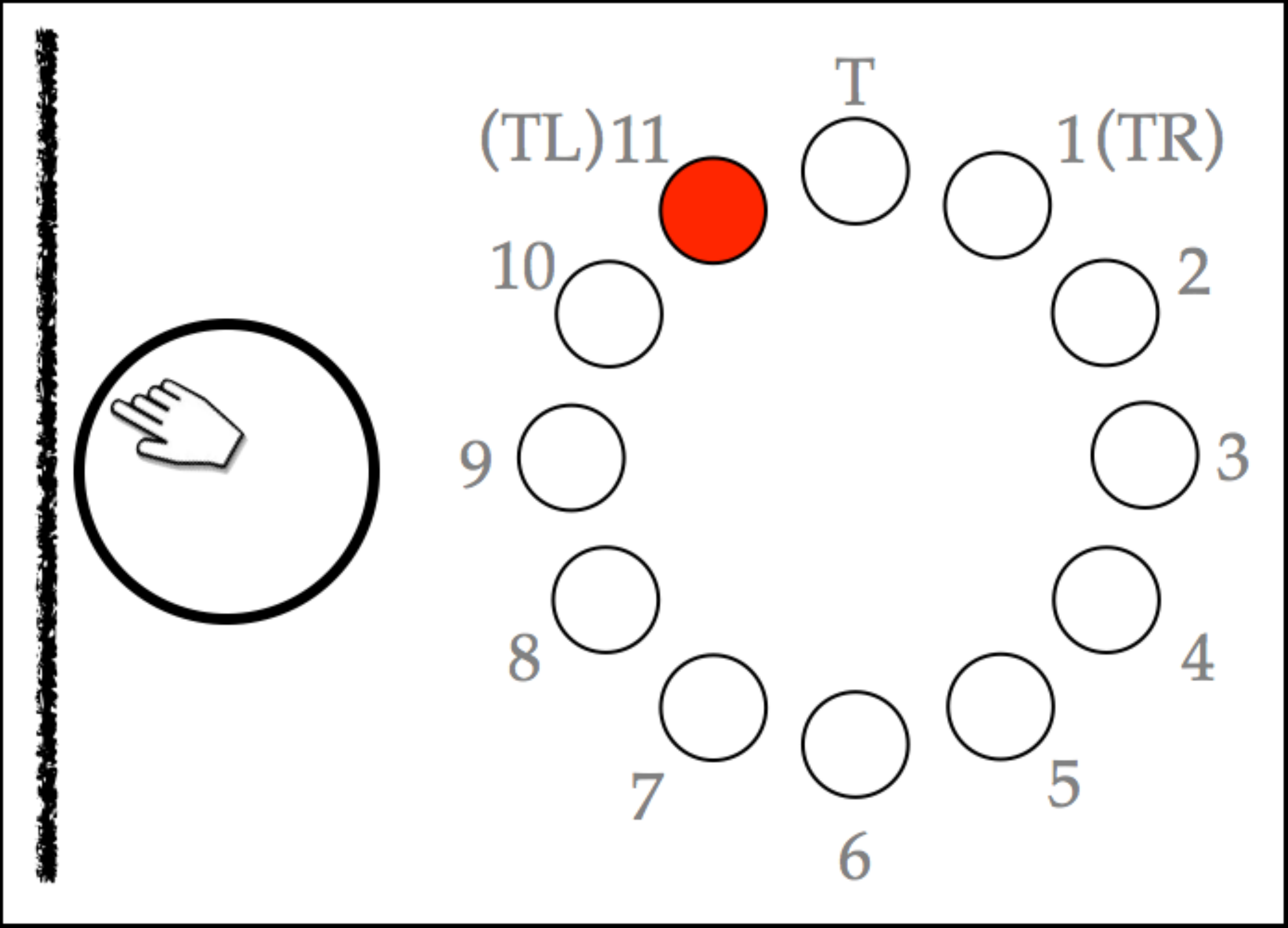} \\
    C
  \end{minipage}
  \begin{minipage}[c]{0.2\textwidth}
    \centering \vspace{20pt}%
    \includegraphics[width=0.99\textwidth]{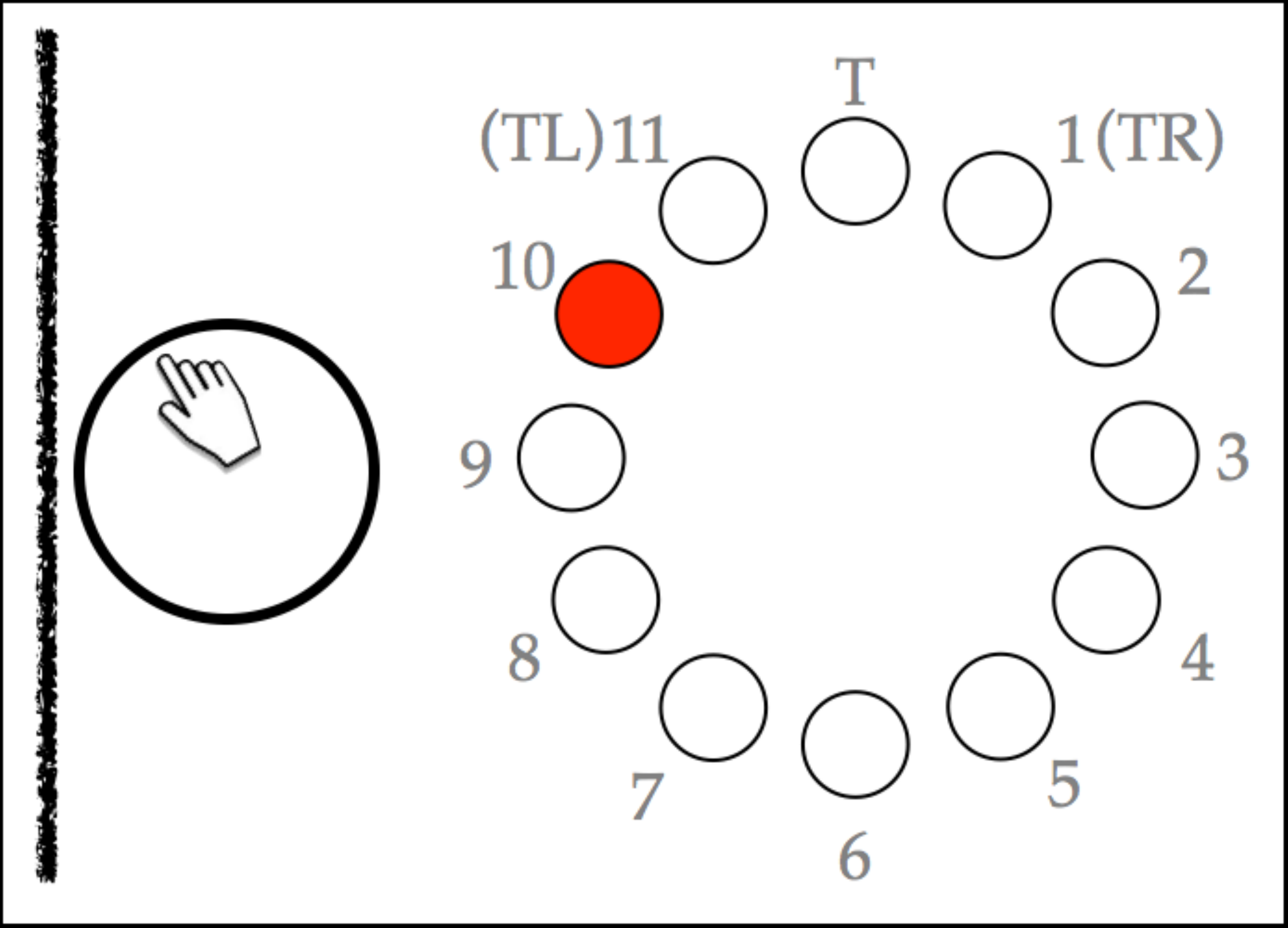} \\
    D
  \end{minipage}
  \begin{minipage}[c]{0.2\textwidth}
    \centering \vspace{20pt}%
    \includegraphics[width=0.99\textwidth]{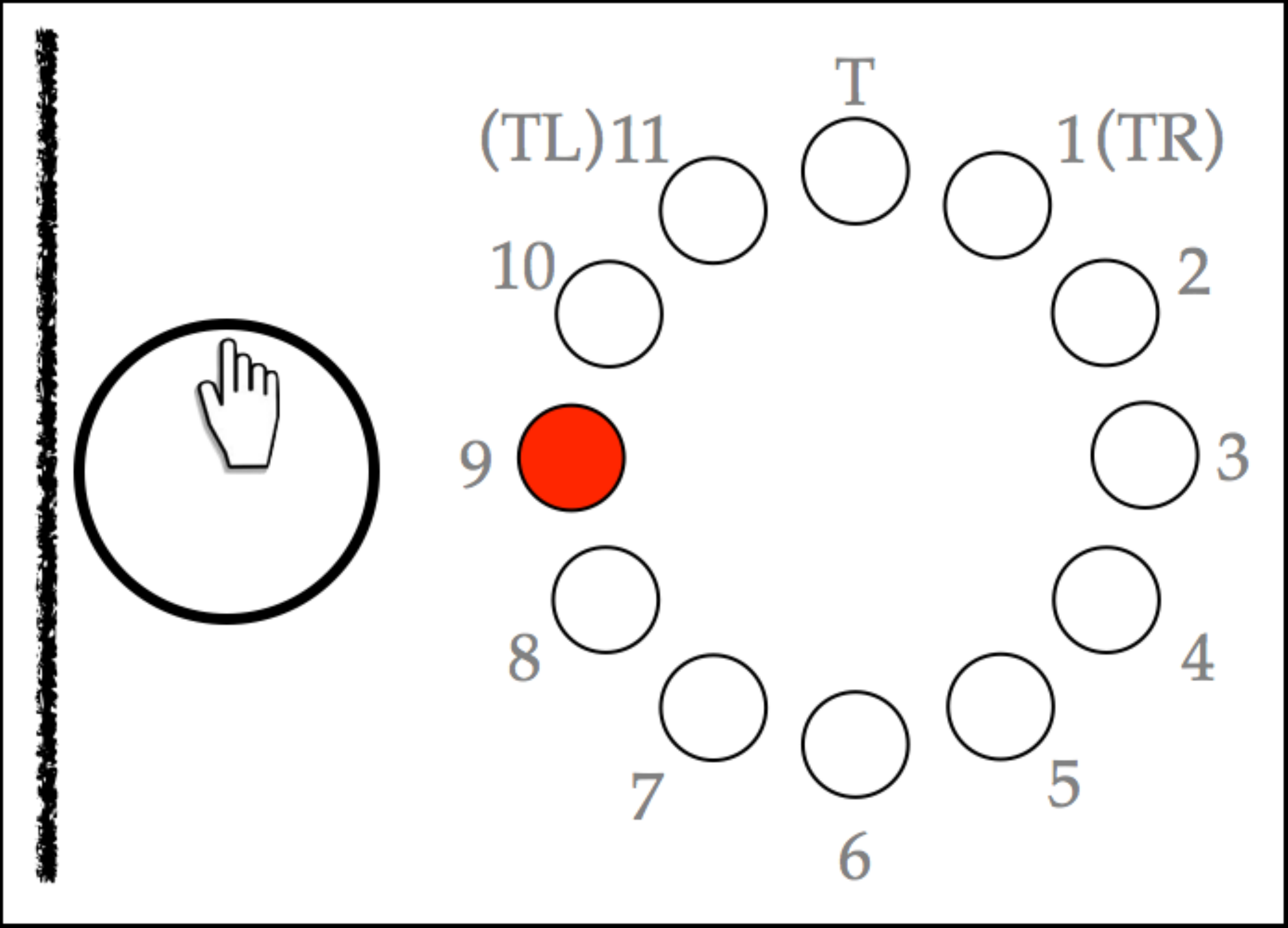} \\
    E
  \end{minipage}
  \caption{ {\bf A:} The twelve \FTMname~(\FTM) forecasts: described formally in Table~\ref{table:FTM}, informally they can be thought of as laid out in a circular ring around the agent like the numbers of a clock, where each estimates whether there is something in the corresponding position that the agent could rotate to, extend its finger toward, and touch.
    {\bf B:} Because the agent can see that it is facing a barrier (black line), the \FT\ forecast predicts that the touch sensor \ST\ will return a value of 1 if the agent extends its finger (red = 1, blank~=~0).
    {\bf C-E: }As the agent rotates to the right (clockwise), its \FTMname\ estimates are updated to reflect its newest predictions.
    In box \textbf{C}, the \FTM(11) forecast predicts that if the agent rotates left, then the \FT\ forecast will have a value of 1.
    In box \textbf{D}, the \FTM(10) forecast predicts that if the agent rotates left, then the \FTM(11) forecast will have a value of 1.
    In box \textbf{E}, the \FTM(9) forecast predicts that if the agent rotates left, then the \FTM(10) forecast will have a value of 1.
  }
\label{fig:TM}
\end{figure}

The next layer extends the last.
In truth, this layer is in itself a small series of layers, each built on the previous.
Forecasts \FnTL\ and \FnTR\ (\FTL\ and \FTR) make predictions about the
value of \FT\ if the robot makes a turn to the left or right, respectively, and this same idea can be replicated in a full circle such that the agent makes predictions about whether it can expect to touch something after some number of rotations to the left or right.
\IA call this the \FTMname.
It is summarized in Table~\ref{table:FTM} and illustrated in Figure~\ref{fig:TM}.
In total there are 12 forecasts, \FTM$(0)$--$\FTM(12)$, one for each 30° rotation, where \FTM$(0) \equiv \FTM(12) \equiv \FT$.

\TableF{TM}

Each forecast \FTM$(1)$--$\FTM(6)$ is a prediction about the forecast numbered below it should the agent rotate once to the right (clockwise).
Each forecast \FTM$(7)$--$\FTM(11)$ is a prediction about the forecast numbered above it should the agent rotate once to the left (counterclockwise).
Thus \FTM$(1)$ is the same function as \FTR, and \FTM$(11)$ is the same as \FTL.

Informally, one can think of these forecasts as being laid out like the numbers of a clock surrounding the agent, each capturing the agent's estimate as to whether there is something to touch at that clock location (see Figure~\ref{fig:TM}A).

Can the agent verify that the forecast values are correct?
Yes, each is described exclusively in terms of the sensorimotor stream: each specifies what the agent can expect to experience after following a policy for a single time step.
Together, this set of forecasts gives the agent a powerful representation of its immediate vicinity.
Each forecasted value can inform the estimation of the others.
For example, imagine that the agent faces a barrier, and the \FT\ forecast accurately predicts that the touch signal will be 1 if the agent should select action \Oef.
If the robot chooses action \Onrotr~(\Orotr) and rotates clockwise, then the function approximator can use the \FT\ estimate from the last time step to correctly estimate the value of \FTM(11) at the next, even in the absence of unambiguous visual data.
Similarly,  a subsequent \Orotr\ action can inform the estimate for \FTM(10).
In the other cases as well, the function approximator can update the values of each forecast estimate using the previous values of its neighbors.

Once the forecast values are being estimated relatively well, they provide reliable signals for the estimation and verification of other forecasts, even in the absence of visual information; but can they be learned initially?
In particular, how can the values of the middle forecasts \FTM(6) and \FTM(7) be learned when the robot is facing away from the barrier?
The values are first learned for \FTM(0) and its neighbors, \FTM(1) and \FTM(11), as described in the previous layer.
These values can then serve as targets for their neighboring forecasts, \FTM(2) and \FTM(10), \etc\ Now, if the robot is situated next to a barrier in a place where visual information is unambiguous (where the robot sees a unique visual image at every rotational position), then the robot can rotate in place and learn to use its visual information to predict the correct values of all the \FTM\ forecasts.
However, as soon as the robot moves to a new place with different visual inputs, then the function approximator will no longer generate the correct predictions; but if the agent finds multiple places in the environment where it can repeat this procedure, then it has a way to generalize beyond the visual images and learn to make its predictions based on the current \FTMname\ values.
At that point the robot can correctly predict the \FTM\ forecast values as it rotates in place, even if it is in a location where visual information may be ambiguous, such as in a large room.%
%
\interfootnotelinepenalty=10000
\footnote{It should be acknowledged that the function approximator is facing some hidden demands here.
In particular, it is being asked to generate forecast estimates at one time step that are used to calculate those same forecast estimates at the following time step.
In other words, there is feedback, or recurrence between outputs and inputs, but the function approximator, being feedforward only, is not equipped to recognize possible feedback relationships.
The function approximator must therefore be robust in terms of possible self feedback, though it is not allowed to search for temporal dependencies.
}

Are the new forecasts high-level and abstract?
At this point, the robot has an understanding of its surroundings that are well beyond that of its immediate sensors, yet the representation is not yet significantly different from that of a standard PSR.
It would be premature to conclude at this point that the representation is powerful enough for the goal of creating a vast range of knowledge in an isolaminar fashion.

\newlayer

As this is a thought experiment, where a large amount of experience is as easy to acquire as a small amount, the robot is trained at each layer until its forecast estimates are sufficiently accurate for learning the next layer. 

The next layer introduces the first non-primitive, learned option, \Ortt~(\Orttname) as summarized in Table~\ref{table:Ortt}.
It learns to maximize \accv\ and \termv\ according to Equation~\ref{eq:ideal}.
To do so it must use only rotation actions and finger extension (actions \Onrotl, \Onrotr, and \Onef\ — or \Orotl, \Orotr, and \Oef, respectively),  and avoid all other actions, which have a large negative \accv\ value.\footnote{Note that the \accv\ value, shown as $-\infty$ for purposes of exposition, can be implemented as any sufficiently large negative number for actions other than rotation or finger extension.}
Because \termP\ is never less than 0.1, there is a significant probability that the policy will terminate while \FT\ is still 0, even if the robot is adjacent to a wall.
Thus, the best policy will be the one that most quickly reaches a high value of \FT.

Informally, the learned policy simply describes how to rotate until the agent is facing something it can reach out and touch. 

\TableOCZ{rtt}

Can the option be learned?
Assuming that the agent has learned the values for the touch map, then learning the policy is fairly simple and can be viewed as a very basic RL problem.%
\footnote{As stated in footnote~\ref{fn:policy-actions}, no module is explicitly described here for taking actions, but each policy could in principle be encoded by a single function approximator such as that of Figure~\ref{fig:FA}, which could in that case share computed intermediate features among all options and forecasts.}
Even so, the agent still might not always have the information necessary to make the correct decisions.
If it should back its way toward a wall, it might not have sufficient information to know precisely where it is with respect to the wall.
But if the agent \emph{does} have the information available, option \Ortt\ will indicate the shortest series of rotations to a position where it can reach out and touch something.


\TableF{TA}

A new forecast can now be defined for predicting the value of \FT\ at the termination of \Ortt.
This new forecast definition, \FTA~(\FTAname) is summarized in Table~\ref{table:FTA}.
The forecast value will answer the following question from the agent's perspective: ``Can I cause the \FT\ forecast to have a high value by choosing actions \Onrotl\ and \Onrotr\ only?''
That is, ``can I rotate to a position where I can extend my finger and receive a signal of 1''?

Informally, one can think of \FTA\ as predicting the probability the agent can rotate to a position where it can touch something;
in other words, it informs the agent as to whether there is anything adjacent to it.

\begin{figure}
  \centering
  \begin{minipage}[c]{0.2\textheight}
    \centering%
    \includegraphics[width=0.9\textwidth]{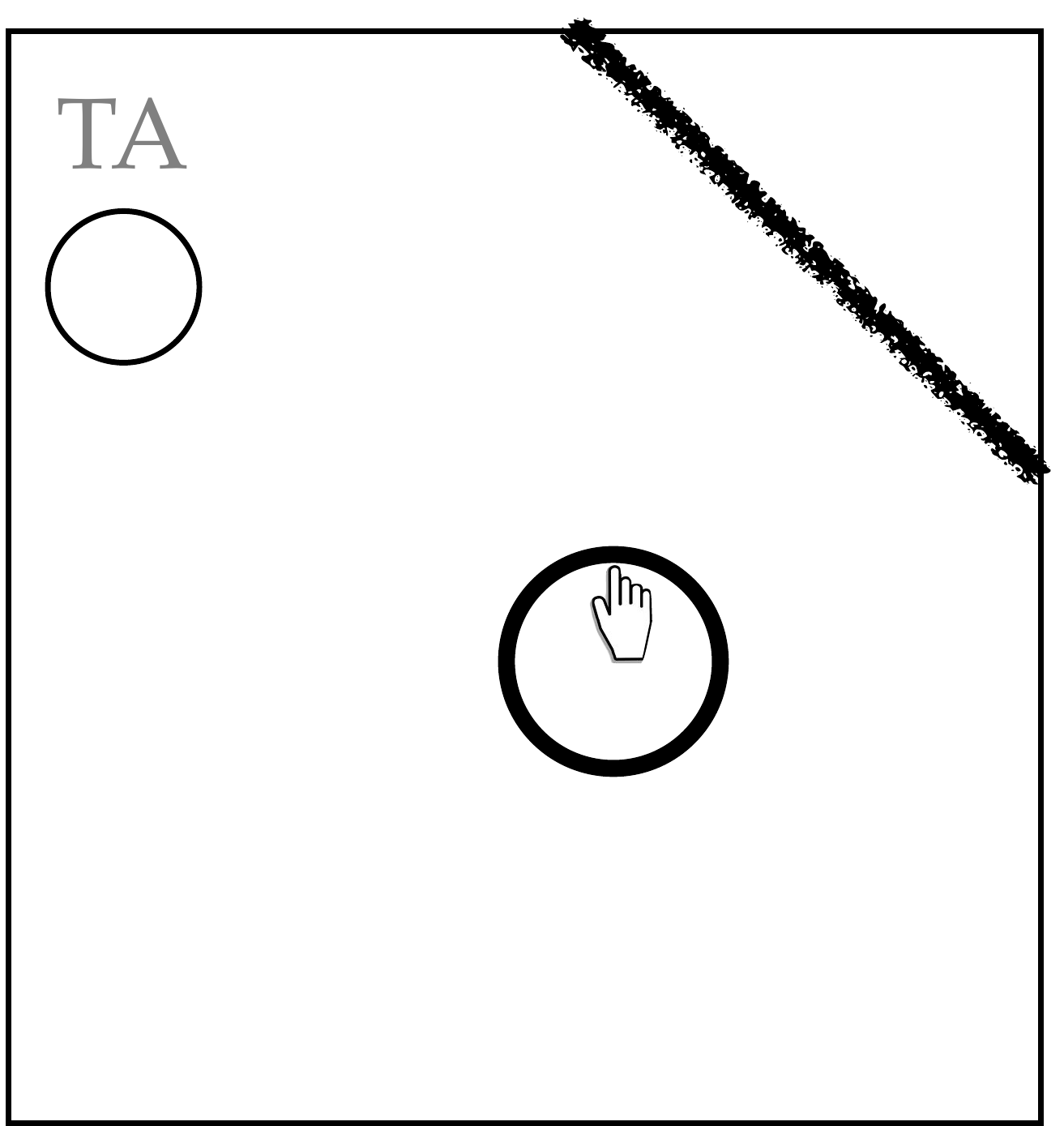}
  \end{minipage}
  \begin{minipage}[c]{0.2\textheight}
    \centering%
    \includegraphics[width=0.9\textwidth]{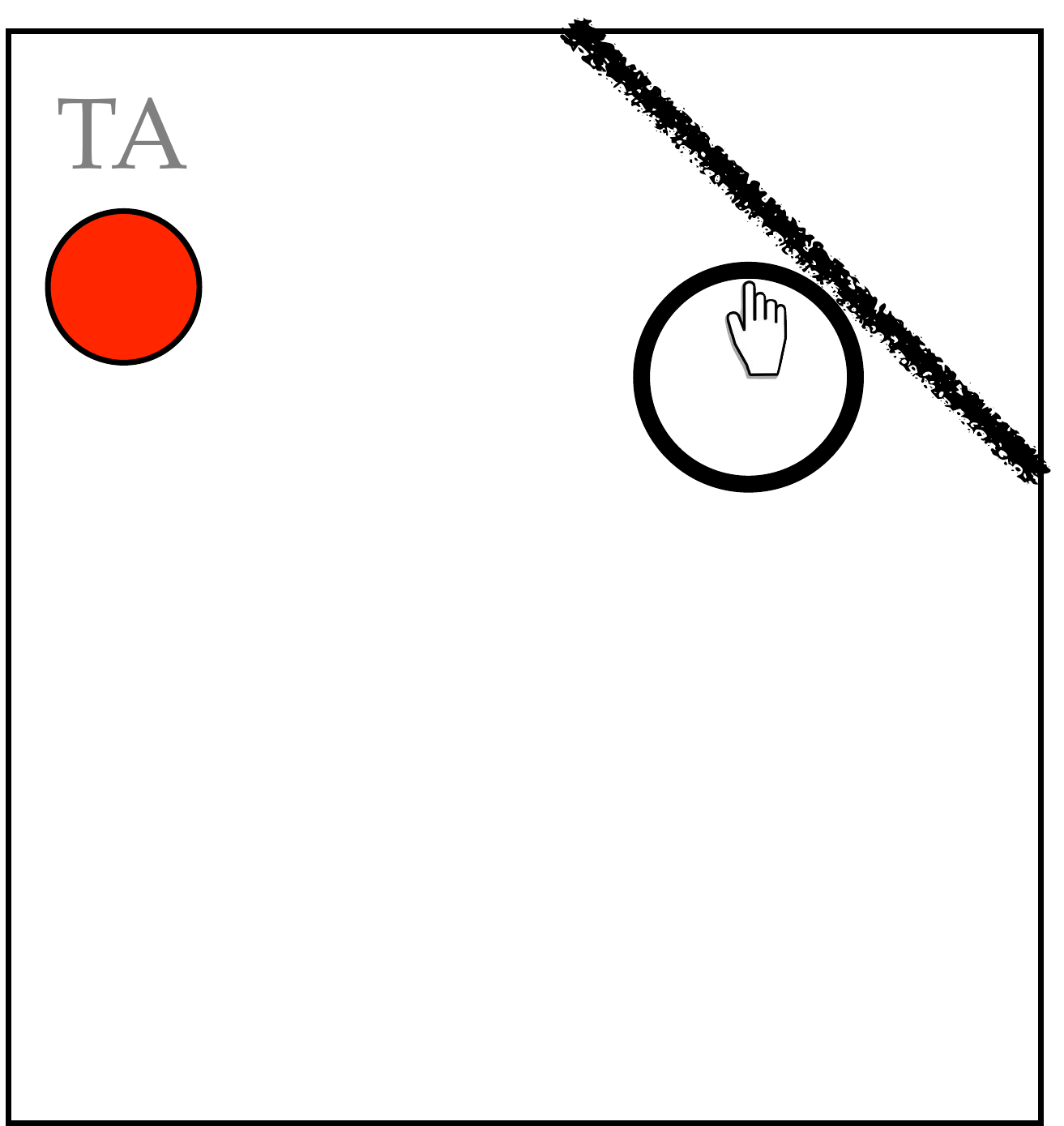}
  \end{minipage}
  \caption{\textbf{(Left)} The agent's \FTA\ forecast estimate (\FnTA) is zero. \textbf{(Right)} The estimate is 1.}
\label{fig:TA}
\end{figure}

The agent can verify the forecasted value by repeatedly following option \Ortt\ until termination, then averaging the resulting value of \FT\ at termination.
In the case where the robot is adjacent to a barrier and has sufficient information from its forecasts and sensors to determine this unambiguously, the \FTA\ forecast will have a high value, because the agent has clear evidence it can follow option \Ortt\ (perhaps multiple times in the case of early termination) until the \FT\ forecast is 1.

Can \FTA\ be learned initially?
Whenever the agent rotates, whether or not it is adjacent to a barrier, it is generating experience that is relevant for training both the \Ortt\ option and the \FTA\ forecast.
(And the off-policy learning methods discussed in Section~\ref{sec:forecast-estimation} allow function approximators to learn from every relevant experience.)
Every time the agent extends its finger, it collects experience that is useful for training \FTA.
If the agent follows the \Ortt\ option (rotates in place) and occasionally extends its finger, it improves its estimate for \FTA.
Assuming it does so a sufficient number of times both when adjacent to a barrier and when not, then the accuracy of its estimates will improve towards the optimal values possible using the information available in the state vector.
If all the \FTMname\ values are correct, then the agent immediately has all the information necessary to predict \FTA\ correctly.
But the \FTMname\ values are not guaranteed to be perfect or complete, only to reflect (under assumption of the perfect function approximator) the agent's full knowledge regarding its opportunities to rotate and touch.
That knowledge is informed by its immediate visual information as well as the \FTMname\ estimates the robot compiled from its visual information and actions as it navigated to its current position and orientation.
If that information was sufficient to estimate good values for the \FTMname, then the agent will be able to use good \FTMname\ estimates to make an equally good prediction for the \FTA\ forecast.
As discussed above, there are situations in which the \FTMname\ values can be perfectly estimated.
In those situations, the agent can also learn to estimate \FTA\ perfectly.
In other situations, the \FTA\ forecast will be as good as the information in the \FTMname.

Is forecast~\FTA\ high-level and abstract?
Under most informal understandings of ``abstract,'' it would seem reasonable to assert that \FTA\ is knowledge significantly more abstract than that captured by the low-level sensors alone.
Whether or not the robot is adjacent to a touchable entity is not information available from just the sensors (for all possible orientations of the robot).
Though \FTA\ appears in many ways to be a statement about the configuration of the world, it is in fact strictly a collection of predictions about the outcomes of possible interactions.
\FTA\ captures something about the robot's relationship with its environment that can be determined in multiple ways and requires multiple time scales, meaning that neither the number of actions nor the amount of time required for verification of the prediction is fixed but depends instead on the intermediate outcomes of the interaction, which are determined by the robot's state.
Thus, the forecast effectively aggregates into a single useful feature a broad range of separate but meaningfully related environmental states.
It would be possible to design a collection of PSR tests of various lengths that represent the same information about the world, and given a mechanism for aggregating that PSR information into a single feature, it would not be misleading to refer to such aggregation as abstraction.

\newlayer

The \FTA\ forecast is an important milestone, and the next layer begins with two new options built upon it (Table~\ref{table:Orfta}).
The first, \Orfta~(\Orftaname), encapsulates a policy for moving forward until termination, where the termination probability is given by the \FTA\ forecast.
Note that $\FTA \in [0,1]$ increases with the agent's certainty (estimated probability) that it can rotate to something it can touch.
But it is not binary.%
\footnote{Even if the agent had perfect knowledge of the world, \FTA\ would not be binary, because there is a chance of early termination before reaching a place where \FT\ is 1.
If the robot can rotate 180° (6 rotations) and then touch something, then if $\termP=0.1$ (Table~\ref{table:FT}) there is a $0.9^6 = 0.53\%$ chance of termination (of option \Ortt) before reaching a state where $\FT=1$.
If \termP\ were instead 0.01, this probability would be $0.99^6=0.94$.
Without perfect knowledge, the forecast represents an expectation over similar states.
If the robot were to roll backward, for example, into a region of its world where it had never been before, it might consistently increase its estimate of \FTA\ (its probability of being able to rotate to touch), despite any direct experience with an obstacle in that area.
With continued experience in its environment, the agent will learn that after rolling forward, its probability of rotating to touch can be estimated from its forward-facing TouchMap values alone $\FTM(10,11,0,1,2)$, which are mostly informed by its visual input.}
%
The \Orfta\ option will simply have a larger chance of terminating when the \FTA\ forecast is high than when it is low.

Informally, the agent following this option will roll forward until (probably) it is adjacent to a wall.
Alternatively, the second option~\Orftt, introduces a threshold $θ$ (0.5, for example) such that the option's termination probability is binary.

\TableO{rfta, rftt}

Two new forecasts, \FDTA\ and \FNTA, can now be built with these new options (Table~\ref{table:FDTA}).
The \FDTA\ forecast is the first to use the \accv\ value of Equation~\ref{eq:ideal}.
It estimates the number of time steps that will elapse if the option \Orfta\ is followed to termination (alternatively, \Orftt\ could also be used as the option).

\TableF{DTA, NTA}

\begin{figure}
  \begin{minipage}[c]{0.24\textwidth}
    \centering%
    \includegraphics[width=0.99\textwidth]{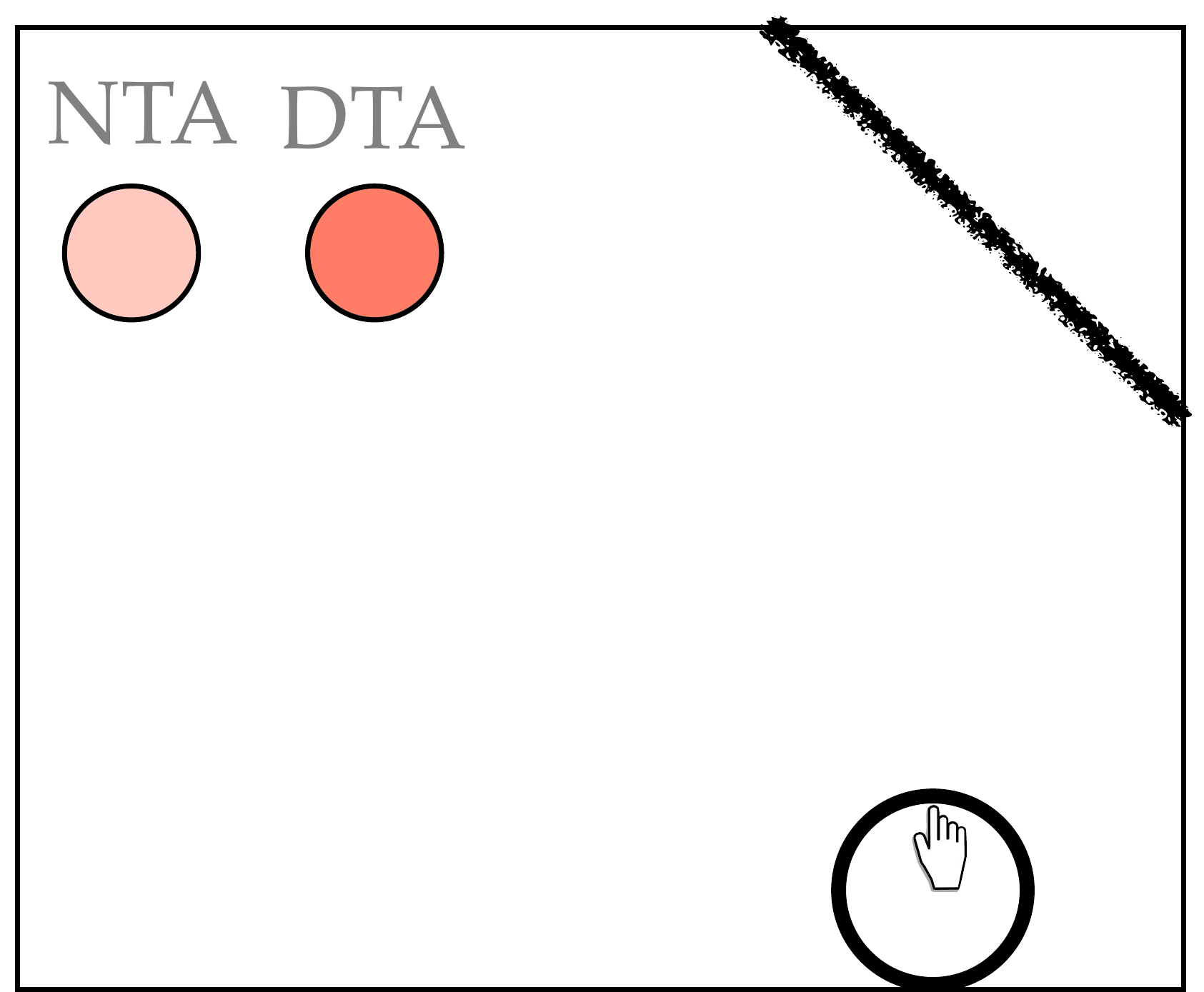}
  \end{minipage}
  \begin{minipage}[c]{0.24\textwidth}
    \centering%
    \includegraphics[width=0.99\textwidth]{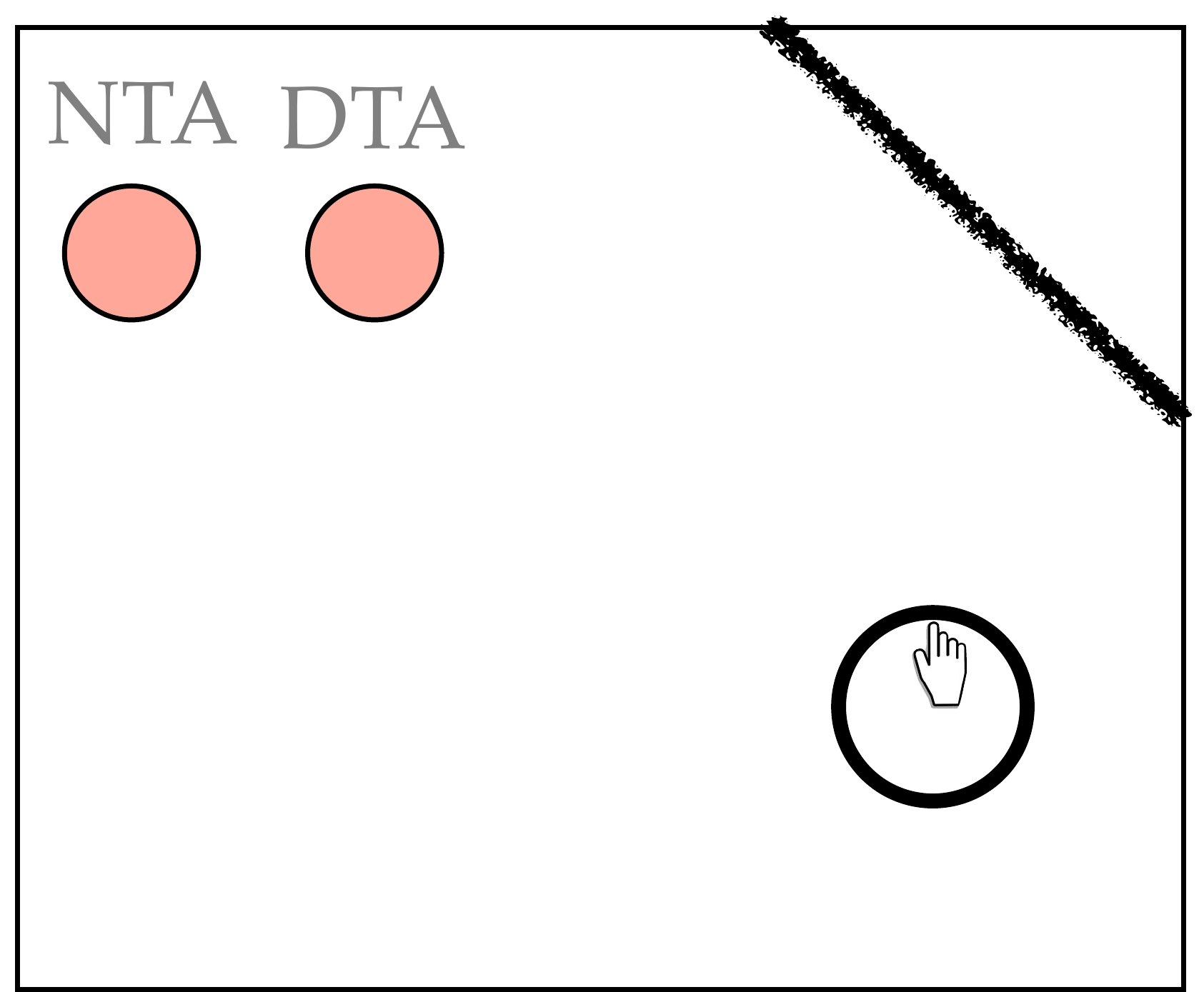}
  \end{minipage}
  \begin{minipage}[c]{0.24\textwidth}
    \centering%
    \includegraphics[width=0.99\textwidth]{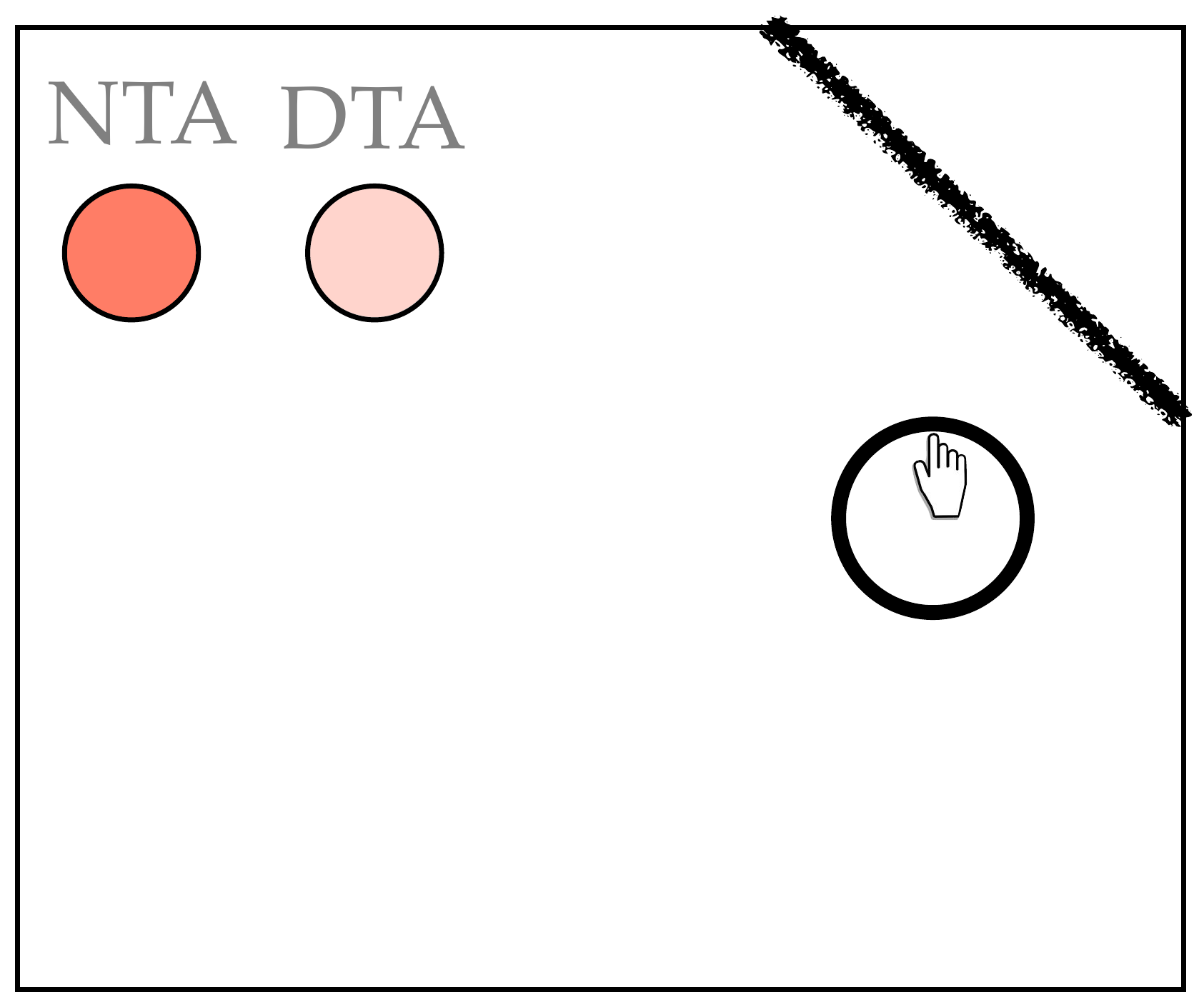}
  \end{minipage}
  \begin{minipage}[c]{0.24\textwidth}
    \centering%
    \includegraphics[width=0.99\textwidth]{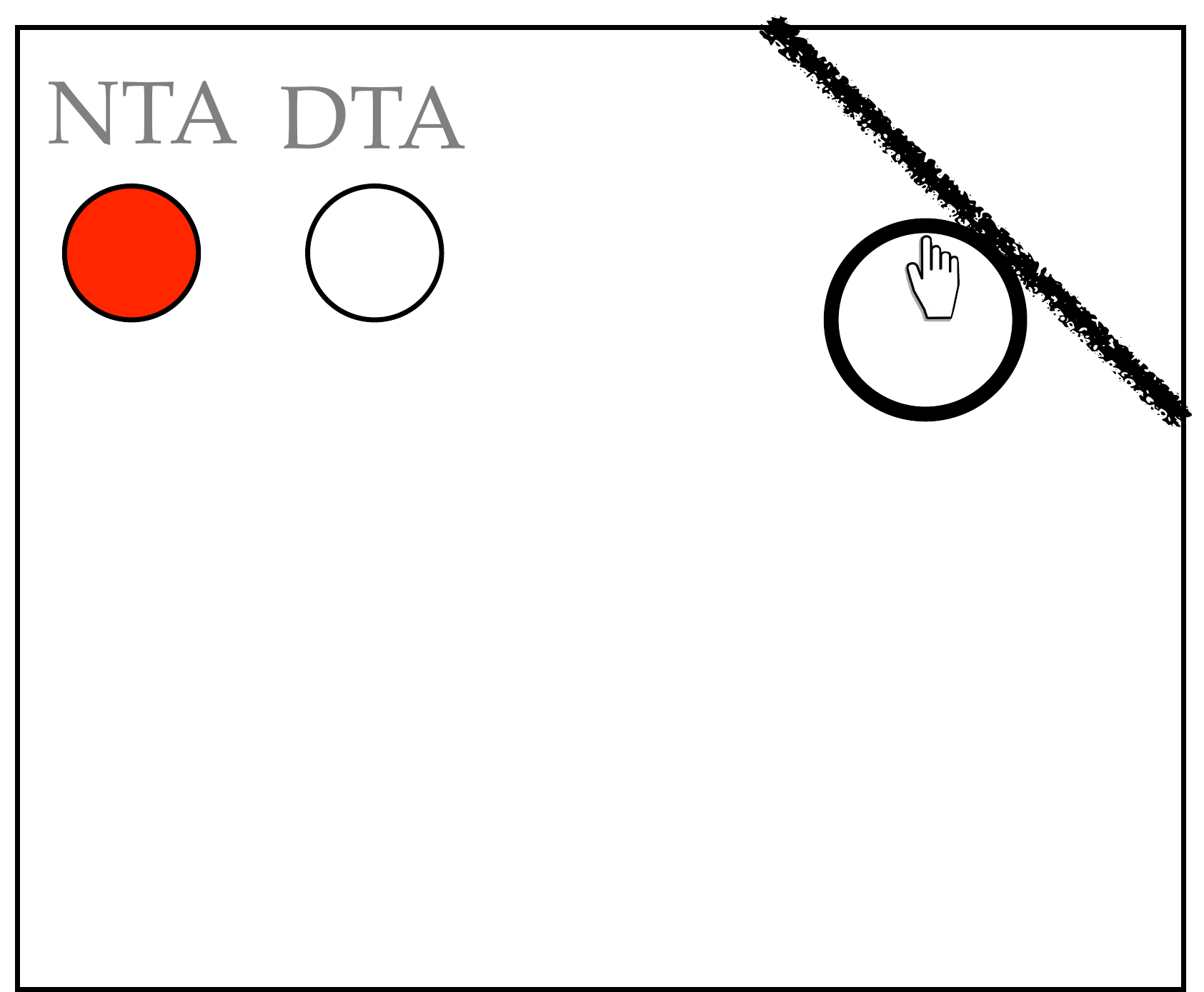}
  \end{minipage}
  \caption{\textbf{Left to right:} As the agent rolls closer to the barrier, its \FDTA\ forecast estimate (\FnDTA)---which predicts how many time steps the agent would need to roll forward to reach a state where \FTA~(\FnTA) is 1---continually decreases, while the inverse estimate \FNTA~(\FnNTA), continually increases. (Darker pink colors represent increasing values where red indicates the maximum and white the minimum values.)}
\label{fig:FDTA}
\end{figure}

Informally, \FDTA\ estimates a subjective notion of forward distance---\ie\ the number of time steps rolling forward to a barrier (see Figure~\ref{fig:FDTA}).
The greater this ``distance,'' the higher the ideal value of the forecast.
Notice that the term ``distance'' here means something very specific, but very subjective.
In fact, it is not specifically a measurement of the environment at all, but a prediction about the agent's future experiences, in particular, its future number of \Orf~(\Orfname) actions before its \FTA\ estimate reaches a high value.
However, from our assumed omniscient point of view as constructors of the thought experiment, it seems so closely related to our own notion of distance that this term, though imprecise, immediately leaps to mind.
To be sure, we could add additional forecasts and capture a notion of distance even more similar to ours, but doing so would not be particularly helpful for the purposes of this experiment.

The forecast \FNTA---the \emph{nearness} of \FTA---is something akin to the \emph{inverse} of \FDTA.
It predicts the probability that termination (while following the \Orfta\ or \Orftt\ option) will occur in a state where the \FTA\ forecast estimate is high.
(Notice that a threshold is used to distinguish between high and low values.)
If the agent will need many \Orf\ actions to reach a state where  $\FTA>θ$, the probability is high that termination will occur while  $\FTA\leq θ$ and $\termv$ is 0, thus \FNTA\ will have a low value.
But when the agent can reach something to touch in only a small number of \Orf\ actions, then there is a high probability that \Orfta\ will terminate when \FTA\ has a high value, and thus \FNTA\ will have a value close to 1.
Therefore, the value of \FNTA\ should increase steadily as the robot rolls forward, until it reaches a state adjacent to something it can touch, in which state the forecast estimate should be 1.

\subheading{Verification and Learnability} The agent can verify that its forecasts are correct by following the \Orfta\ and \Orftt\ options (\ie\ by rolling forward) for one or more time steps.
In fact, whenever the agent rolls forward, it can collect data about the \FDTA\ and \FNTA\ forecasts.
Whenever it rolls into a position where it is adjacent to a wall, it receives information about the \FTA\ forecast.
Whenever it rotates while in a position adjacent to a wall, it receives information about the \FTMname\ forecasts.
And whenever it extends its finger, it receives information about the \FT\ forecast.
It can use its camera to receive visual information that will change as it rolls forward from one time step to the next (as a requirement for the thought experiment, see Section~\ref{sec:a&m}), allowing the function approximator to estimate the number of \Orf\ actions needed to reach a high value of \FTA.
Thus, every piece of this chain is verifiable and learnable using only the forecast mechanism, the function approximator, and the robot's hardware to interact with its world.

\subheading{High-level abstraction} If placed in the middle of any open area in the microworld, the agent can now use its camera and forecasts to estimate the number of \Orf\ actions needed to reach a state where it can rotate and then touch. 
It is a \emph{subjective} estimate based exclusively on its own experience and would be different for a different agent with different hardware (such as different diameter wheels, a noisy control apparatus, \etc).
The estimate will account for visually detectable variations in the environment that might impact the robot's rolling efficiency, such as slippery floors or thick carpet.
An estimate of distance to entities visible in the environment seems to be of a character very different from the primitive senses and actions of Table~\ref{table:s&a} on which they are based.
The primitive senses and actions do not recognize any patterns in the agent's environment (e.g., the things we call ``obstacles'' and ``open spaces'') nor say anything about what they might look like.
They make no estimates of distances nor of any other relationships between the agent and its environment.
Thus, it seems clear that the \FDTA\ and \FNTA\ forecasts do indeed capture something abstract from the sensorimotor stream, and something \emph{useful} as well.

\newlayer

Like the TouchMap of Layer~\FlTM, the forecasts of this new layer will create a map of related predictions.
This \FDTAMname~(\FDTAM) is summarized in Table~\ref{table:FDTAM} and is built on top of the \FDTA\ forecast, where $\FDTAM(0) \equiv \FDTAM(12)  \equiv \FDTA$.
Each forecast in the set (other than \FDTA) predicts what the value of one of its neighbors would be if the agent were to execute a single rotation action.
In this way, the agent can keep track of its \FDTAM\ estimates as it rotates in place.

\TableF{DTAM}

\begin{figure}[tb]
  \begin{minipage}[c]{0.24\textwidth}
    \centering%
    A\\[7pt]
    \includegraphics[width=0.9\textwidth]{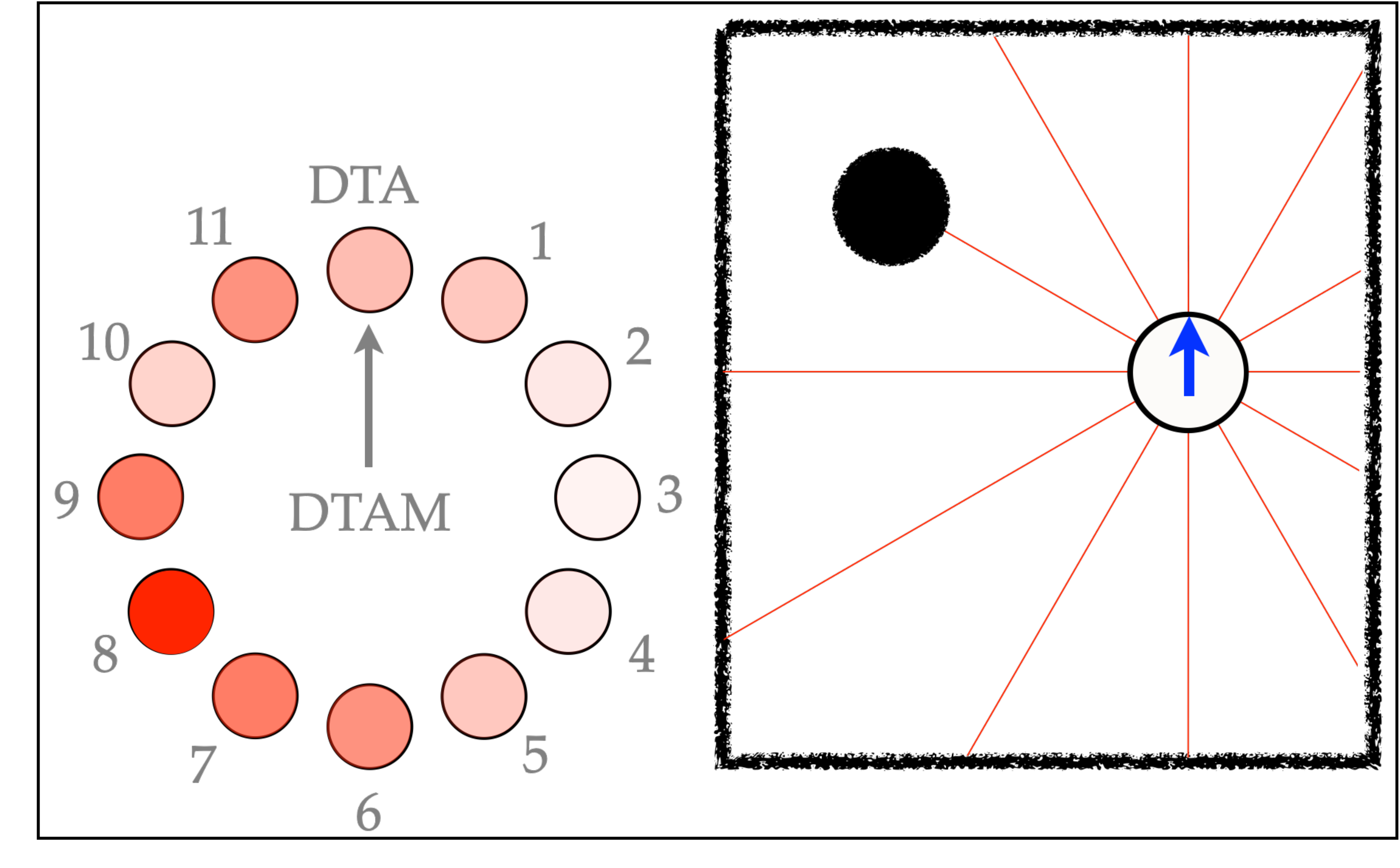}\\[5pt]
    \includegraphics[width=0.9\textwidth]{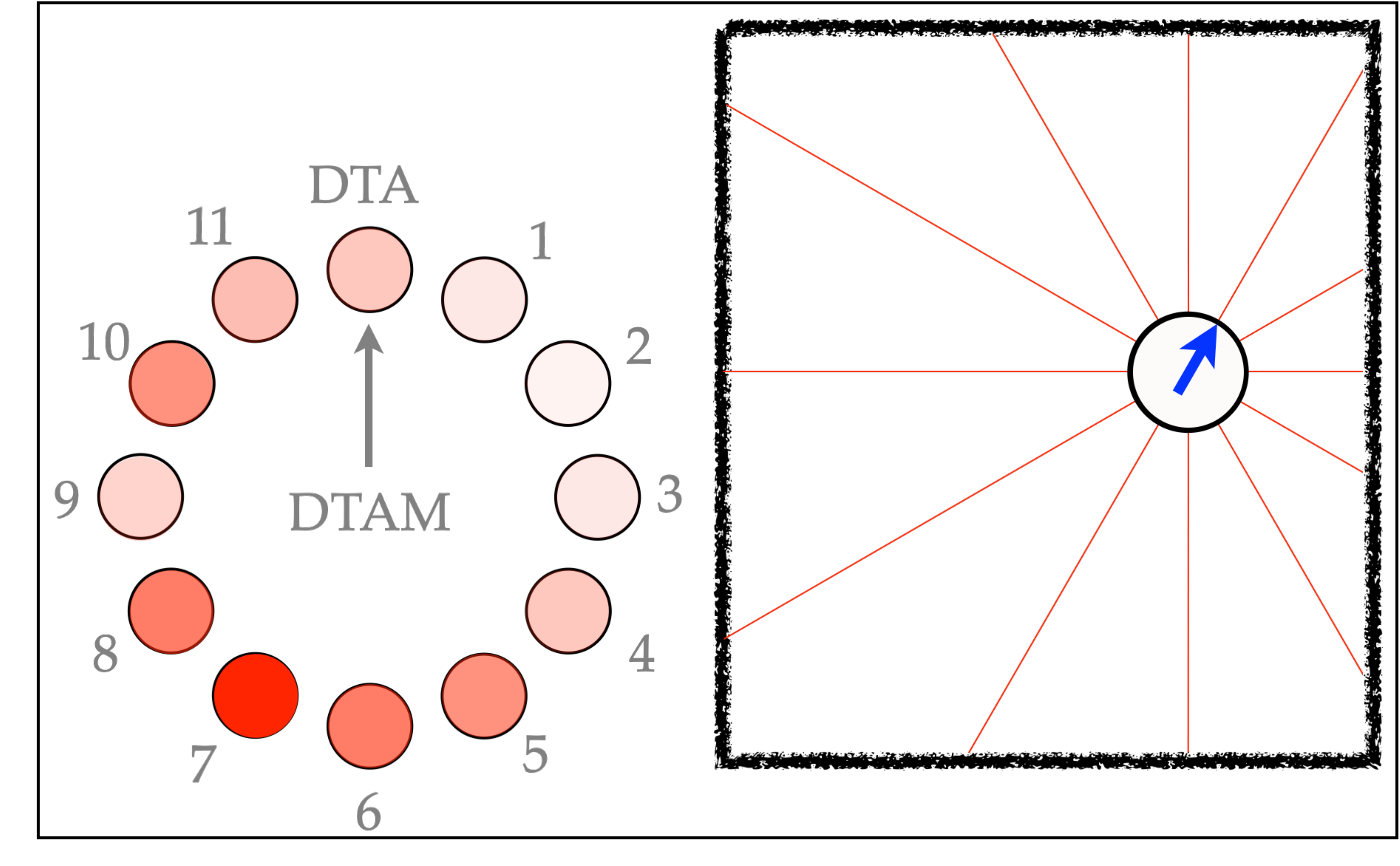}\\
    E\\
  \end{minipage}
  \begin{minipage}[c]{0.24\textwidth}
    \centering%
    B\\[7pt]
    \includegraphics[width=0.9\textwidth]{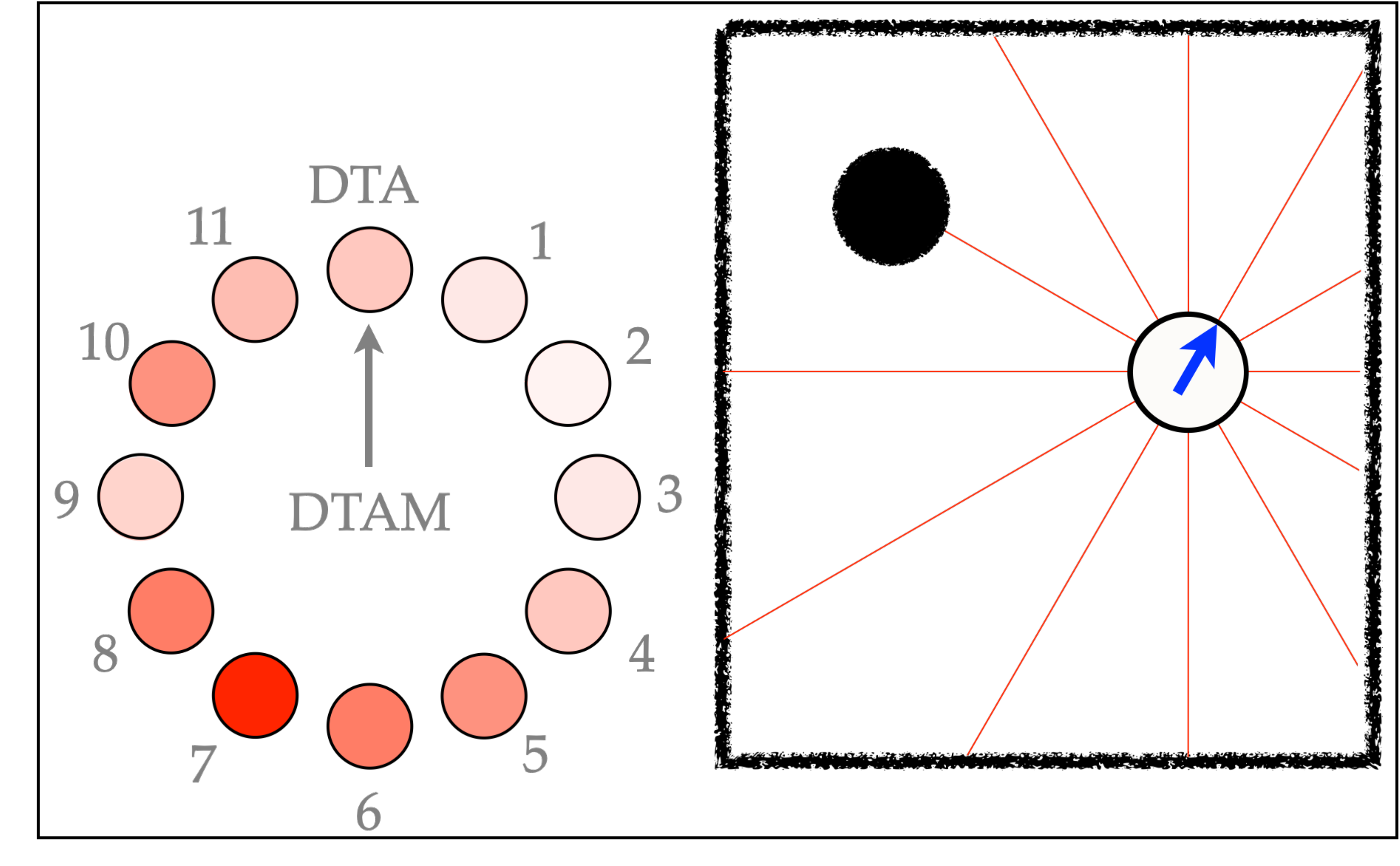}\\[5pt]
    \includegraphics[width=0.9\textwidth]{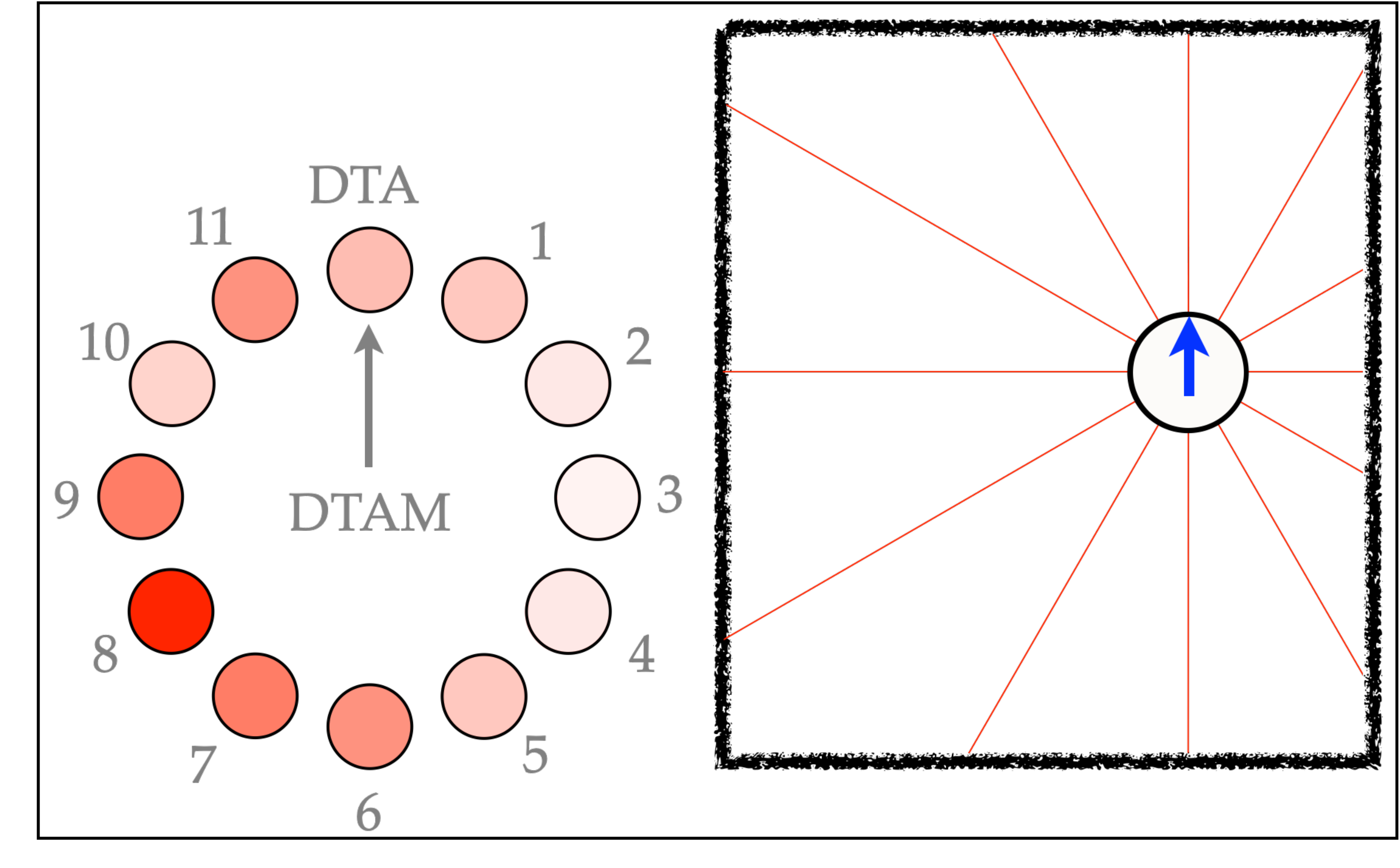}\\
    F\\
  \end{minipage}
  \begin{minipage}[c]{0.24\textwidth}
    \centering%
    C\\[7pt]
    \includegraphics[width=0.9\textwidth]{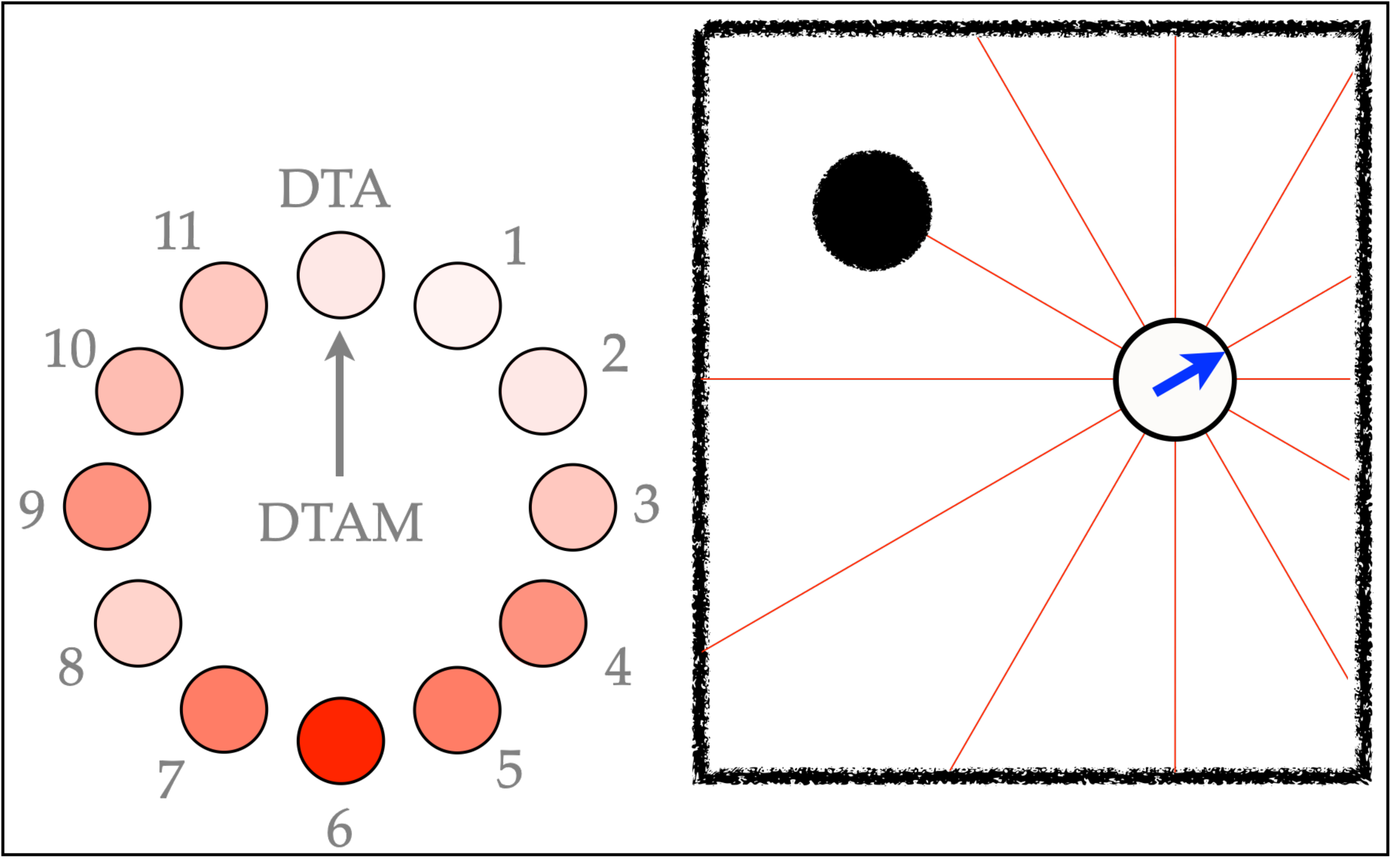}\\[5pt]
    \includegraphics[width=0.9\textwidth]{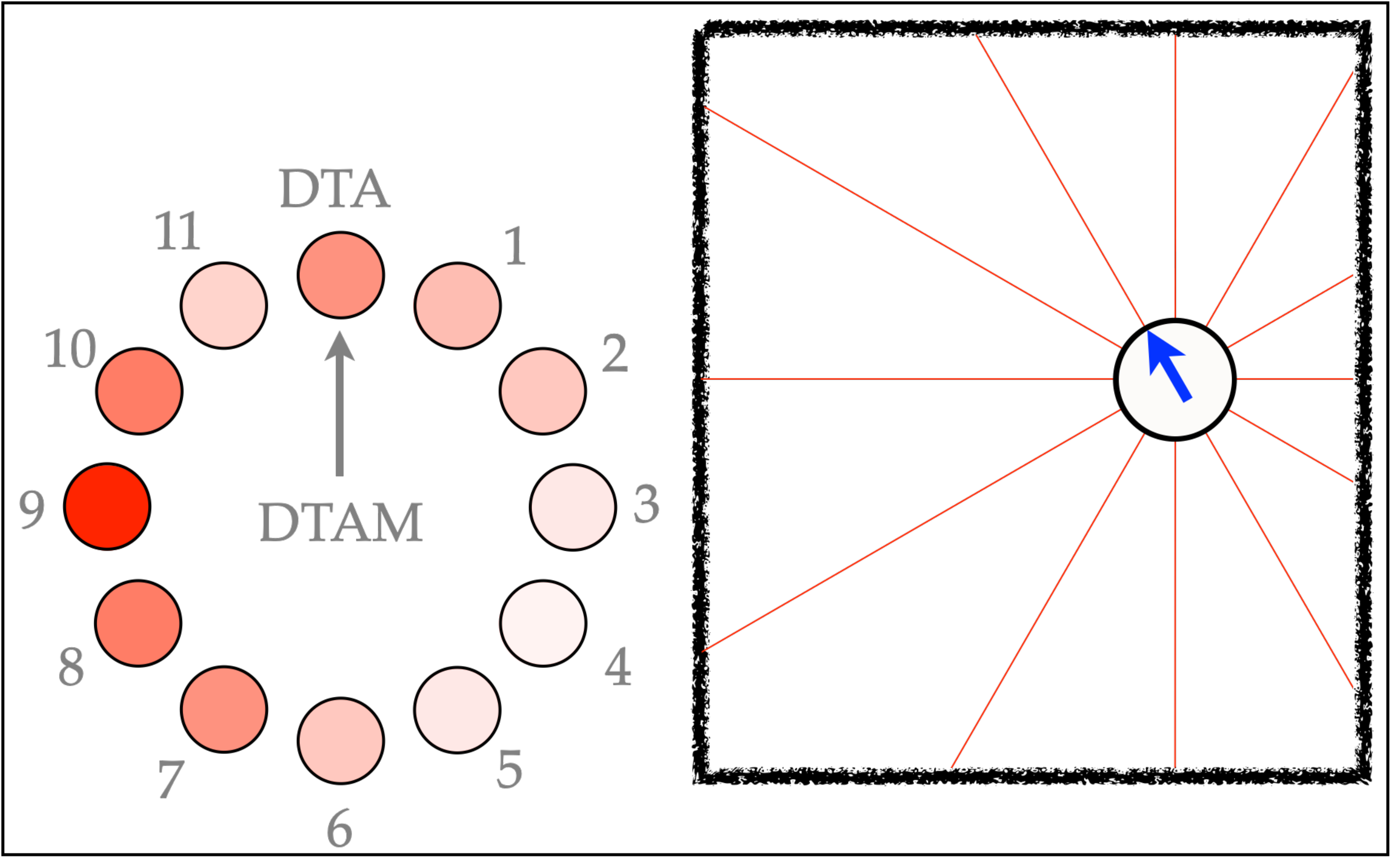}\\
    G\\
  \end{minipage}
  \begin{minipage}[c]{0.24\textwidth}
    \centering%
    D\\[7pt]
    \includegraphics[width=0.9\textwidth]{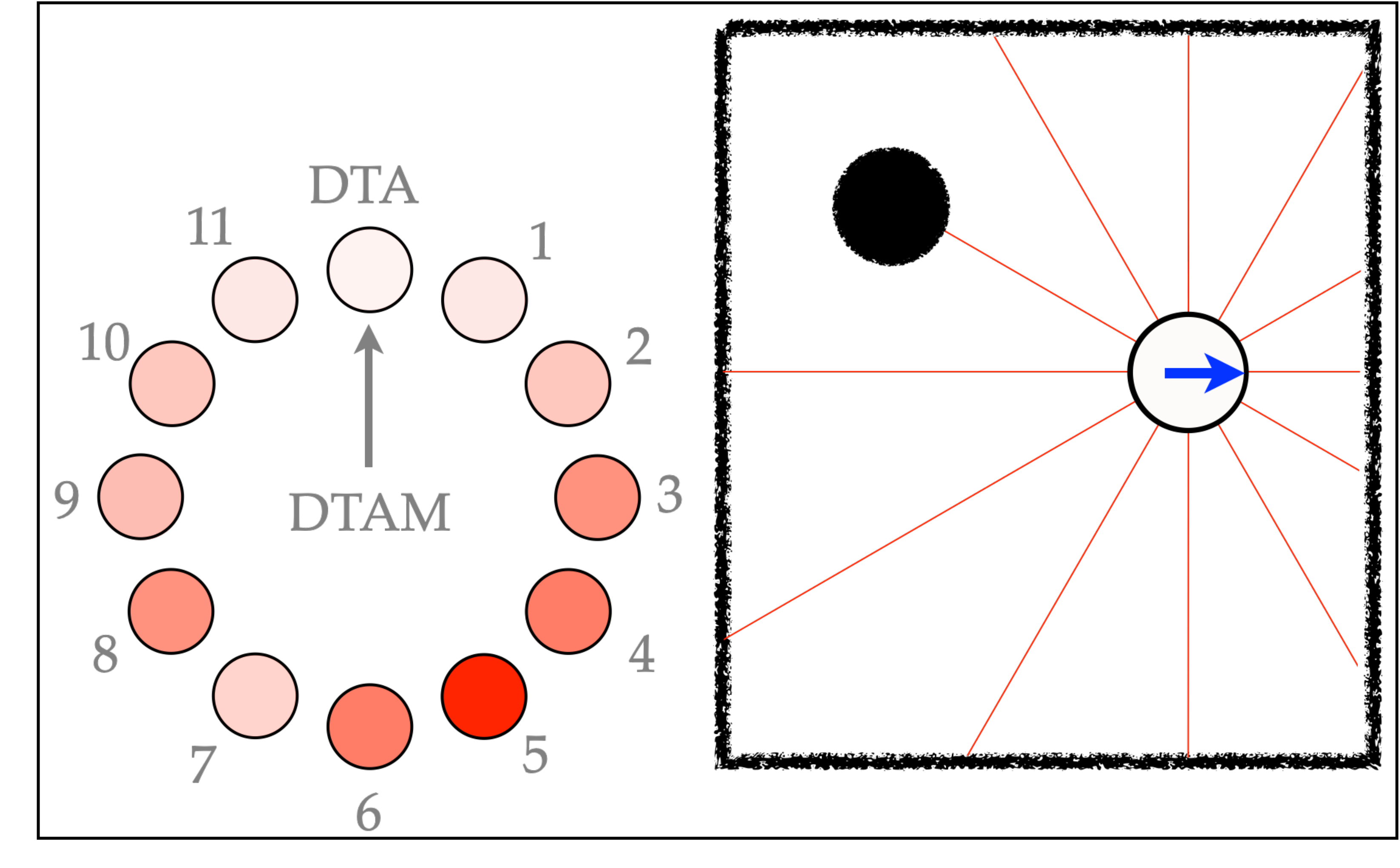}\\[5pt]
    \includegraphics[width=0.9\textwidth]{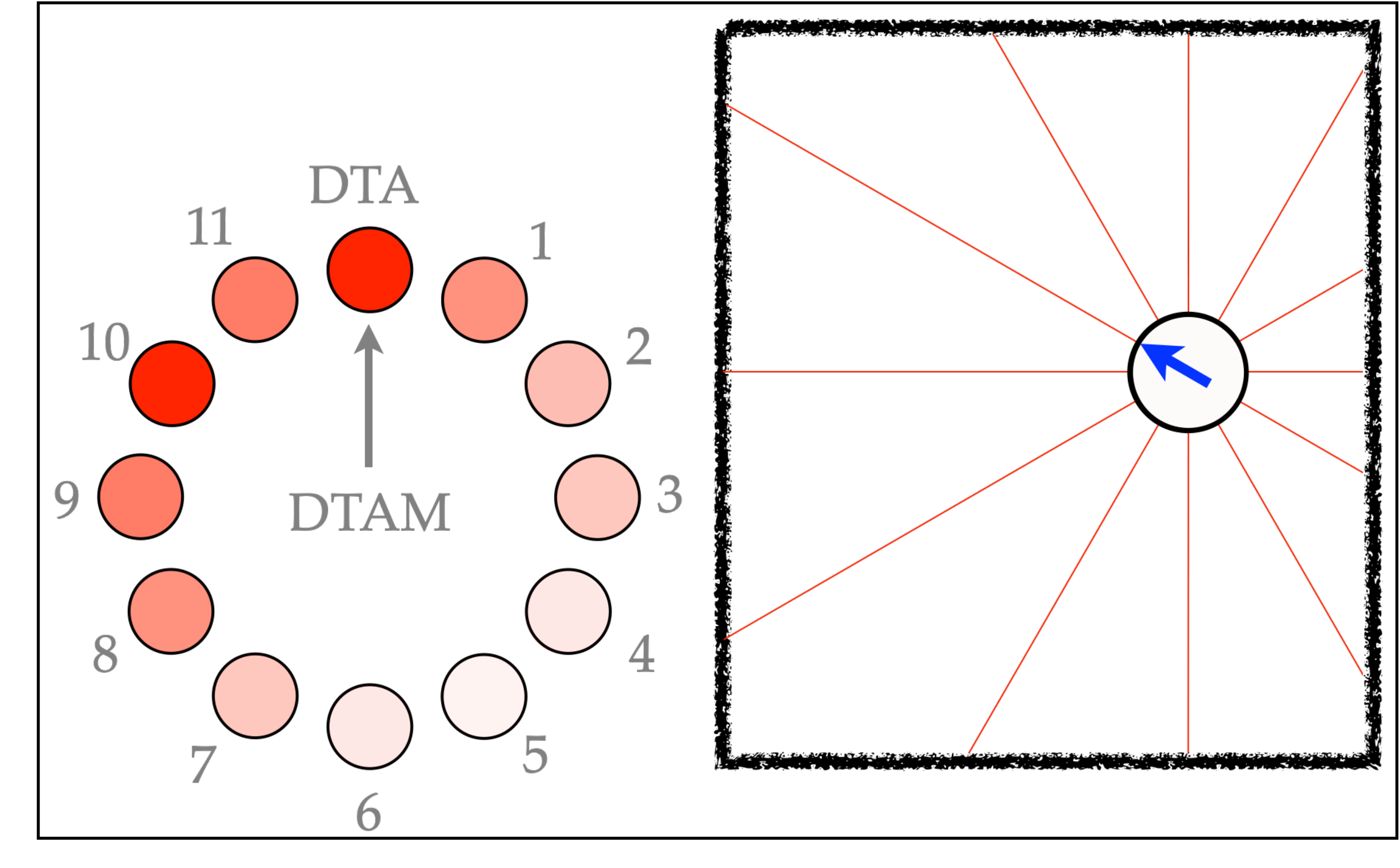}\\
    H\\
  \end{minipage}
  \caption{The agent's \FDTAM\ forecast estimates provide a picture to the agent of how many forward steps (\Orf\ actions) it must take in each direction (red lines) to reach a place where \FTA\ is 1.
As with the \FTM\ forecasts in Figure~\ref{fig:TM}, forecasts \FDTAM(0) through \FDTAM(12) are shown here arranged like a clock: position ``1'' shows forecast \FDTAM(1); position ``2'' shows forecast \FDTAM(2), etc., and
$\FDTAM(0) \equiv \FDTAM(12) \equiv \FDTA$.
(The pointing hand has been replaced by a blue arrow for easier reading.)
\textbf{A:} The agent estimates relatively few steps to the obstacle at its 3 o'clock and 10 o'clock positions, and more steps to the obstacle at its 7, 8, and 9 o'clock positions.
\textbf{B-D}: As the agent rotates, the \FDTAM\ values are updated to reflect the agent's changing estimates.
\textbf{E-G:} If the round obstacle suddenly disappears outside of the agent's visual field, the updates will be made as before.
\textbf{H:} When the agent finally observes the new information (the missing obstacle) it can revise its forecast estimates accordingly.}
\label{fig:FDTAM}
\end{figure}

Informally, the \FDTAM\ forecasts provide the robot an estimate in each clock direction of the distance (how many \Orf\ actions it will need) to reach something it can touch.
Figure~\ref{fig:FDTAM} shows the map of forecasts arranged in a circle surrounding the agent.
This figure also illustrates what occurs when the agent encounters unexpected information, immediately updating its knowledge (its forecast estimates) to incorporate the latest sensorimotor data.

\subheading{Verification and Learnability}
Each forecast in the set gets a target value after each rotation action.
The agent can compare its prediction to the target value and learn from the discrepancy.
The target values are computed by the function approximator from the previous forecasts and the current observation, relying heavily on the visual information to compute $\FDTAM(0) \equiv \FDTAM(12)  \equiv \FDTA$ which is the anchor for the other forecasts.
As with the \FTMname, the forecasts farther in the map from \FDTA\ can be learned after the forecasts closer are learned.
The nature of thought experiments here again allows the assumption that the agent has collected enough experience for the function approximator to achieve sufficient accuracy in the estimate of each forecast before those dependent on it are learned.

\subheading{High-level abstraction}
At this point, the agent has greatly extended its built-in ability to probe the space immediately in front of it, and it now has its first representation of the space moving out from it in all directions.
The \FDTAM\ forecasts give the agent a sense of its surroundings---a sense that describes how far it must travel in each direction to reach something it can touch.
With this sense, it can begin to build a picture of its environment, a picture that it can update at every time step and use to make predictions about what would happen should it rotate or roll in arbitrary directions.
This picture is informed heavily by the visual image from its current position and recent rotations.

\subheading{Low-level foundation}
Yet the entire picture is ultimately built on predictions the agent makes about what it will sense if it extends its finger.
For example, \FDTAM(2) is a prediction about what will happen if it rotates twice to the right (clockwise) then rolls forward until it reaches a state where it can rotate to a position from which it can reach out its finger and receive a signal of 1 from its touch sensor.

\newlayer

\begin{figure}[tb]
  \begin{minipage}[c]{0.15\textwidth}
    \centering%
    \includegraphics[width=0.9\textwidth]{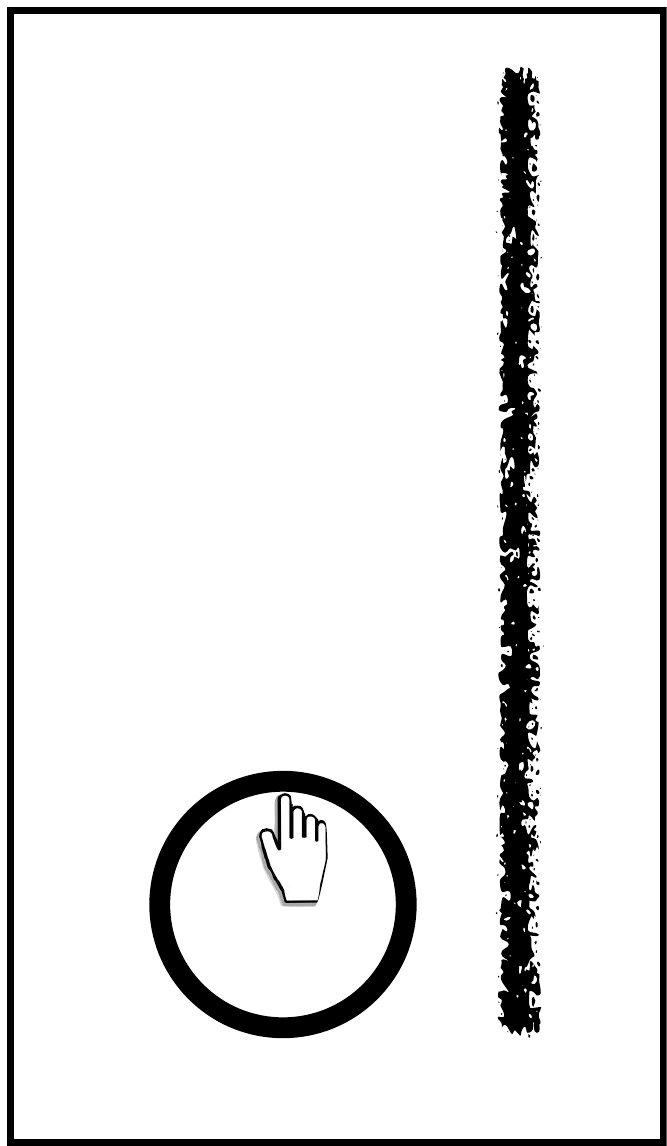}
  \end{minipage}
  \hfill
  \begin{minipage}[c]{0.15\textwidth}
    \centering%
    \includegraphics[width=0.9\textwidth]{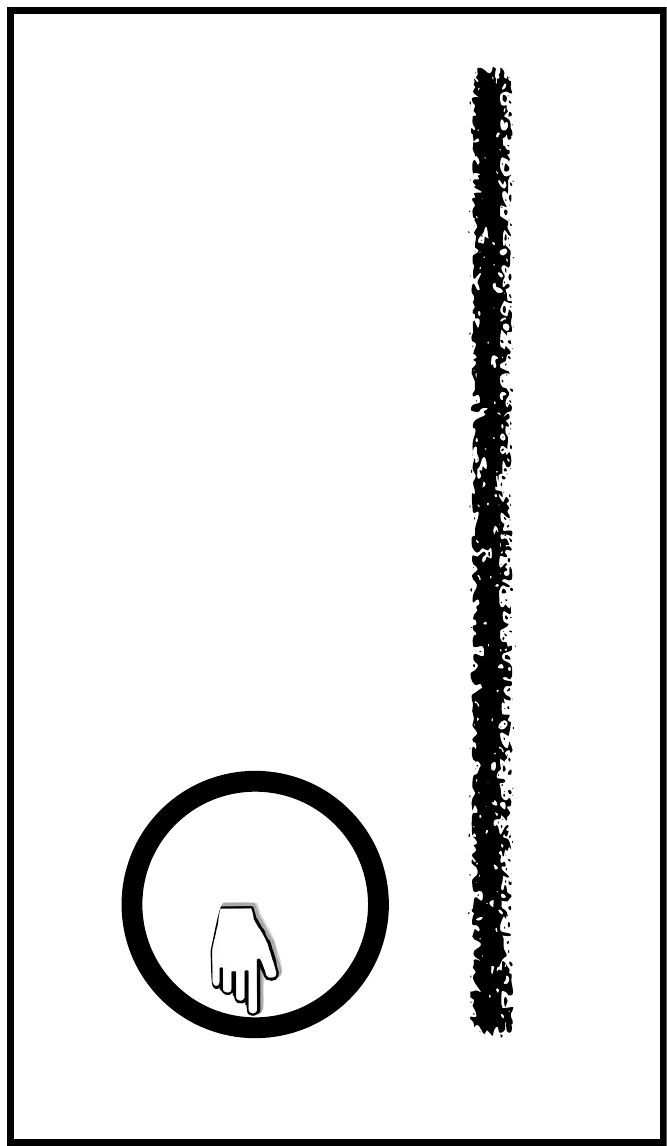}
  \end{minipage}
  \hfill
  \begin{minipage}[c]{0.15\textwidth}
    \centering%
    \includegraphics[width=0.9\textwidth]{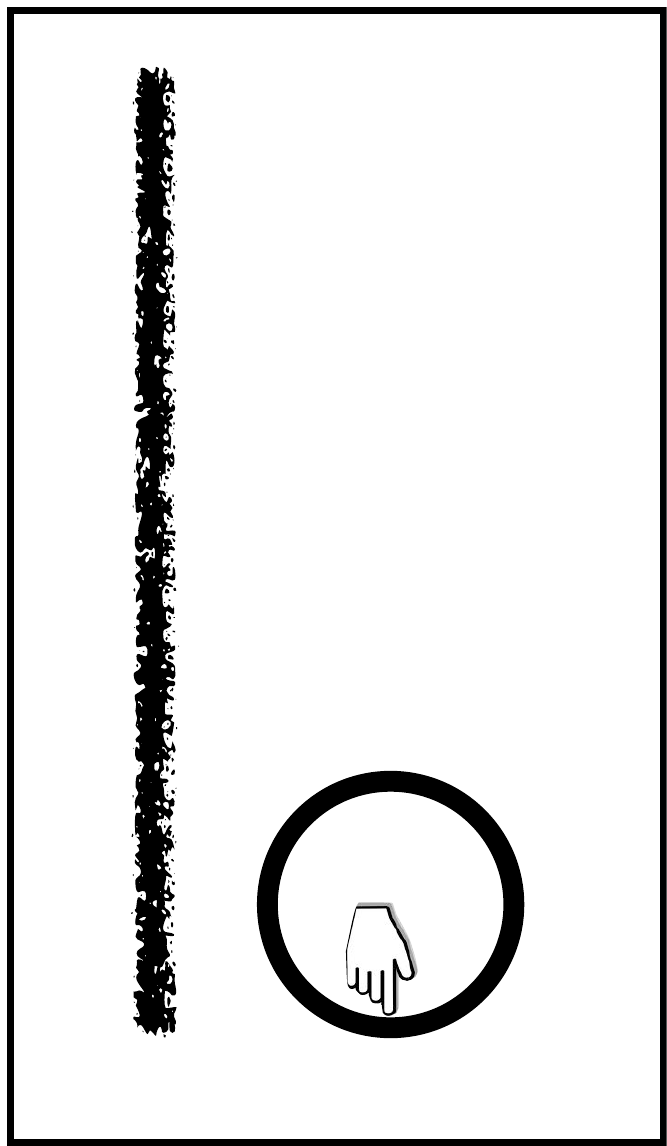}
  \end{minipage}
  \hfill
  \begin{minipage}[c]{0.15\textwidth}
    \centering%
    \includegraphics[width=0.9\textwidth]{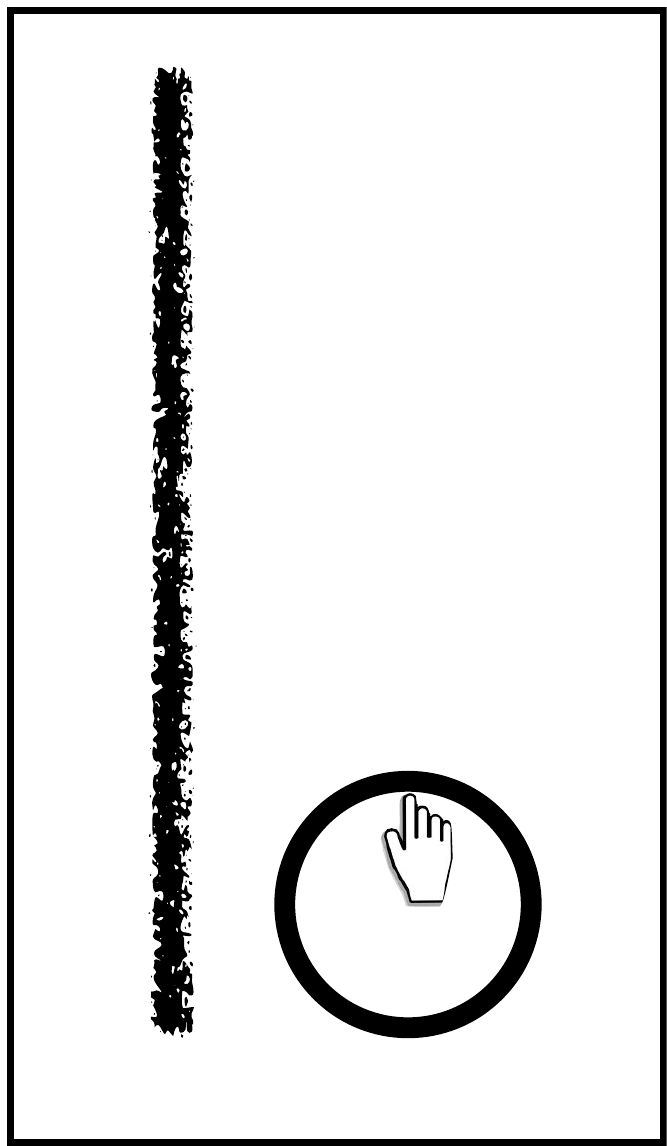}
  \end{minipage}
\caption{The robot in four different canonical positions in which it can roll forward or backward while maintaining something to its side that it can rotate toward, roll forward to, and touch.
}
\label{fig:canonical wall}
\end{figure}

The next layer begins with a set of four options for guiding the agent along a barrier.
The barrier can be to the left or right side, and the agent can move along forward or backward, as shown in Figure~\ref{fig:canonical wall}.
To engage in this behavior, the agent requires estimates of its distance to the obstacles in its vicinity, as were learned in Layer~\FlDTAM. 

The four options are shown in Table~\ref{table:OrfWR}.
Each chooses a single action (\Orf\ or \Orb) under the condition that either the agent's \FDTAM(3) or \FDTAM(9) estimate falls within a certain range, denoted by the threshold values $\theta_1$ and $\theta_2$.
Each option has a non-zero chance of termination in every state, which guarantees eventual termination, so the agent cannot follow the option indefinitely.
Termination also occurs if the agent's \FTA\ forecast is sufficiently high (thus predicting that the robot could extend its finger and touch something, which happens in states where an obstacle is immediately in front of the robot and its movement is therefore impeded).
Furthermore, termination also occurs if the agent's \FDTAM(3) or \FDTAM(9) estimate falls outside the given range.
The range is chosen to correspond to those cases where the agent estimates it can rotate three times and then quickly reach something it can touch, but not so quickly that its motion will likely be impeded should it choose the \Orf\ or \Orb\ action.

\TableO{rfWR, rfWL, rbWR, rbWL}

The four new options allow four new forecasts (Table~\ref{table:FWRf}), each making a prediction about the results of following one of the options until termination.
In all cases, the termination value \termv\ is 0 and the \cumu\ value \accv\ is 1.
Termination is guaranteed to occur with a minimum probability of 0.1, and thus the maximum possible expected value of any of the four forecasts is $\sum_{t=0}^\infty (1-0.1)^t = 10$, which occurs when the agent predicts it can roll forward indefinitely while maintaining a value of either \FDTAM(3) or \FDTAM(9) between the two threshold values.

\TableF{WRf, WLf, WRb, WLb}


Informally, the four forecasts describe how far the robot can roll while maintaining something nearby on its side that it could turn to and touch.
In other words, the forecasts indicate whether there is a wall near the agent's side.
We might think of this as similar to our notion of standing next to a wall, but keep in mind that it is actually a formal description of a specific set of relationships regarding the prediction of action-contingent observations.\footnote{This representation could easily be enhanced through more sophisticated options that enable more subtle predictions, allowing (for example) the agent to follow a slightly curved wall by making small course corrections (rotations) as it rolls.}

\subheading{Verification and Learnability}
When the agent's \FDTAM(3) forecast is between $θ_1$ and $θ_2$ and the agent takes a policy action (\Orf), the agent can compare its \FWRf\ forecast to that made at the following time step, which is estimated mostly from the camera image.
The function approximator will learn to distinguish those images in which the camera is aimed parallel along a wall from those in which it is not, which even a human can do fairly well (hardly a perfect function approximator).
The agent does not need to roll to the end of the wall and test it at every position along the way but will automatically get appropriate updates each time its position changes and its estimates can be improved.
For the two forecasts predicting the results of rolling backward (\FWRb\ and \FWLb), the agent must rely primarily on its \FDTAM\ forecasts, which may be much less accurate; but note that even noisy forecast estimates  may still provide useful information for the function approximator's computation of other forecast estimates.%
%

\subheading{High-level abstraction}
Each of the four newest forecasts predict the outcome of a complex set of interactions the agent could have with its environment, should it choose to do so.
The agent can verify the validity of its predictions, and the function approximator can refine them to increase their accuracy.
The resulting estimates correspond reasonably well to what one might informally think of as detection of a wall next to the agent, an informal interpretation that is now noticeably different from the formal description.
Whereas it may seem very convenient at this point to refer to the forecasts as ``encodings'' of an objective ``concept'' of a ``wall,'' this linguistic convenience could come at the cost of obscuring the fact that forecasts are subjective predictions.
These predictions are individually simple but rely on a deep structure representing quite a complex set of possible interactions with the environment.

\subheading{Low-level foundation}
Each forecast estimate embodies the agent's prediction of the number of time steps it could roll forward (or backward) while maintaining the possibility of rotating three times to the right or left and subsequently rolling forward a small number of times to reach a state in which it can rotate in place and reach a position where it can extend its finger (execute the \Oef\ action) and receive a 1 from the \ST\ sensor.
Thus, the high-level abstractions described in this section are fully connected to the agent's low-level actions and sensory signals.

\newlayer

In this layer the agent learns to predict how far it is from a state where one of the Layer~\FlWRf\  \emph{wall} forecasts (\FWRf, \FWLf, \FWRb\ or \FWLb) has a high estimated value.
As a convenience to the thought experiment, \IA postulate a hidden feature or ``alias'' \AWLR\ (Table~\ref{table:AWLR}) computed by the function approximator to have a value of 1 whenever any of the wall forecasts does.\footnote{This does not imply that the function approximator \emph{must} compute such a function explicitly, only that it \emph{could} do so: all the information is now available to estimate this value with a feedforward computation.
Assume for simplicity that the function is calculated as shown in Table~\ref{table:AWLR}, but it could also be computed in other ways.
More of such aliases will appear in later layers.}

\TableA{WLR}

The forecasts in this layer could be represented in multiple ways, including the following two alternatives, which convey some of the variety possible.
The first forecast, \FWDa~(Table~\ref{table:FWDa}), is based on option \OmrW~(Table~\ref{table:OmrW}), which chooses actions at random (uniformly) until \AWLR\ has a value of 1.
(Note that in Table~\ref{table:OmrW}, option \OmrW\ does not have a learned policy, so $\accv$ and $\termv$ are not applicable.)
The advantage of the random policy is that it is simple to implement and always available.
The agent can use it as the policy in any number of forecasts to provide a rough estimate of how many time steps are needed to reach a desired state.
(As with forecasts~\FnWRf-\FnWLb, the maximum ideal value of \FWDa\ is bounded because the policy has a non-zero minimum termination probability.)

\TableF{WDa, WDb}

The second forecast, \FWDb\ in Table~\ref{table:FWDa}, is based on option \OmCWp\  (Table~\ref{table:OmrW}), whose policy is learned.
The policy of option \OmCWp\ learns to maximize the value of \AWLR\ (\ie\ tries to reach a state as quickly as possible in which \AWLR\ has a value of 1).
This learned option provides more focused information than the random option \OmrW\ because it calculates the \emph{minimal} number of actions required to reach a state in which $\termv=\AWLR =1$.


\TableOCZ{mrW,mCWp}


\subheading{Verification and Learnability}
As in all previous layers, the agent receives target values at every time step when the options are applicable, thus allowing it to verify and learn from its predictions and their actual outcomes.
The agent can update its \FWDa\ forecast every time it takes an action, because all actions are part of the random policy and thus every action results in some degree of learning.
The \FWDb\ forecast is updated whenever a \OmCWp\ policy action is taken.
(For clarity and convenience, as before assume that the forecast is learned only after the policy has been learned.)
The \OmCWp\ policy can be learned with the camera and \FDTAM\ map, at least in those places in the microworld where the agent is relatively close to a wall.
However, if the agent is not near a wall, or if it is near a barrier that does not form a wall (see Figure~\ref{fig:world}), then it may not have enough information to find a path to a place where \AWLR\ is 1.
Therefore, though the information it provides is less precise, the random policy may sometimes be superior for use in forecasts to a learned policy, because it provides predictable results in situations where a dependable policy cannot be learned.


\subheading{High-level abstraction and low-level foundation}
The \FWDa\ and \FWDb\ forecasts are the most abstract so far, built on \numberstringnum{\thestagecounter} previous layers.

Informally, one can think of these forecasts as representing roughly how far the agent is from a wall, though this informality conceals the fact that the agent never posits nor assumes the existence of walls, objects, nor anything else beyond the regularities in its sensorimotor stream.
Instead, the agent merely makes predictions about the outcomes of its behavior, but through forecasts those predictions can be so complex, nested, and intertwined with relationships that they evoke in \us descriptions of objects familiar to us.
But the \FAbbr{WD} forecasts are simply predictions, having low values when the agent predicts it can quickly reach a state from which it can choose actions \Onrf\ or \Onrb~(\Orf\ or \Orb) for several time steps while maintaining the ability to rotate 90° (choose the same rotation action three times successively) and thereby reach a place where it can roll forward (\Orf~\!)
a small number of times to reach a state where it can rotate to a place where it can extend its finger (\Oef~\!)
and receive a signal of 1 from its touch sensor.
The description is quite complex---but formed exclusively from the sensorimotor stream.

\newlayer

This layer consists of a map for the agent much like the \FTMname~(\FTM) and the \FDTAMname~(\FDTAM) forecasts.
As with the \FDTAM\ forecasts, the new forecasts measure the agent's distance to states meeting a certain condition.
In the current case that condition is defined by the \AWA\ alias (Table~\ref{table:AWA}), which has a value of 1 when the agent's \FTA\ estimate is sufficiently high (predicting it can rotate to something it can touch) and the \FWDa\ estimate is sufficiently low (predicting it is near a wall).

\TableA{WA}

A new option \OrfW\ (Table~\ref{table:OrfW}) helps calculate steps to \AWA, choosing action~\Orf\ until termination, which occurs reliably when $\AWA=1$ or with a small minimum probability otherwise (just as with options $\OnrfWR-\OnmCWp$).

\TableO{rfW}

The map is based on the forecast \FDW~(Table~\ref{table:FDW}), which uses option \OrfW\ to estimate the number of \Orf\ actions to a state where $\AWA = 1$.
This forecast and the map that goes with it (\FDWM, Table~\ref{table:FDW}) is not more complex than the previous two, so \IA will forego a lengthy description.
Similar to the previous cases, $\FDWM(0)\equiv\FDWM(12)\equiv\FDW$.

\TableF{DW,DWM}


\begin{figure}
  \centering%
  \includegraphics[height=0.25\textheight]{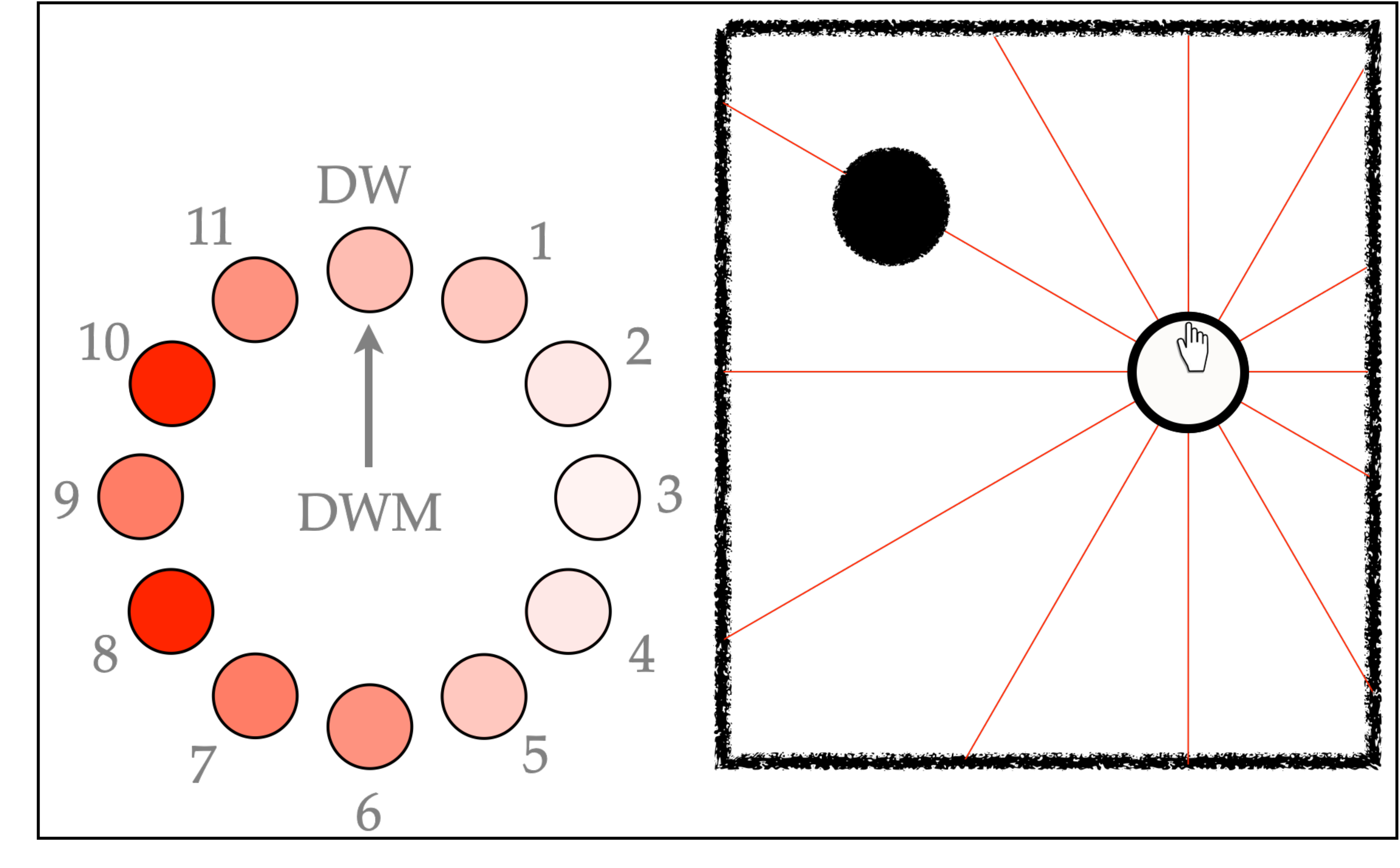}
  \caption{The \FDWM\ forecasts estimate the distances of (number of \Orf\ actions to) the walls surrounding the robot.
Unlike the distances to states where the robot can touch (Figure~\ref{fig:FDTAM}), these estimates make predictions about the robot's ability to reach \emph{walls}---\ie large contiguous collections of states where it can roll foreword while maintaining something nearby on its side that it can rotate toward, roll to, and touch.
Thus, the estimate from \FDWM(10) is much greater than the \FDTAM(10) estimate in the first position of Figure~\ref{fig:FDTAM}.}
\label{fig:FDWM}
\end{figure}

\subheading{Verification and Learnability}
Verification and learnability are not different in this case from previous cases. With each rotation, the agent can compare its predictions from one time step to those of the next.

\subheading{High-level abstraction and Low-level foundation}
Figure~\ref{fig:FDWM} shows the agent's \FDWM\ estimates in a situation similar to that shown in Figure~\ref{fig:FDTAM}, highlighting the differences between the \FDWM\ and \FDTAM\ distance maps.
For the first time the agent can now truly ``see'' its surroundings.
When it looks out from the center of the room, its camera provides far more information than the random pixels of Layer~0.
The camera now informs the function approximator, which can use a large amount of the visual information to fill in a complex network of predictions about how the agent can expect to interact with its surroundings \emph{even though its predictions are all ultimately about a single binary sensor}.
When it rotates in place, it takes in more visual information and fills in its predictions.
After turning 360° it has estimates about its possible interactions with all the walls surrounding it, predicting how far it must travel in each direction to reach a wall (\ie a place where it is adjacent to something that it can travel along for an extended number of time steps while maintaining the ability to rotate by 90° to a state where it can roll forward, extend its finger, and receive a signal of 1).

\newlayer

\begin{figure}
  \def\nfigs{0.132}
  \begin{minipage}[b]{\nfigs\textwidth}
    \centering%
    \includegraphics[width=0.99\textwidth]{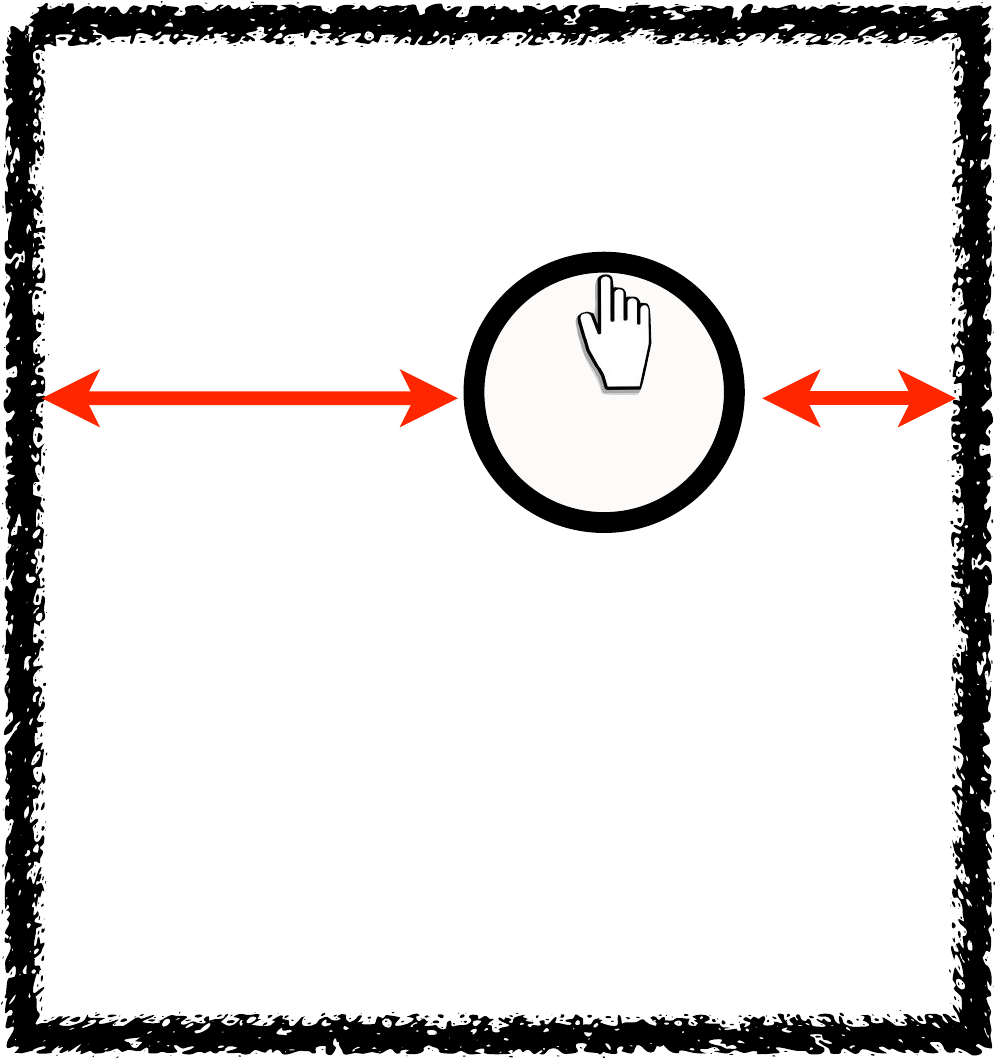}
    \ALRFS
  \end{minipage}
  \begin{minipage}[b]{\nfigs\textwidth}
    \centering%
    \includegraphics[width=0.99\textwidth]{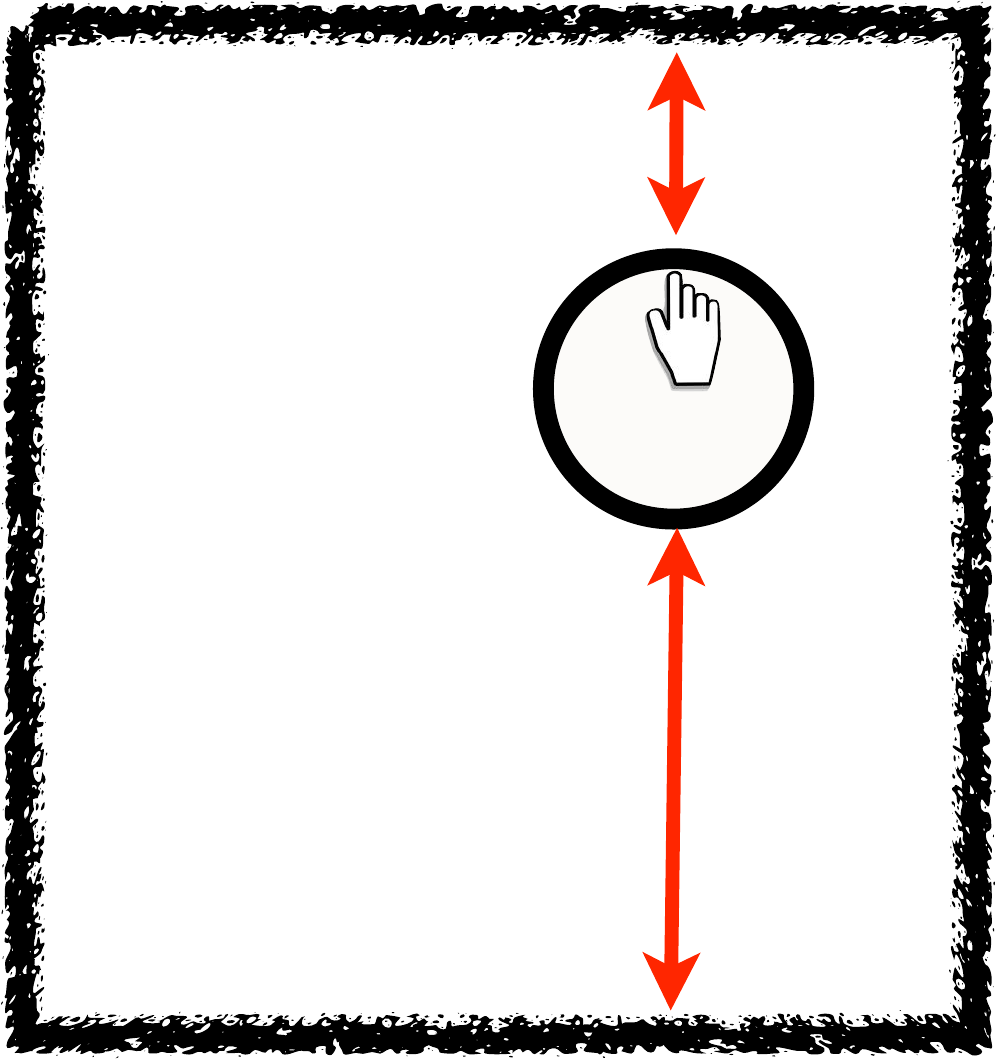}
    \AFBFS
  \end{minipage}
  \begin{minipage}[b]{\nfigs\textwidth}
    \centering%
    \includegraphics[width=0.99\textwidth]{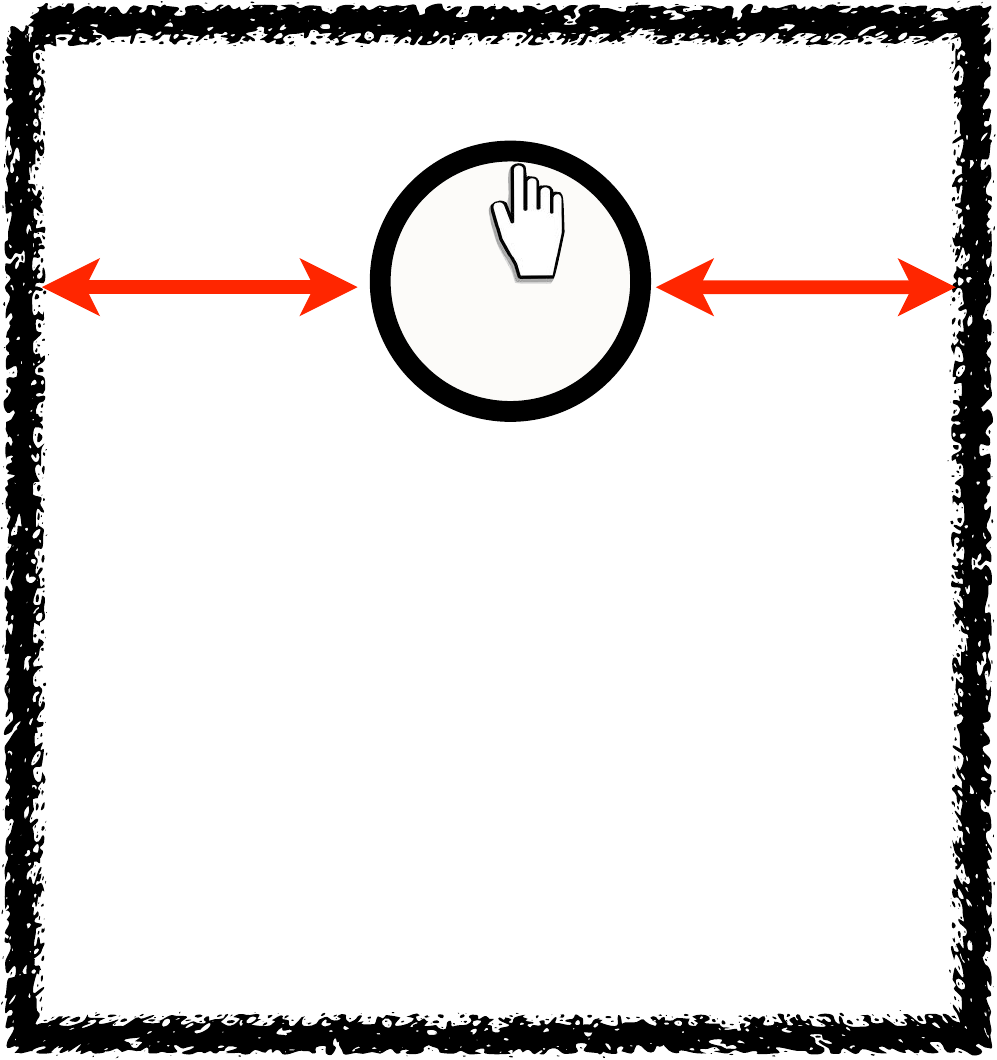}
    \ALRC
  \end{minipage}
  \begin{minipage}[b]{\nfigs\textwidth}
    \centering%
    \includegraphics[width=0.99\textwidth]{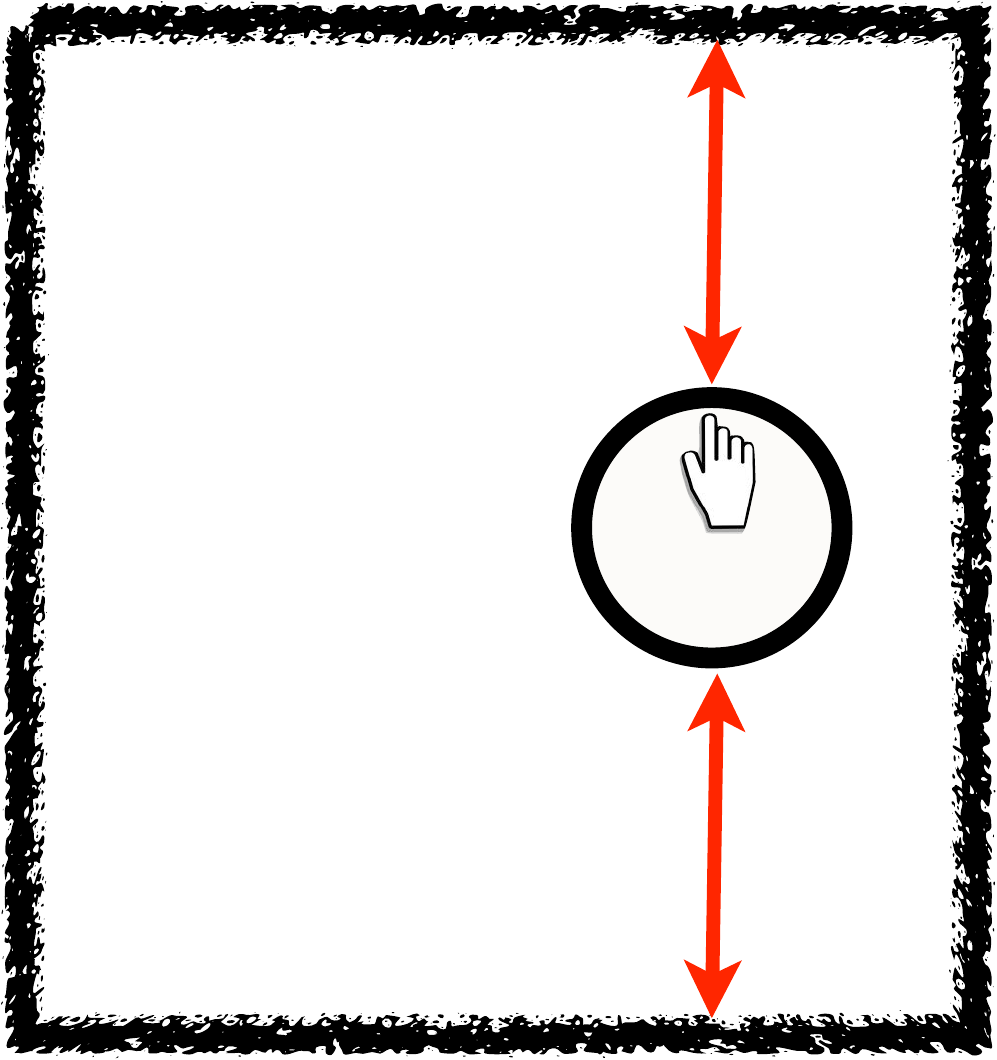}
    \AFBC
  \end{minipage}
  \begin{minipage}[b]{\nfigs\textwidth}
    \centering%
    \includegraphics[width=0.99\textwidth]{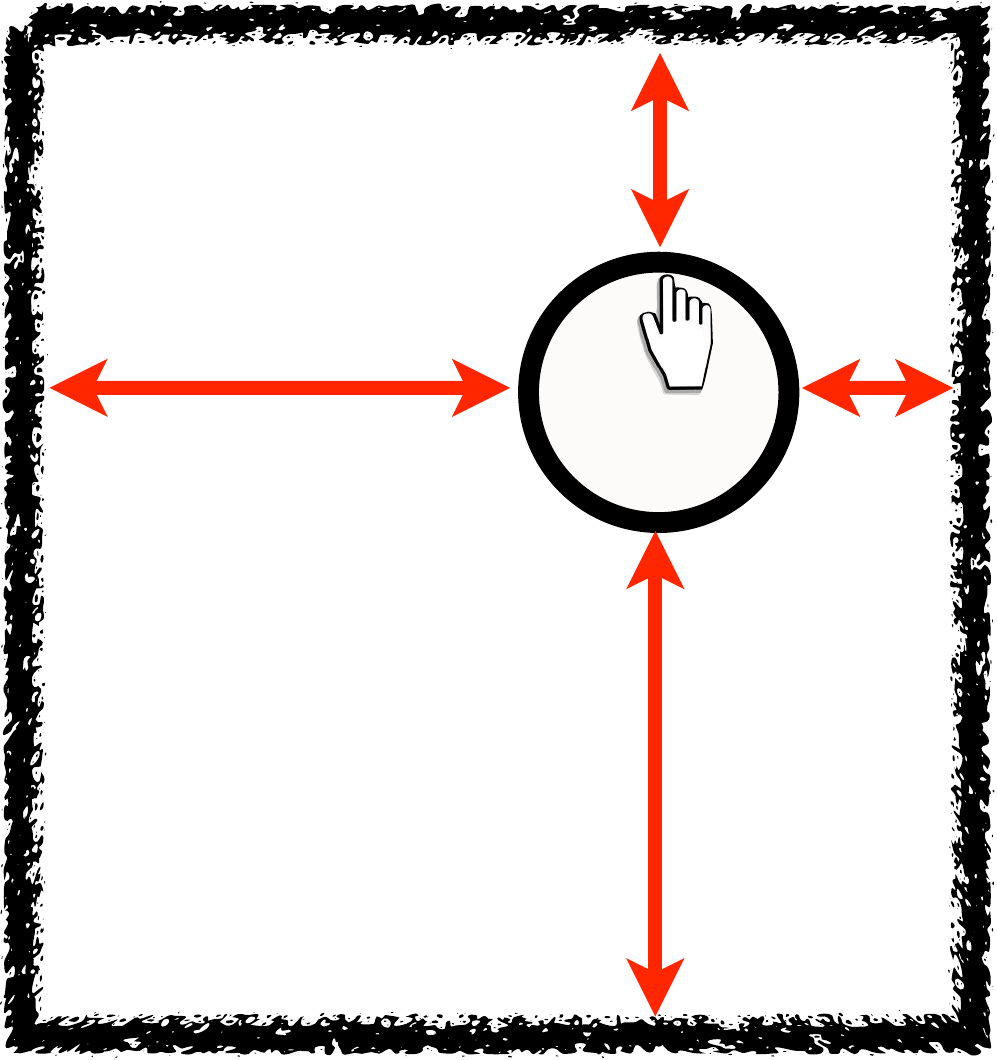}
    \AR
  \end{minipage}
  \begin{minipage}[b]{\nfigs\textwidth}
    \centering%
    \includegraphics[width=0.99\textwidth]{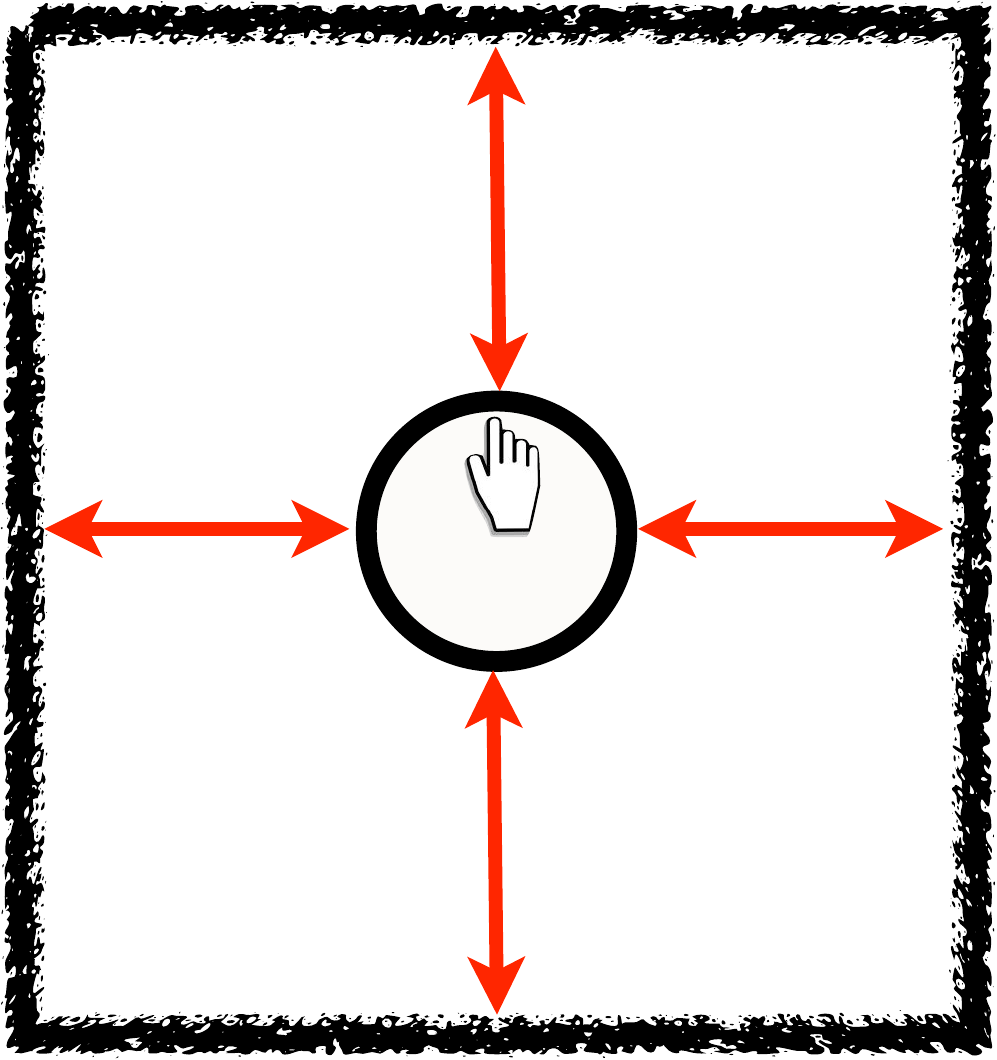}
    \ACR
  \end{minipage}
  \begin{minipage}[b]{\nfigs\textwidth}
    \centering%
    \includegraphics[width=0.99\textwidth]{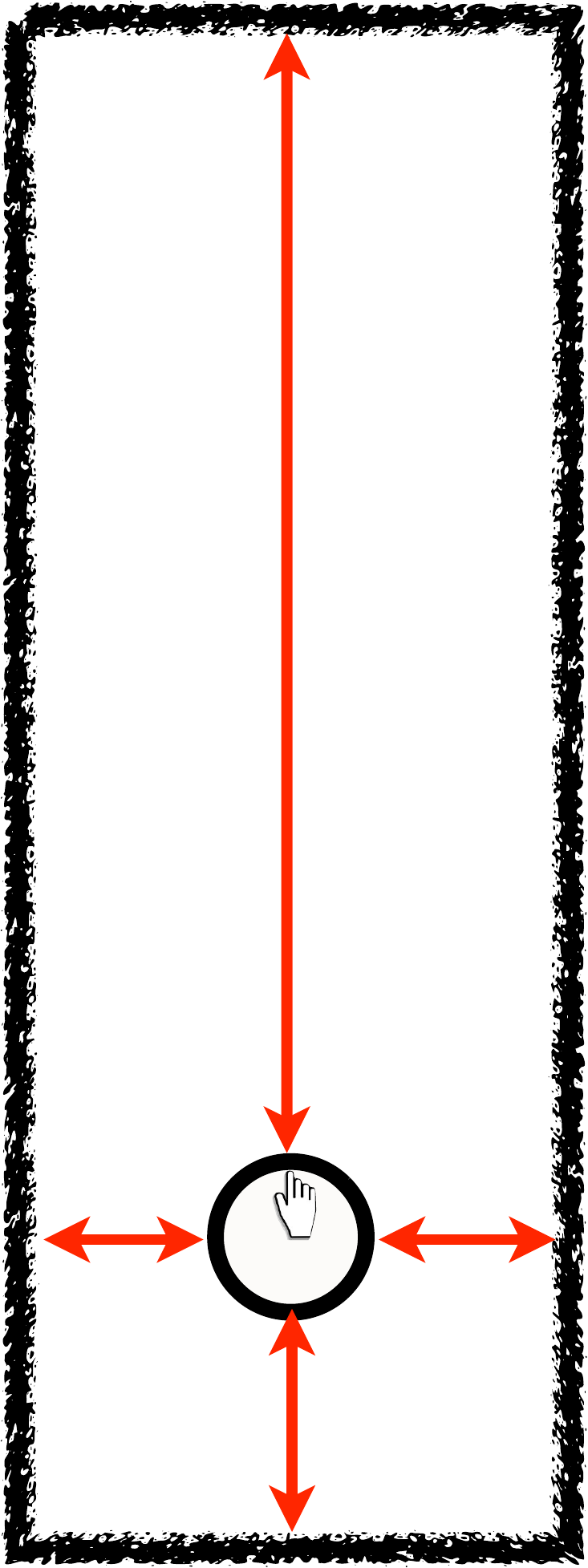}
    \AMH
  \end{minipage}
  \caption{Example positions describing (from left to right): Left-right free space~(\ALRFS), front-back free space~(\AFBFS), being left-right centered~(\ALRC), front-back centered~(\AFBC), surrounded by walls on all sides~(\AR), centered within a room~(\ACR), and positioned in the middle of a hallway~(\AMH). All aliases are defined in Tables~\ref{table:ALRFS} and~\ref{table:AR}.}
\label{fig:alias diagrams}
\end{figure}

The wall map created in the previous layer is extremely powerful and allows the computation of a large number of useful aliases---Tables~\ref{table:ALRFS} and~\ref{table:AR} list several; most  are shown diagrammatically in Figure~\ref{fig:alias diagrams}.
The first (\ALRFS) allows estimation of the combined distance to walls on either of the robot's two sides, \ie\ its left-right free space.
The next (\AFBFS) estimates total free space to the nearest wall in front and back of the robot.
The third (\ALRC) estimates whether the robot has roughly equal free space between walls on its left and right.
The fourth (\AFBC) estimates whether it is centered between walls in front and back of it.

\TableA{LRFS, FBFS, LRC, FBC}

Notice that, despite the convenience of informal expressions such as ``free space'' and ``centered,'' these aliases are strictly feedforward functions of forecasts, all of which are entirely (and formally) defined as predictions about the sensorimotor stream.\footnote{For clarity these definitions leave out some easily added details (such as specification of the robot's orientation with respect to the walls) that might prove useful in an actual robot (or  a more extensive thought experiment).}

The next alias, \AR\ (Table~\ref{table:AR}), defines the situation where the agent estimates that it can quickly reach a wall on four sides.
\ACR\ estimates whether the robot is in a centered position within a room (both left to right and front to back) and \AMH\ is 1 exactly when the agent is in a left-right centered position where the left-right free space is small and the front-back space is large.
The remaining aliases allow estimation of the overall size of the space containing the agent (\ARA) and discrimination of that size into categories (\ASR\ and \ALR).

\TableA{R, CR, MH, RA, SR, LR}

New options can now be made based on any of these computed values.
Table~\ref{table:Ogmh} shows three, one that learns to maximize the \AMH\ value (\Ogmh), one that learns to maximize the \ACR\ value (\Ogcr), and one that simply rolls forward until the \AR\ value is 1.

\TableOCZ{gmh, gcr, gfr}

\subheading{Verification, Learnability and abstraction}
All the aliases are simply feedforward functions of previously defined forecasts, each of which is learnable and verifiable.
These aliases are presented here to show the expressive power of the forecasts already defined.
The agent's forecast estimates and aliases allow it to assess whether it is inside a room, how large the room is, whether it is in a hallway, and whether the robot is in the center of the room.
All of this can be derived from the predictions the agent makes about its future interactions and all significantly more abstract than the agent's raw sensors and effectors.
The camera, which provides much of the information the agent uses to make its predictions, reveals nothing outside of its 30° field of view.

The options \Ogmh\ and \Ogcr\ must be learned, but all the information necessary for learning these options is available: the agent's four \FDWM\ estimates are updated at every time step and reflect the agent's estimates of how far away the surrounding walls are.
It can use these to orient itself and move toward the location that maximizes the target value for the two options.

At this point the agent has collections of forecasts and options that give it high-level features for navigating the microworld that closely resemble the descriptive features \IA used in the text to initially describe the microworld.
All have been built from forecasts—predictions about the results of following a policy until its termination.

\newlayer

\begin{figure}[tb]
  \centering%
  \includegraphics[height=0.25\textheight]{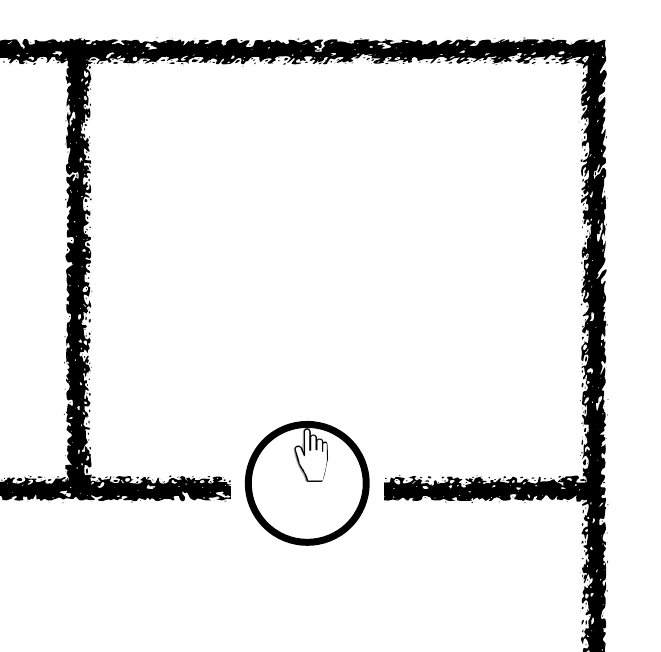}
  \caption{The robot finds itself in a canonical doorway position---a state in which the \FDR\ forecast will have a low value, and where it can rotate three times either to the left or right and reach a state from which it can roll forward, extend its finger, and receive a signal of 1 from its touch sensor.}
\label{fig:doorway}
\end{figure}

This layer attempts to use the previously built forecasts, aliases, and options together to capture the unique set of predictions the agent can make from the common situation diagrammed in Figure~\ref{fig:doorway}.
From this position, the robot can predict that if it rotates 90° in either direction, then it will be able to roll a short distance forward and reach a state where it can extend its finger and receive a signal of 1 from its touch sensor.
It can also predict that if it rolls forward it will soon arrive in a state where its \AR\ alias has a value of 1.
This combination of predictions is captured by alias~\AD~(Table~\ref{table:AD}).

\TableA{D}

Alias~\AD\ relies on the previously defined layers as well as a new forecast~\FDR~(Table~\ref{table:FDR}), which estimates the number of time steps needed to reach a room by rolling forward.
The \AD\ alias is a fairly simple function that only partially corresponds to the common notion of a doorway; however, if we wished to capture more of the subtleties that to us are associated with the state of being in a doorway, this would certainly be possible---capturing, for example, the expectation of finding a wall on one or both sides of the doorway should the agent venture out to look for one, etc.
Again, such details are not different in spirit from those already described.

\TableF{DR}

It should be clear at this point that this layer is not different from the others: it can be verified and learned from experience and it is built entirely from the sensorimotor stream.



%
\refstepcounter{stagecounter}
\subsection*{Layer~\thestagecounter\ and Beyond...}

The techniques shown above can be reused to continue the progression toward higher layers of abstractions, where each layer knits together its existing forecasts, options, and aliases to form new ones.
One could, for example, design additional maps that estimate the agent-centric distances to the forecasts and aliases given in the previous sections: rooms, hallways, doorways, etc.
With those maps the agent would gain a subjective awareness of its distances to all the environmental landmarks that \IA have been using when discussing the environment from \our external perspective.
The agent can then build new forecasts to make new predictions about how these forecasts and aliases are related.
For example, a “kitchen” alias might have a value of 1 whenever a certain group of forecasts have certain values, encompassing the agent's specific expectations about how it can reach nearby walls and doorways.
“Living room” could be a set of states from which the “kitchen” states can be reached by rolling through a “doorway.”

All the different rooms in a floor plan could be distinguished by the various expectations the agent has about the options it can execute to reach each room from the others.
The same layering could then continue to the level of houses, predicting how each floor plan is reachable from the others.
Once the agent has these forecast definitions and can estimate their values, it has a highly informed picture of its world.
The picture is not a perfect one-step model, nor any kind of one-step model, but it makes predictions about what the agent can expect to experience in a set of common situations that allows it to navigate its environment.
Yet all of this is built out of the sensorimotor stream, from forecasts ultimately consisting of predictions about a single sensor.

\subsection{Discussion}

As is the purpose of thought experiments, this one examined a single issue and ignored tangential issues better solved elsewhere; in particular, it imagined the existence of a perfect feedforward function approximator with access to all experience necessary for convergence to accurate values.
This support from the imagination made it possible to focus exclusively on the representational abilities of the forecasts.
In an actual system, of course, these tangential issues impact the results, which could obscure the question under study.

The goal of the thought experiment was to investigate the ability of forecasts to capture high-level knowledge, starting from the raw sensorimotor stream and building up layer by layer in an isolaminar fashion.
At each layer, methods of construction were the same: simpler components were combined into slightly more complex structures, the degree of abstraction increasing layer by layer.
These new structures were then able to serve as the components from which more complex structures were built.
The initial components were sensory signals and motor signals.
These were combined into forecasts and options that served as the building blocks for later forecasts and options.
At no point in layers 0 through~\thestagecounter\ did the representation incorporate techniques, mechanisms, knowledge, or skills other than those described in previous sections, and no source of information was available to the agent other than its raw sensorimotor stream.
Thus, all the agent's knowledge was always expressed in the language of forecasts.
All knowledge was learnable, tunable, verifiable by the agent through its interaction with the world.
By the end of the demonstration, layer~\thestagecounter, the forecasts represented subjective abstractions, patterns from the sensorimotor stream reminiscent of common notions of walls, doorways, rooms and floor plans.
It therefore does not seem unreasonable to conclude that forecasts can indeed capture high-level knowledge. 

Thus, the thought experiment has shown that at least some of what seems intuitively to be high-level knowledge can in fact be expressed exclusively in terms of the raw sensorimotor stream.
By making long-term, policy dependent predictions about the sensorimotor data, forecasts organize the statistical regularities of the sensorimotor stream, and provide an isolaminar mechanism for the construction and representation of high-level knowledge and skills.
Starting from a complex stream of raw data that describes only the robot's immediate inputs and outputs, forecasts can provide an agent with a rich understanding of its world, capturing knowledge and making predictions about highly complex, temporally extended, useful abstract quantities.
Whether forecasts can capture all the critical knowledge that an agent might require, or whether there might be limits, is an important question whose answer is not yet known.


%



Does the fact that all forecasts ultimately boil down to predictions about the sensorimotor layer in some way contradict their claim to abstraction?
Could the hierarchy simply be flattened and turned into a single one-step model?
If so, is the hierarchy simply superfluous and unnecessary?
These are fair questions, but they somehow miss the essence of the work, and so \IA would  answer each of them with a \emph{no}. 
Critically, \myA overall goal here has not been to build a one-step model, but to \emph{build up knowledge}, knowledge as a scaffold for learning new knowledge, a continually growing structure that is continually increasing its ability to support new, additional knowledge. 
And forecasts appear to provide the raw materials for building exactly such structures. 





%% file: acknowledgements.tex
This paper began in 2010 as a collaboration with Rich Sutton, having its roots in work we did together beginning in 2004.
It is deeply indebted to his suggestions and to his contributions of important ideas and, perhaps more importantly, to the pruning of other ideas.
The paper has also benefited from my collaboration with Tom Schaul and our efforts to test out some of these ideas, as reported in our 2013 IJCAI paper~\cite{SchaulRing:IJCAI2013}, and from many thought-provoking comments by Joseph Modayil (2016) and my colleagues at Cogitai (2016--2019).
It has also benefited from feedback on talks I gave on its major contents at the Gatsby Institute in 2012—--especially comments made by Peter Dayan, at Tsinghua University in 2013, and at the University of Alberta in 2017. 
To all those who have suggested improvements and encouraged me to finish the paper and make it public, thank you!


%% file: appendix-forward-view.tex
\subsection{Forecast forward view}
\label{sec:forwardview}
This appendix derives a one-step, forward-view update for the forecast as defined in Equation~\ref{eq:ideal}:
\begin{equation}
\forecastfun(\state) = \Exp\left[\accv_1 + \accv_2 + \ldots + \accv_{k-1} + \termv_k \mid \pi, β, \state_0 = s \right].
\end{equation}

For the current state $\state_0$, and for termination at future time step $k$, %
\begin{align}
\label{eq:termination-at-k}
\forecastfun_k(\state) & = \Exp\left[\accv_1 + \accv_2 + \ldots + \accv_{k-1} + \termv_k \mid \pi, β, \state=\state_0, \text{termination at } k \right].
\end{align}
However, $k$ is unknown and can be any non-negative integer; therefore,
\begin{align}
  \nonumber
  \forecastfun(\state) & = \Exp[\forecastfun_k(\state)\mid β] 
  \\
  \nonumber
  &= \sum_{k=0}^\infty P(\text{termination at } k) \forecastfun_k(\state) 
  \\
  \label{eq:forecastfunction}
  &=
       \sum_{k=0}^\infty
          \sum_{\state'\in\allStates} 
            P_k(\state,\state')\ 
            \left[\beta(\state') \termv(\state') 
                   + (1-\beta(\state')) \accv(\state') \right],
\end{align} 
where the states that the agent visits when following the policy $\pi$ are treated as a Markov chain with a special absorbing state, $\absorbingState$, in which $\accv = \termv = 0$ and to which the MDP can transition with probability $β(\state')$ from every state $\state'$ reachable from $\state \in I$; $P_k(\state,\state')$ is the probability that the agent will be in $\state'$ exactly $k$ steps after visiting $\state$ (while following π); and $\accv(\state)$ is the average $c$ value received by taking a policy action in $\state$.
It is assumed that all options will eventually terminate, thus avoiding any infinite sums.\footnote{One way to implement this in practical applications---though not in this paper---would be to constrain β to (0,1].}
\begin{align}
\nonumber
\forecastfun(\state) &= 
       \sum_{\state'\in\allStates} 
         \sum_{k=0}^\infty
            P_k(\state,\state')\ 
            \left[\beta(\state') \termv(\state') 
                   + (1-\beta(\state')) \accv(\state') \right] 
\\
\nonumber
  &=
       \sum_{\state'\in\allStates} 
       \left[
            P_0(\state,\state')\ 
            \left(
              \beta(\state') \termv(\state') 
              + (1-\beta(\state')) \accv(\state')
            \right)
    + \sum_{k=1}^\infty
            P_k(\state,\state')\ 
            \left(
              \beta(\state') \termv(\state') 
              + (1-\beta(\state')) \accv(\state')
              \right)
            \right] 
\\
\nonumber
  &=
    \beta(\state) \termv(\state) 
          + (1-\beta(\state)) \accv(\state) 
          + 
          \sum_{\state'\in\allStates} 
            \sum_{k=1}^\infty
                   P_k(\state,\state')\ 
                   \left[\beta(\state') \termv(\state') 
                           + (1-\beta(\state')) \accv(\state')
                   \right] 
\\
\nonumber
  &=
    \beta(\state) \termv(\state) 
    + (1-\beta(\state)) \accv(\state)
          + 
          \sum_{\state'\in\allStates} 
          \sum_{k=1}^\infty
          \sum_{\state''\in\allStates}
          P_1(\state,\state'')  
             P_{k-1}(\state'',\state')\ 
                   \left[\beta(\state') \termv(\state') 
                           + (1-\beta(\state')) \accv(\state')
                   \right]
\\
\nonumber
  &=
    \beta(\state) \termv(\state) 
    + (1-\beta(\state)) \accv(\state)
          + 
          \sum_{\state'\in\allStates} 
          \sum_{\state''\in\allStates}
          P_1(\state,\state'')  
          \sum_{k=0}^\infty
             P_{k}(\state'',\state')\ 
                   \left[\beta(\state') \termv(\state') 
                           + (1-\beta(\state')) \accv(\state')
                   \right] 
\\
\nonumber
  &=
    \beta(\state) \termv(\state) 
          + (1-\beta(\state)) \accv(\state) 
          + 
          \sum_{\state''\in\allStates}
          P_1(\state,\state'')  
          \sum_{\state'\in\allStates} 
          \sum_{k=0}^\infty
             P_{k}(\state'',\state')\ 
                   \left[\beta(\state') \termv(\state') 
                           + (1-\beta(\state')) \accv(\state')
                   \right] 
\\
\nonumber
  &=
    \beta(\state) \termv(\state) 
          + (1-\beta(\state)) \accv(\state) 
          + 
          \sum_{\state''\in\allStates}
          P_1(\state,\state'')
          \forecastfun(\state'')
 \\
\commentout{
\nonumber
  &=
    \beta(\state) \termv(\state) 
          + (1-\beta(\state)) \sum_\action \pi(\state,\action) \accv(\state,\action)
          + \sum_{\state'\in\allStates}
             (1-\beta(\state)) \sum_a \pi(\state,\action) Pr(\state,\action,\state')
             \forecastfun(\state') 
\\
}
\label{eq:forecastfunction1}
  &=
    \beta(\state) \termv(\state) 
          + (1-\beta(\state)) \accv(\state) 
          + \sum_{\state'\in\allStates}
             (1-\beta(\state)) \sum_a \pi(\state,\action) Pr(\state,\action,\state')
             \forecastfun(\state') 
\\
\commentout{
\nonumber
  &=
    \beta(\state) \termv(\state) 
          + (1-\beta(\state)) 
              \sum_\action \pi(\state,\action)
            \left[
              \accv(\state,\action)
              + \sum_{\state'\in\allStates}
              Pr(\state,\action,\state')
              \forecastfun(\state')
              \right]
\\
}
\label{eq:forecastfunction2:A}
  &=
    \beta(\state) \termv(\state) 
          + (1-\beta(\state)) 
          \left(
             \accv(\state) 
             + 
             \Exp[\forecastfun(\state' \mid π,β, \state=\state_t,\state'=\state_{t+1})]
             \right).
\end{align} 
This holds for any state $\state'$ reachable from any $\state\in I$ before termination of π.
While the agent is following π, this recursive definition allows a temporal difference update, just as with GQ~\cite{GQ}, because the states are experienced with those probabilities given in Equation~\ref{eq:forecastfunction1}.
However, when the agent is following a different policy, then corrections are required to compensate for the differing distributions.


%% file: appendix-collected-tables.tex
\subsection{Collected Tables}
\label{sec:collected-tables}
The tables of Section~\ref{sec:demo} are combined and indexed here for easy reference.

\newcommand{\summarytables}[2]{
  \begin{table}[h]
    \centering
    \small
    \begin{tabular}{cccccc}
      \mysplit{#1 \\ Number} & Name & Abbrev & Layer \\
      \midrule
      \expandafter\docsvlist\expandafter{#2}
      \bottomrule
    \end{tabular}
  \end{table}
}

\renewcommand*{\do}[1]{
  \FaNumber{#1} & \FaName{#1} & \FAbbr{#1} & \FaLayer{#1}\\}

\summarytables{Forecast}{\allforecastabbreviations}

\renewcommand*{\do}[1]{
  \OaNumber{#1} & \OaName{#1} & \OAbbr{#1} & \OaLayer{#1}\\}

\summarytables{Option}{\alloptionabbreviations}

\renewcommand*{\do}[1]{
  \AaNumber{#1} & \AaName{#1} & \AAbbr{#1} & \AaLayer{#1}\\}

\summarytables{Alias}{\allaliasabbreviations}

\clearpage
